\documentclass{article}
\PassOptionsToPackage{numbers, compress}{natbib}
\usepackage[preprint]{arxiv_2026}

\usepackage[utf8]{inputenc} %
\usepackage[T1]{fontenc}    %
\usepackage{hyperref}       %
\usepackage{url}            %
\usepackage{booktabs}       %
\usepackage{amsfonts}       %
\usepackage{nicefrac}       %
\usepackage{microtype}      %
\usepackage{xcolor}         %
\usepackage{wrapfig}
\usepackage{graphicx}
\usepackage{enumitem}
\usepackage{subcaption}
\usepackage{multirow}
\usepackage{booktabs} 
\usepackage{graphicx}
\usepackage{xcolor}   
\usepackage{float}
\usepackage{algorithm}
\usepackage{algpseudocode}
\algrenewcommand\algorithmicrequire{\textbf{Input:}}
\algnewcommand{\LineComment}[1]{\Statex \hfill $\triangleright$ #1}
\usepackage{caption}
\usepackage{amsmath}
\usepackage{amssymb}
\usepackage{mathtools}
\usepackage{amsthm}
\usepackage{makecell}
\usepackage{booktabs}
\usepackage{multirow}
\title{Dual-Process Atomic Skill Learning: Decoupling Semantic Reasoning and Real-Time Control}

\author{
  Jun Chen$^{1*}$, Erdemt Bao$^{2*}$, Wenlong Dong$^3$, Jierui Liu$^4$, Qi Cai$^4$, Hao Wan$^1$, Shaopeng Li$^5$, \\
  \textbf{Weijun Qin}$^5$, \textbf{Jing Liang}$^1$, \textbf{Huiping Zhuang}$^{4\dagger}$ \\[5pt]
  \normalfont\scriptsize $^{1}$University of Electronic Science and Technology of China \quad $^{2}$Huazhong University of Science and Technology \\[-2pt]
  \normalfont\scriptsize $^{3}$Southern University of Science and Technology \quad $^{4}$South China University of Technology \quad $^{5}$EbTech Co. Ltd. \\[-1pt]
  \normalfont\scriptsize $^*$Equal contribution. \quad $^\dagger$Corresponding author. Email: \texttt{hpzhuang@scut.edu.cn}
  \vspace{-5pt}
}

\begin{document}

\maketitle

\begin{abstract}
  Language-conditioned Imitation Learning (IL) is essential for enabling robots to perform complex tasks following natural language instructions. However, generalizing to multi-step compositional tasks remains a significant challenge. While hierarchical approaches attempt to address this by decomposing tasks into atomic skills, existing methods often suffer from training instability and codebook collapse due to the tight coupling between high-level skill reasoning and low-level action generation in joint training paradigms. Inspired by the Dual-Process Theory of cognition, we propose Dual-Process Atomic Skill Learning (\textbf{DASL}), a novel asynchronous hierarchical imitation learning framework that decouples slow semantic reasoning from fast, real-time motion control. DASL comprises a Slow-Frequency Policy that predicts interpretable, discrete skills via Vector Quantization, and a High-Frequency Policy that leverages a latent diffusion model and a Decision Transformer to generate precise actions conditioned on these latent skills. By asynchronously coordinating these modules and utilizing diffusion to structure the latent space, our framework mitigates the skill codebook interference problem common in joint training paradigms. Evaluations across simulation benchmarks and  experiment demonstrate that DASL significantly outperforms state-of-the-art baselines, excelling in skill acquisition and compositional generalization to unseen instructions. \textbf{GitHub page:} \href{https://github.com/Hatakekaka/DASL}{\color[RGB]{0, 32, 128}\texttt{https://github.com/Hatakekaka/DASL}}
\end{abstract}

\section{Introduction}

\begin{wrapfigure}{r}{0.45\textwidth} %
  \vspace{-1.5em} %
  \centering
  \includegraphics[width=\linewidth]{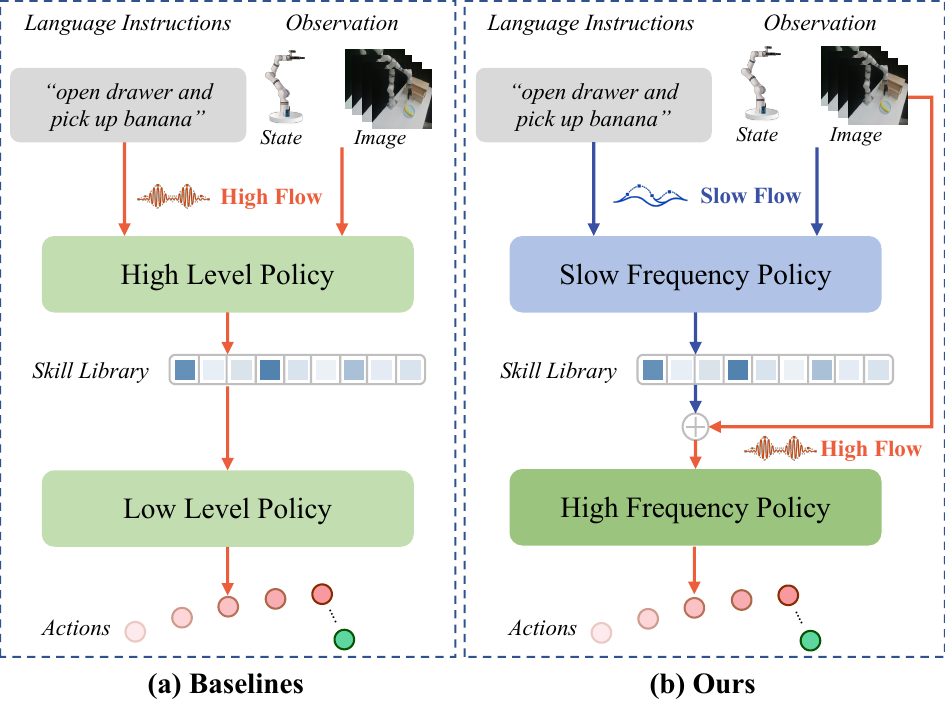} 
  
  \vspace{-0.5em} %
  \caption{\small Baselines vs. our dual-process control. (a) Baselines: synchronous high-frequency skill and action generation. (b) Ours: asynchronous decoupling, with slow-frequency skill reasoning and high-frequency action execution.}
  \label{DASL_introduction}
  \vspace{-3em} %
\end{wrapfigure}

Language-conditioned imitation learning (IL) aims to train robots to execute tasks specified by natural language instructions using paired language-trajectory demonstrations. However, this objective becomes increasingly arduous when tasks involve sequential execution of multiple sub-goals \citep{gupta2019relay,yu2020meta}. A key strategy to address this complexity exploits the inherent hierarchical structure of natural language to transform complex task solving into learning atomic skills. Recombining these atomic skills enables robots to generalize to unseen task configurations, a capability particularly critical in data-constrained regimes where covering the entire task space is infeasible.

Despite progress in hierarchical skill learning methods \cite{lynch2020learning,rosete2023latent}, a significant limitation persists: high-level skill transition and low-level action generation remain tightly coupled. This necessitates end-to-end co-training of both policies, inducing severe instability during skill learning. The latent low-level and language-conditioned high-level policies become intertwined, causing mutual interference. This frequently leads to codebook collapse, forcing reliance on cumbersome multi-stage training strategies or additional constraints to optimize and stabilize the skill codebook \cite{garg2022lisa,liang2024skilldiffuser,ju2024rethinking}.

Inspired by Kahneman's Dual-Process Theory, distinguishing fast, intuitive "System 1" from slow, deliberative "System 2" cognition \cite{shentu2024llms,wen2025dexvla,chenfast}, we argue similar decoupling benefits language-conditioned skill learning. While recent Vision-Language-Action (VLA) models explore dual-system architectures, their application in hierarchical skill acquisition remains limited. We hypothesize asynchronously decoupling high-level semantic reasoning from low-level action generation substantially improves training stability and execution efficiency. Specifically, allowing high-frequency control to operate independently of slower reasoning mitigates interference between skill abstraction and motion generation common in synchronous frameworks.

Building on this, we introduce \textbf{DASL}, an asynchronous hierarchical imitation learning framework explicitly decoupling high-level skill reasoning from low-level action execution (Figure~\ref{DASL_introduction}). DASL comprises three components: a Slow-Frequency Policy, a High-Frequency Policy, and an Asynchronous Coordination mechanism. The Slow-Frequency Policy operates at coarse temporal resolution, processing language and sparse observations to predict fixed-horizon continuous skills, subsequently discretized via Vector Quantization (VQ) for interpretability. The High-Frequency Policy operates at the original control frequency, conditioned on the selected skill. It employs a skill-conditioned latent diffusion model during training to strengthen skill-trajectory correspondence, while a Decision Transformer generates real-time actions at inference. The Asynchronous Coordination mechanism decouples execution schedules, balancing compositional reasoning and precise real-time control.

We evaluate DASL on four simulation benchmarks (LOReL \cite{nair2022learning}, Franka Kitchen \citep{gupta2019relay}, CALVIN \citep{mees2022calvin}, and BabyAI \cite{chevalier2018babyai}) and  experiment, spanning robotic manipulation and grid-world navigation. Empirical results show DASL consistently outperforms state-of-the-art baselines, excelling in skill acquisition and compositional generalization to unseen instructions. Ablation studies further validate the critical roles of asynchronous execution and skill-conditioned diffusion modeling in overall performance.

In summary, our main contributions are as follows:
\begin{enumerate}[leftmargin=2.0em,nosep]
\item We propose DASL, an asynchronous hierarchical framework learning reusable, interpretable atomic skills directly from language-conditioned demonstrations.
\item We design an architecture integrating asynchronous execution and latent skill diffusion. This unifies skill representation and mitigates codebook interference while maintaining high-frequency action execution by decoupling semantic reasoning from low-level motion control.
\item We achieve state-of-the-art performance across simulation benchmarks and  experiment, driven by stable skill learning and strong compositional generalization.
\end{enumerate}

\section{Related Work}
\subsection{Language-Conditioned Imitation Learning}

IL allows embodied agents to learn behaviors directly from expert demonstrations \cite{macmahon2006walk,tellex2011understanding}. Scaling IL to complex multi-task settings demands effective task specification. Unlike goal images or trajectories \cite{nair2018visual,finn2017one}, natural language serves as a uniquely powerful and general interface \cite{garg2022lisa}. Existing language-conditioned IL methods fall into two paradigms: aligning pretrained language embeddings with visual observations for direct action prediction \cite{shridhar2022cliport,kim2024openvla,wen2025dexvla,wang2026rethinking}, or leveraging Large Language Models (LLMs) to decompose instructions into sub-task sequences \cite{brown2020language,ahn2022can,shentu2024llms,liu2025delta,gu2026language}. Instead of using external LLM planners, we adopt pretrained language embeddings to condition a hierarchical policy, executing complex multi-step tasks via structured skill learning.

\subsection{Language-Conditioned Skill Learning}

Traditional skill learning relies on latent variable models \cite{eysenbach2018diversity}, the options framework \cite{konidaris2009skill}, or mutual information maximization \cite{hausman2018learning}. However, these unsupervised methods lack linguistic guidance and exhibit high sample complexity, complicating real-world expermentt. Recent work focuses on learning semantically interpretable skills from language-annotated data. LISA \cite{garg2022lisa} introduced an end-to-end hierarchical paradigm by jointly training language-conditioned high-level and low-level policies. SkillDiffuser \cite{liang2024skilldiffuser} and RoLD \cite{tan2024rold} extend this by using diffusion models to generate diverse actions over discrete skills, while DynaMind \cite{wang2025dynamind} reasons over abstract video dynamics. LADS \cite{jiang2025discrete} decouples high-level policy training from latent plan abstraction. Still, achieving effective skill representations and ensuring high-frequency, stable execution remains challenging.

To address this, we introduce a skill-conditioned diffusion module to regularize latent skill representations, employing a Decision Transformer to generate continuous actions conditioned on discrete skills. This hybrid approach learns reusable, interpretable skills directly from language-conditioned demonstrations, enhancing stability in compositional tasks.

\subsection{Hierarchical Policy Learning}

Hierarchical Policy Learning (HPL) improves sample efficiency and generalization by decomposing complex tasks into sub-goals. Early works used latent variables to span the state space \cite{eysenbach2018diversity}, but static skill encodings limited their adaptive flexibility. Recent advancements \cite{zhang2024extract,wan2024lotus,kim2025uniskill} integrate vision foundation models \cite{oquab2023dinov2,nair2022r3m} to bolster high-level skill abstraction. However, lacking explicit action or language grounding, these representations remain susceptible to visual distractors and often fail to capture meaningful temporal abstractions.

Meanwhile, Dual-System Architectures have gained traction in VLA models \cite{chenfast, bjorck2025gr00t, song2025rationalvla}. This paradigm decouples high-level reasoning from low-level execution, significantly improving system adaptability. Inspired by this cognitive paradigm, we propose an asynchronous architecture for hierarchical skill learning, integrating a slow-frequency high-level planner and a high-frequency low-level executor within a unified model. Crucially, this asynchronous design increases overall action generation frequency while mitigating the common codebook index collapse in joint hierarchical training.

\section{Method}
\subsection{Overview}

\begin{figure*}
  \begin{center}
    \centerline{\includegraphics[width=1\linewidth]{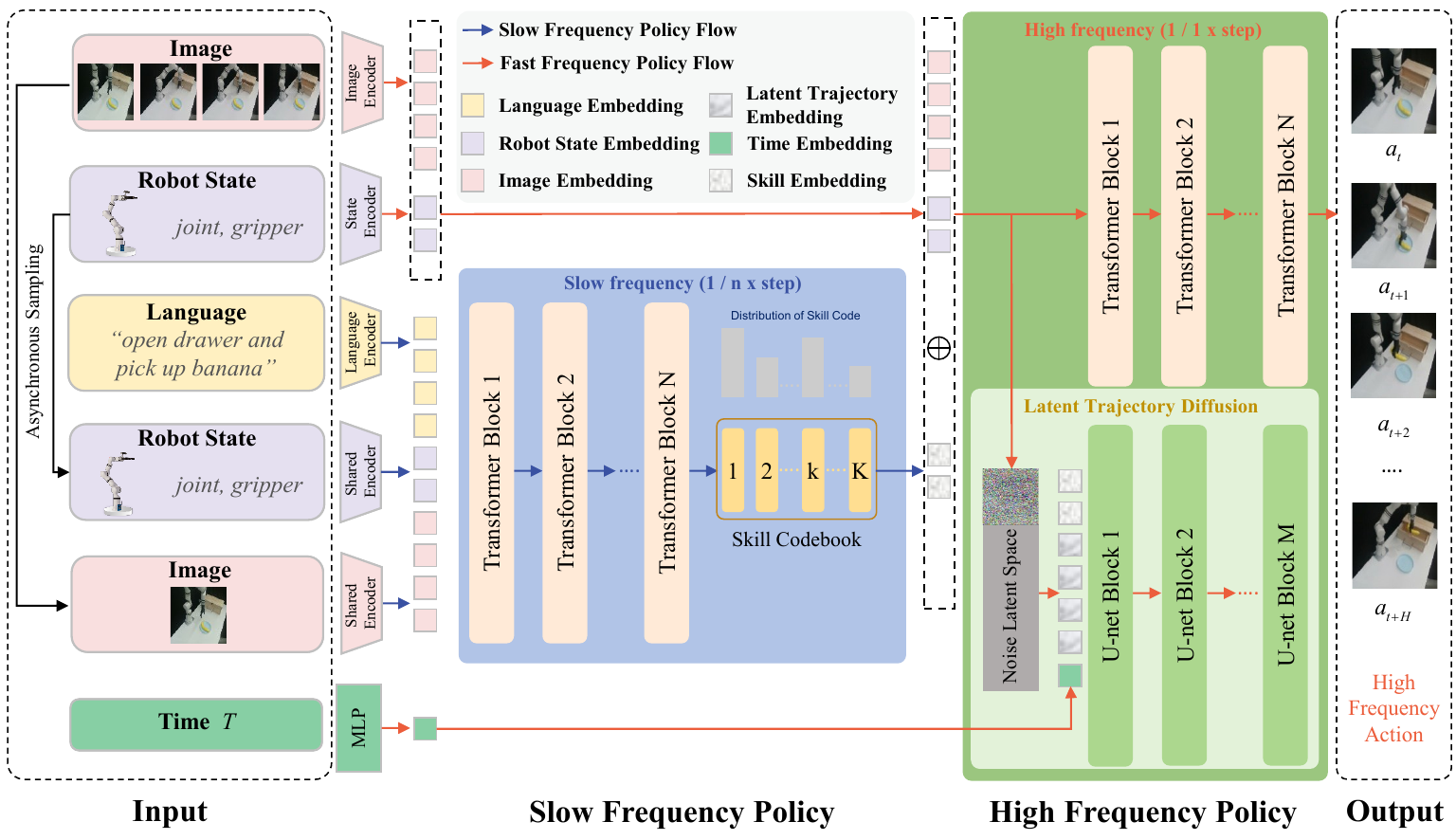}}
    \caption{
      Overview of DASL, an asynchronous hierarchical imitation learning framework. The high-level policy operates at a slow timescale to generate discrete semantic skills from language instructions and sparse observations, while the low-level policy executes skill-conditioned actions at a fast timescale. A latent diffusion module regularizes the latent trajectory space during training and is removed at inference for efficient real-time control.
    }
    \label{DASL_overview}
  \end{center}
    \vspace{-3em}
\end{figure*}

We propose DASL, an asynchronous hierarchical imitation learning framework for multi-task robotic manipulation combining language-conditioned skill discovery with latent diffusion models. As shown in Figure \ref{DASL_overview}, DASL decouples high-level semantic planning from low-level action execution via distinct temporal resolutions. A slow-timescale high-level policy deliberatively infers discrete semantic skills from language instructions, while a fast-timescale low-level policy executes skill-conditioned actions. This asynchronous design stabilizes skill learning and preserves the high-frequency control essential for precise manipulation.

\subsection{Problem Setup}

We model multi-task imitation learning as a Task-Augmented Markov Decision Process (MDP) \cite{garg2022lisa} with tasks specified by natural language instructions $l \in \mathcal{L}$. Given an offline dataset of language--trajectory pairs $\mathcal{D} = \{ (l^i, o_1^i, a_1^i, \dots, o_{T_i}^i, a_{T_i}^i) \}_{i=1}^N$, we aim to learn a language-conditioned sequential policy $\pi(a_t \mid o_{\le t}, a_{<t}, l)$ that generalizes to unseen task and environment. Furthermore, the lack of sub-task annotations necessitates the unsupervised discovery of reusable skill primitives.

\subsection{High-Level Semantic Skill Planning}

Within our dual-process architecture, the high-level policy governs language-guided skill planning. Operating on a slower timescale than the low-level controller, its core objective is to achieve temporal abstraction by mapping multimodal observations into a sequence of discrete behavioral primitives.

\textbf{Slow-Frequency Timescale.} To enable temporal abstraction and dynamically align with the low-level Decision Transformer, the high-level policy processes a sliding context window of length $L$ at a coarser fixed interval $I$. Formally, for a given sliding window, the high-level policy operates on a 1-based strided timeline, denoted as $\mathcal{T}_{\mathrm{slow}} = \{ 1 + k I \mid k = 0, 1, \ldots, \bigl\lfloor (L-1)/I \bigr\rfloor \}$. At each slow timestep $t_s \in \mathcal{T}_{\mathrm{slow}}$, the policy predicts a continuous latent skill $\bar{z}_{t_s}$ conditioned on the instruction and visual observations up to $t_s$.

\textbf{Skill Predictor.} We employ an Option Transformer for cross-modal reasoning. First, observations at each slow timestep are encoded via a visual encoder $\phi$ to form the sequence $\mathbf{O} = [\phi(o_{t_s})]_{t_s \in \mathcal{T}_{\mathrm{slow}}} \in \mathbb{R}^{|\mathcal{T}_{\mathrm{slow}}| \times d}$. This is temporally concatenated with embedded instructions $\mathbf{E}^{\mathrm{lang}} \in \mathbb{R}^{L_{\mathrm{lang}} \times d}$ to form the joint sequence $\mathbf{S} = [\mathbf{E}^{\mathrm{lang}}; \mathbf{O}]$. Processing $\mathbf{S}$ through causal self-attention allows observation tokens to attend to the global semantic context. We then extract the hidden states $\mathbf{h}_{t_s}$ exclusively from the observation positions and project them via a linear head to yield continuous latent skills $\bar{z}_{t_s} = \mathbf{W}_{\mathrm{skill}} \mathbf{h}_{t_s} + b_{\mathrm{skill}}$. Stacking these yields the continuous skill matrix $\bar{\mathbf{Z}} \in \mathbb{R}^{|\mathcal{T}_{\mathrm{slow}}| \times d_{\mathrm{option}}}$.

\textbf{Discrete Skill Vector Quantization.} To obtain a discrete skill library, we map each continuous pre-activation $\bar{z}_{t_s}$ to a learnable codebook $\mathcal{C} = \{c_m\}_{m=1}^M \subset \mathbb{R}^{d_{\mathrm{option}}}$ via a VQ bottleneck. The discrete skill $z_{t_s}$ is retrieved via nearest-neighbor lookup:
\begin{equation}
    \label{eq:vq_lookup}
    z_{t_s} = \operatorname{VQ}(\bar{z}_{t_s}) = \operatorname*{argmin}_{c \in \mathcal{C}} \|\bar{z}_{t_s} - c\|_2,
\end{equation}
Applying this quantization row-wise to $\bar{\mathbf{Z}}$ yields the discrete skill matrix $\mathbf{Z} \in \mathbb{R}^{|\mathcal{T}_{\mathrm{slow}}| \times d_{\mathrm{option}}}$, serving as the high-level temporal guidance for the low-level controller. During training, we optimize the codebook using an auxiliary commitment loss and approximate gradients through the quantization step via the Straight-Through Estimator (STE).

\subsection{Low-Level Skill-Conditioned Decision Transformer}

Operating at a high frequency, the low-level policy synthesizes actions conditioned on observations and skills. We introduce a hybrid architecture coupling a Decision Transformer for real-time inference with a latent diffusion module employed solely during training. This design uses diffusion to structure the latent manifold, ensuring semantic alignment between skills and trajectories without compromising the efficiency of the transformer backbone.

\textbf{Skill Fusion.} We condition the low-level policy by fusing discrete skills with high-frequency observations. Specifically, the retrieved skill sequence is projected into embeddings $e_{t_s}^{\mathrm{skill}}$ and upsampled by repeating each vector $I$ times. The resulting sequence is truncated or padded to match the exact trajectory sequence length $L$. This yields the aligned high-frequency skill matrix, which is added element-wise to the observation embeddings $\mathbf{E}^{\mathrm{obs}}$, incorporating semantic guidance without increasing the feature dimensionality:
\begin{equation}
    \label{eq:skill_fusion}
    \mathbf{E}^{\mathrm{fused}} = \mathbf{E}^{\mathrm{obs}} + \big[ \underbrace{e_{t_s^{(1)}}^{\mathrm{skill}}, \dots, e_{t_s^{(1)}}^{\mathrm{skill}}}_{I \text{ times}}, \dots, \underbrace{e_{t_s^{(|\mathcal{T}_{\mathrm{slow}}|)}}^{\mathrm{skill}}, \dots, e_{t_s^{(|\mathcal{T}_{\mathrm{slow}}|)}}^{\mathrm{skill}}}_{I \text{ times}} \big]^\top.
\end{equation}

\textbf{Latent Trajectory Diffusion Module.} To robustly regularize the latent dynamics, we apply an auxiliary diffusion process. Let $\tau_0 = \mathbf{E}_{1:H}^{\mathrm{fused}}$ denote the pre-timestep-embedded trajectory prefix of length $H = \min(\text{horizon}, L)$. The forward process adds noise at step $n$ via a cosine schedule $\bar{\alpha}_n$:
\begin{equation}
    \label{eq:forward_diff}
    \tau_n = \sqrt{\bar{\alpha}_n}\tau_0 + \sqrt{1 - \bar{\alpha}_n}\epsilon, \quad \epsilon \sim \mathcal{N}(\mathbf{0}, \mathbf{I}).
\end{equation}
For the reverse process, a 1D U-Net $\epsilon_\theta$ is trained to predict this injected noise, conditioned on the skill sequence and the diffusion step $n$. Crucially, this module serves solely as a representation regularizer during training, we do not perform iterative denoising sampling during rollout.

\textbf{Decision Transformer.} Using a GPT-2 backbone, the DT interleaves timestep-encoded action and fused observation embeddings $\mathbf{E}^{\mathrm{fused}}$ to predict the action $\hat{a}_t$ at each timestep. During deployment, the policy executes the action predicted at the final position of the sliding context window $L$.

\subsection{Training Objectives}
\label{sec:loss}

To achieve stable end-to-end learning from high-level language instructions to low-level continuous actions, DASL employs joint optimization. To mitigate codebook collapse and visual overfitting common in synchronous hierarchical training \cite{garg2022lisa}, our objective decouples slow semantic reasoning from real-time motion control. Concurrently, latent diffusion disperses skill representations across the data manifold, maximizing codebook utilization and preventing index collapse.

\textbf{Action Prediction Loss.} The low-level Decision Transformer performs behavioral cloning to reproduce expert actions given observations and target skills. The loss over a context window of length $L$ is the mean squared error between predicted actions $\hat{a}_t$ and ground-truth $a_t^{\mathrm{expert}}$:
\begin{equation}
    \label{eq:action_loss}
    \mathcal{L}_{\mathrm{action}} = \frac{1}{L} \sum_{t=1}^{L} \| \hat{a}_t - a_t^{\mathrm{expert}} \|_2^2.
\end{equation}
In standard synchronous architectures, backpropagating high-frequency action loss directly into a slow-frequency latent space often corrupts high-level semantics with execution noise. To mitigate this timescale discrepancy, DASL accumulates gradients across $I$ execution steps, effectively suppressing noise prior to semantic updates.

\textbf{VQ Commitment Loss.} To stabilize discrete skill representations, we optimize the Option Transformer's continuous outputs $\bar{z}_{t_s}$ and codebook $\mathcal{C} = \{c_m\}_{m=1}^M$ via the VQ objective:
\begin{equation}
    \label{eq:vq_loss}
    \mathcal{L}_{\mathrm{vq}} = \|\bar{z}_{t_s} - \operatorname{sg}[z_{t_s}]\|_2^2 + \beta \|\operatorname{sg}[\bar{z}_{t_s}] - z_{t_s}\|_2^2,
\end{equation}
where $z_{t_s}$ is the retrieved code, $\beta$ is the commitment weight, and $\operatorname{sg}[\cdot]$ denotes the stop-gradient operator. This objective induces latent skills to cluster around discrete, reusable centroids, promoting structured and compositional high-level policies.

\textbf{Latent Diffusion Loss.} Prior methods \cite{garg2022lisa} typically rely on retrospective sequence modeling, a paradigm that often overfits to historical visual observations. To overcome this limitation, we introduce an auxiliary latent diffusion loss $\mathcal{L}_{\mathrm{diff}}$ to explicitly model the trajectory manifold:
\begin{equation}
      \label{eq:diff_loss}
    \mathcal{L}_{\mathrm{diff}} = \mathbb{E}_{\tau_0, n, \epsilon \sim \mathcal{N}(\mathbf{0}, \mathbf{I})} \left[ \big\| \epsilon - \epsilon_\theta( \underbrace{\sqrt{\bar{\alpha}_n}\tau_0 + \sqrt{1-\bar{\alpha}_n}\epsilon}_{\tau_n}, n, c ) \big\|_2^2 \right].
\end{equation}
Combining the inverse denoising process with skill sequence $c$, this objective explicitly links discrete skills to the future skill trajectory distribution. This disperses potential skills across the data manifold, forcing the model to use distinct codebooks for different action behaviors, avoiding index collapse.

\textbf{Total Training Objective.} This decoupled formulation allows DASL to jointly optimize the asynchronous modules while learning stable behavioral semantics:
\begin{equation}
    \label{eq:total_loss}
    \mathcal{L}_{\mathrm{total}} = \mathcal{L}_{\mathrm{action}} + \lambda_{\mathrm{vq}}\mathcal{L}_{\mathrm{vq}} + \lambda_{\mathrm{diff}}\mathcal{L}_{\mathrm{diff}},
\end{equation}
where $\lambda_{\mathrm{vq}}$ and $\lambda_{\mathrm{diff}}$ are hyperparameters. This multi-objective loss function enables the model to learn a complete mapping from language instructions to action sequences. The complete procedure is detailed in Appendix~\ref{app:pseudo_code}. Details regarding the network architecture are provided in Appendix~\ref{app:Network_Architecture}.

\begin{table*}[tb]
    \centering
    \small
    \caption{\textbf{Rephrasal-wise success rates (\%) on LOReL Sawyer.} We report success rates (\%) of different algorithms under both state and image observation settings. DASL outperforms all other methods across almost all rephrasal types. Best results are \textbf{bolded}, second-best are \underline{underlined}, and {\color{blue}blue} values indicate results cited from original papers. All other results are averaged over 3 seeds.}

    \newcommand{\hpad}{\hspace{2pt}} 

    \newcommand{\res}[3][]{%
        \def\corecontent{%
            \makebox[2.0em][r]{#2} %
            \scriptsize{\raisebox{.5pt}{$\pm$}} %
            \makebox[1.6em][l]{#3}%
        }%
        \ifx\relax#1\relax%
            \hpad\corecontent\hpad%
        \else%
            #1{\hpad\corecontent\hpad}%
        \fi%
    }
    \resizebox{\textwidth}{!}{%
        \begin{tabular}{@{}l c c c c c c c@{}} %
        \toprule
        \textbf{Rephrasal Type} & \textbf{Obs} & \textbf{Lang DT} & \textbf{LISA} & \textbf{LCSD} & \textbf{SkillDiffuser} & \textbf{LADS} & \textbf{DASL(ours)} \\
        \midrule
        \multirow{2}{*}{seen} & State & \res{49.67}{5.77} & \res{12.78}{9.02} & \res[\underline]{60.20}{\textcolor{blue}{0.00}} & \res{38.89}{14.30} & \res{51.56}{1.89} & \res[\textbf]{62.22}{3.85}\\
         & Image & \res{15.00}{\textcolor{blue}{0.00}} & \res{40.00}{\textcolor{blue}{0.00}} & \res[\underline]{50.80}{\textcolor{blue}{0.00}} & \res{43.65}{4.70} & \res{40.20}{4.40} & \res[\textbf]{53.33}{5.36}\\
        \multirow{2}{*}{unseen noun} & State & \res{35.44}{0.53} & \res{10.00}{6.70} & \res{50.20}{\textcolor{blue}{0.00}} & \res{36.67}{3.34} & \res[\underline]{50.56}{11.33} & \res[\textbf]{61.11}{7.03}\\
         & Image & \res{13.33}{\textcolor{blue}{0.00}} & \res{33.33}{\textcolor{blue}{0.00}} & \res{30.40}{\textcolor{blue}{0.00}} & \res[\underline]{36.01}{6.30} & \res{35.10}{6.50} & \res[\textbf]{50.11}{5.47}\\
        \multirow{2}{*}{unseen verb} & State & \res{31.56}{5.85} & \res{11.22}{9.88} & \res{29.30}{\textcolor{blue}{0.00}} & \res{31.11}{10.75} & \res[\underline]{44.56}{1.37} & \res[\textbf]{50.00}{10.00}\\
         & Image & \res{28.33}{\textcolor{blue}{0.00}} & \res{30.00}{\textcolor{blue}{0.00}} & \res{27.50}{\textcolor{blue}{0.00}} & \res[\underline]{36.70}{9.50} & \res{31.10}{0.50} & \res[\textbf]{58.67}{9.53}\\
        \multirow{2}{*}{unseen noun\&verb} & State & \res{17.33}{9.19} & \res{11.00}{7.20} & \res{28.70}{\textcolor{blue}{0.00}} & \res{26.67}{3.34} & \res[\underline]{34.44}{5.19} & \res[\textbf]{45.56}{5.08}\\
         & Image & \res{6.70}{\textcolor{blue}{0.00}} & \res{20.00}{\textcolor{blue}{0.00}} & \res{36.50}{\textcolor{blue}{0.00}} & \res[\textbf]{42.02}{3.80} & \res{21.70}{1.20} & \res[\underline]{39.55}{0.39}\\
        \multirow{2}{*}{human provided} & State & \res{34.35}{7.47} & \res{10.42}{7.23} & \res{35.50}{\textcolor{blue}{0.00}} & \res{35.47}{3.93} & \res[\underline]{50.06}{4.75} & \res[\textbf]{52.20}{0.77}\\
         & Image & \res{26.98}{\textcolor{blue}{0.00}} & \res{27.35}{\textcolor{blue}{0.00}} & \res{33.40}{\textcolor{blue}{0.00}} & \res[\underline]{40.16}{2.10} & \res{33.90}{1.30} & \res[\textbf]{49.01}{2.10}\\
          \midrule
        \multirow{2}{*}{\textbf{Average}} & State & \res{33.67}{5.41} & \res{11.08}{7.98}\textsuperscript{*} & \res{40.78}{\textcolor{blue}{0.00}} & \res{33.76}{4.83} & \res[\underline]{46.23}{3.95} & \res[\textbf]{54.22}{4.65}\\
         & Image & \res{18.07}{\textcolor{blue}{0.00}} & \res{30.14}{\textcolor{blue}{0.00}} & \res{35.70}{\textcolor{blue}{0.00}} & \res[\underline]{39.71}{\textcolor{blue}{0.00}} & \res{32.40}{1.40} & \res[\textbf]{50.13}{0.43}\\
        \bottomrule
        \end{tabular}%
    }
    \label{tab:lorel_res_final_long}
    \vspace{-1.0em}
\end{table*}

\begin{figure}[tb] %

    \vspace{0.5em} 
    
    \centering
    \renewcommand{\arraystretch}{1.1} 

    \begin{minipage}[t]{0.56\linewidth}
        \vspace{0pt} %
        \centering
        \small
        \captionof{table}{N-rates of different methods on seen and unseen instructions in Kitchen with state and image observation.}
        \label{tab:kitchen_N_rate}

        \vspace{0pt} 
        {%
        \setlength{\tabcolsep}{2pt} 
        \newcommand{\hpad}{\hspace{1pt}} 

        \newcommand{\res}[3][]{%
            \def\corecontent{%
                \makebox[1.6em][r]{#2}%
                \scriptsize{\raisebox{.5pt}{$\pm$}}%
                \makebox[1.6em][l]{#3}%
            }%
            \ifx\relax#1\relax%
                \hpad\corecontent\hpad%
            \else%
                #1{\hpad\corecontent\hpad}%
            \fi%
        }

        \resizebox{\linewidth}{!}{%
        \begin{tabular}{@{}l c c c c c@{}}
            \toprule
            \textbf{Rephrasal Type} & \textbf{Obs} & \textbf{Lang DT} & \textbf{LISA} & \textbf{LADS} & \textbf{DASL(ours)} \\
            \midrule
            \multirow{2}{*}{seen}   & State & \res{1.39}{0.12} & \res{0.71}{0.44} & \res[\textbf]{2.25}{0.14} & \res{1.66}{0.05}\\
                                    & Image & \res{1.32}{0.41} & \res{0.93}{0.12} & \res{1.61}{0.28} & \res[\textbf]{1.74}{0.09}\\
            \multirow{2}{*}{unseen} & State & \res{1.27}{0.17} & \res{0.71}{0.53} & \res[\textbf]{2.23}{0.16} & \res{1.76}{0.14}\\
                                    & Image & \res{1.07}{0.32} & \res{1.29}{0.20} & \res{1.44}{0.32} & \res[\textbf]{1.71}{0.17}\\
            \bottomrule
        \end{tabular}%
        }%
        }%
    \end{minipage}\hfill %
    \begin{minipage}[t]{0.41\linewidth}
        \vspace{-5pt} %
        \centering
        \includegraphics[width=\linewidth, height=3.0cm, keepaspectratio]{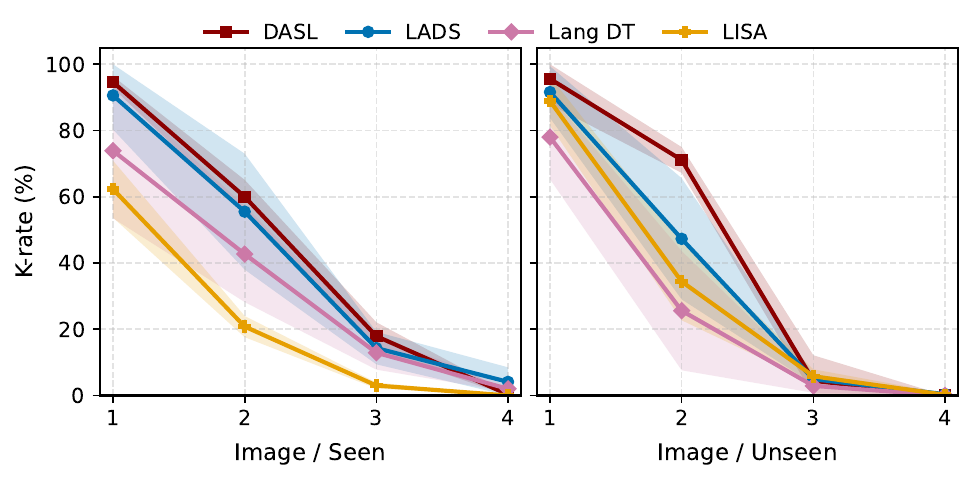}
        
        \vspace{-5pt} 
        \captionof{figure}{K-rates on seen and unseen tasks in Kitchen (image).}
        \label{fig:kitchen_image_k_rate}
    \end{minipage}

    \vspace{-1.5em} 
\end{figure}

\section{Experiments}
\label{Experiments}
In this section, we evaluate DASL on robotic manipulation and grid-world navigation tasks, comparing it against strong hierarchical and non-hierarchical baselines. We further analyze the generalization of the framework to compositional instructions, the contributions of the architectural components, the efficiency of inference, and the practical viability of the approach through a real-world experiment.

\textbf{Environments.} We evaluate the proposed method across simulated benchmarks, including robotic manipulation tasks (LOReL Sawyer \cite{nair2022learning}, Franka Kitchen \citep{gupta2019relay}, and CALVIN \citep{mees2022calvin}) and grid-world navigation tasks (BabyAI \citep{chevalier2018babyai}). Beyond simulations, we validate DASL in a real-world experiment to demonstrate its effectiveness. Please refer to Appendix~\ref{app:datasets_descriptions} for dataset specifications and Appendix~\ref{app_Experiment} for the real-world experiment setup.

\textbf{Baselines.} We compare DASL with a diverse set of baseline methods, covering standard behavior cloning approaches and state-of-the-art hierarchical learning algorithms, including LCBC \cite{stepputtis2020language}, LangDT \cite{garg2022lisa}, LISA \cite{garg2022lisa}, HULC \cite{mees2022calvin}, 
LCSD \cite{ju2024rethinking} , SkillDiffuser \cite{liang2024skilldiffuser}, and LADS \cite{jiang2025discrete}. Detailed descriptions of these baselines are provided in Appendix~\ref{app:baselines_descriptions}.

\subsection{Does DASL achieve superior performance in skill acquisition compared to state-of-the-art baselines?} 
To rigorously validate DASL's skill learning capabilities, we conduct comparative evaluations across three simulation benchmarks. Ensuring fair comparison, we align experimental settings with baselines; specifically, DASL employs the same vision and language encoder architectures as LISA. Reported success rates are directly sourced from original publications or derived from evaluating trained models over 50 episodes. Appendix~\ref{app:Hyperparameters} provides detailed configurations. Additional heatmap visualizations and word clouds are provided in Appendix~\ref{app:skill_visuals}.

\textbf{Performance on the LOReL Sawyer dataset.} We evaluate performance across both state and image observation modalities. Table~\ref{tab:lorel_res_final_long} summarizes quantitative rewriting task results; see Appendices~\ref{app:lorel_descriptions}, \ref{app:lorel_result}, and \ref{app:Qualitative_Results_in_LOReL} for detailed operational instructions and further results. DASL outperforms all baselines, demonstrating exceptional cross-task adaptability. Notably, LISA suffers from codebook collapse on LOReL (state) (see Figure~\ref{fig:codebook_collapse}, indicated by an asterisk (*)), performing worse than non-hierarchical architectures like Lang DT. This aligns with observations in LCSD \cite{ju2024rethinking} and LADS \cite{jiang2025discrete}. Compared to other Decision Transformer-based methods (e.g., Lang DT and LISA), our results underscore the necessity of modeling decision-making within a latent vector space. This approach effectively mitigates inherent skill learning instability and resolves LISA's index collapse. Compared to SkillDiffuser, we validate that skill-conditioned diffusion models effectively learn to denoise latent vectors, enhancing alignment between skills and trajectories. Finally, compared to LADS, we demonstrate that a hierarchical architecture with asynchronous synergy and heterogeneous modality alignment facilitates unified skill representations, enabling rapid and precise manipulation.

\textbf{Performance on Franka Kitchen.} We evaluate DASL on the Franka Kitchen benchmark using two metrics: (1) \textit{N-rate}, representing the average number of successfully completed sub-tasks out of four, and (2) \textit{K-rate}, measuring the success rate of completing at least $K$ out of four sub-tasks. Results are summarized in Table~\ref{tab:kitchen_N_rate} and Figure~\ref{fig:kitchen_image_k_rate}. For detailed operational instructions and additional results, see Appendix~\ref{app:Franka-Kitchen-descriptions} and Appendix~\ref{app:Kitchen_result}. DASL demonstrates exceptional robustness in tasks with image-based observations and unseen instructions. Comparisons with LISA reveal that the tight coupling between high-level reasoning and low-level control during synchronous training leads to codebook collapse. Consequently, LISA's performance in the seen-image setting ($0.93$) falls below the non-hierarchical Lang DT baseline ($1.32$). In contrast, DASL circumvents this pitfall through its asynchronous hierarchical architecture, ensuring the semantic consistency of learned skills. Although LADS achieves optimal performance in state-based observations by leveraging coordinate information within a discrete latent action space, its performance degrades significantly in more practical image-based settings. This suggests that visual representation noise inherent in synchronous end-to-end training severely destabilizes high-level planning. Conversely, DASL’s latent diffusion effectively models skill trajectory distributions, maintaining a high N-rate ($1.71$) on unseen image-based tasks.

\begin{figure*}[t!]
    \centering 

    \begin{minipage}[t]{0.62\textwidth} 
        \vspace{0pt} 
        
        \newcommand{\res}[3][]{%
            \def\corecontent{\makebox[2.1em][r]{#2}\hspace{1pt}\scriptsize{\raisebox{.5pt}{$\pm$}}\hspace{1pt}\makebox[1.5em][l]{#3}}%
            \ifx\relax#1\relax\corecontent\else#1{\corecontent}\fi%
        }
        
        \small 

        \captionof{table}{Performance on CALVIN benchmark.}
        \label{tab:calvin_results}
        \vspace{2pt} %
        \resizebox{\linewidth}{!}{
        \setlength{\tabcolsep}{3pt} 
        \begin{tabular}{l ccccc}
            \toprule
            \textbf{Method} & \textbf{Lang DT} & \textbf{LISA} & \textbf{HULC} & \textbf{LCSD} & \textbf{DASL(ours)} \\
            \midrule
            SR & \res{11.70}{0.80} & \res{10.10}{3.30} & \res{32.50}{2.50} & \res[\underline]{35.60}{1.80} & \res[\textbf]{48.05}{1.27} \\
            \bottomrule
        \end{tabular}
        }
        
        \vspace{0.5em} %

        \captionof{table}{Performance on LOReL Compositional Tasks}
        \label{tab:lorel_compos_task_res}
        \vspace{4pt} %
        \resizebox{\linewidth}{!}{
        \setlength{\tabcolsep}{3pt} 
        \begin{tabular}{l c c c c c}
        \toprule
        \textbf{Obs} & \textbf{Lang DT} & \textbf{LISA} & \textbf{SkillDiffuser} & \textbf{LADS} & \textbf{DASL(ours)} \\
        \midrule
        State & \res{11.11}{1.78} & \res{0.17}{0.29} & \res{10.55}{2.52} & \res[\underline]{23.11}{2.76} & \res[\textbf]{23.61}{2.10}\\
        Image & \res{13.33}{1.30} & \res{20.89}{0.60} & \res[\underline]{25.21}{2.70} & \res{20.40}{1.80}   & \res[\textbf]{32.50}{3.99}\\
        \bottomrule
        \end{tabular}
        }
        
    \end{minipage}%
    \hfill 
    \begin{minipage}[t]{0.34\textwidth} 
        \vspace{0pt} 
        \centering 
        \includegraphics[width=\linewidth]{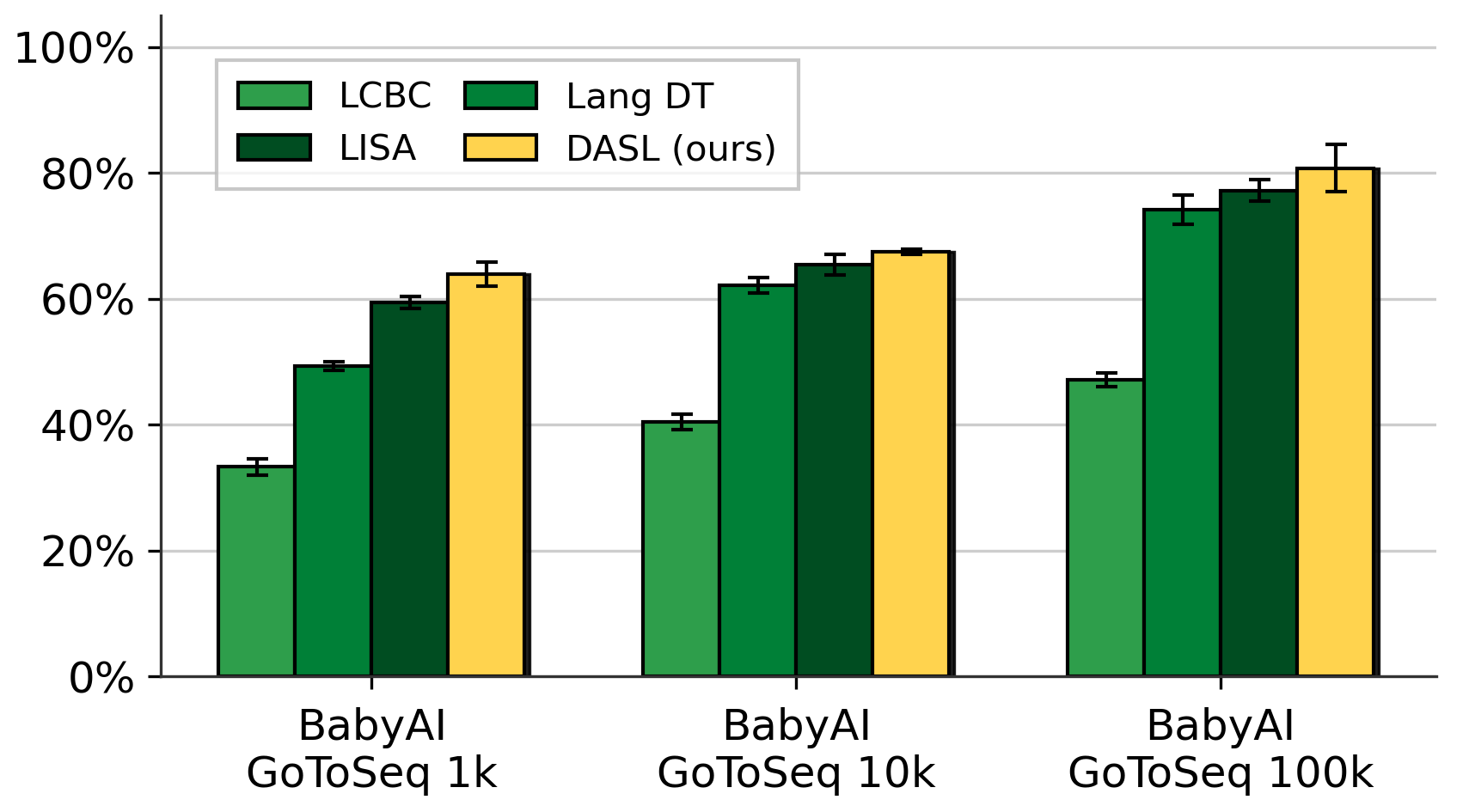} 
        \captionof{figure}{Success rates (\%) on BabyAI GoToSeq task with varying numbers of demonstrations.}
        \label{fig:babyai_gotoseq_bar}
    \end{minipage}
    \vspace{-2em} %
\end{figure*}

\textbf{Performance on CALVIN.} We directly select trajectories from the CALVIN-D dataset corresponding to six modified tasks (see Appendix~\ref{app:CALVIN-descriptions}). As shown in Table~\ref{tab:calvin_results}, DASL achieves a $48.05\%$ SOTA success rate, significantly outperforming the official HULC baseline ($32.50\%$) and LCSD ($35.60\%$). Following LCSD \cite{ju2024rethinking}, the setup employs a 21-dimensional perspective state and a simplified MLP visual encoder, isolating decision-making from visual representation bias. The $15.55\%$ improvement over HULC under this constrained data regime indicates DASL's asynchronous dual-process architecture handles low-dimensional state vectors more robustly than Lang DT ($11.70\%$) or synchronous LISA ($10.10\%$). These baselines likely suffer from codebook collapse and structural instability under sparse state inputs. Conversely, DASL's performance validates decoupling slow semantic planning from fast action execution, preventing low-level control constraints from disrupting high-level reasoning. Further results are provided in Appendix~\ref{app:CALVIN_result}.

\textbf{Performance on BabyAI.} As shown in Figure~\ref{fig:babyai_gotoseq_bar}, DASL consistently achieves superior success rates across all data regimes on the BabyAI GoToSeq task, outperforming three baselines. For detailed operational instructions, see Appendix~\ref{app:BabyAI-descriptions}. Notably, when trained on only 1k randomly sampled trajectories, DASL attains a $63.93\% \pm 1.91\%$ success rate, surpassing LISA ($59.3\% \pm 0.9\%$) by $4.63\%$. These results indicate our approach effectively extracts valuable information from limited data, demonstrating the robust skill learning capabilities of DASL in grid-world navigation tasks.

\subsection{How does DASL compare to baselines in terms of generalization capabilities when facing compositional tasks?
}

To evaluate DASL's generalization toward compositional language instructions, we assess its performance on the LOReL Sawyer dataset. Table~\ref{tab:lorel-composite-instrs} details these specifications. DASL recomposes learned atomic skills to adapt to novel instructions. As Table~\ref{tab:lorel_compos_task_res} shows, DASL outperforms competing methods, underscoring its capacity to master and combine atomic skills. Facilitated by the asynchronous hierarchical architecture and latent diffusion model, DASL exhibits robust language understanding and generalization, significantly surpassing baselines, especially in image-based settings.

\subsection{Are the proposed architectural components of DASL essential for its effectiveness?}
\label{sec:ablation}

To rigorously verify individual component contributions within DASL, we conducted ablation studies on the LOReL Sawyer dataset. Adopting LISA as our baseline, we progressively incorporated Latent Skill Embedding, Asynchronous Hierarchical Architecture, and the Skill-Conditioned Diffusion Mechanism. More ablation studies can be found in Appendix~\ref{app_Additional_Ablation_Studies}.

\begin{table}[hbtp]
    \vspace{-1.5em} %
    \centering
    \caption{Ablation study of individual component contributions  on the LOReL Sawyer dataset.}
    \label{tab:ablation_res}
    \vspace{4pt} %
    \small 
    \renewcommand{\arraystretch}{1.2}
    \resizebox{\linewidth}{!}{
    \begin{tabular}{l c c c c c c}
        \toprule
        \textbf{Method} & \textbf{Embedding} & \textbf{Synchronous} & \textbf{Asynchronous} & \textbf{Diffusion} & \textbf{State SR (\%)} & \textbf{Image SR (\%)} \\
        \midrule
        LISA                 & $\times$     & $\checkmark$ & $\times$     & $\times$     & 19.08 & 30.14 \\
        Variant A            & $\checkmark$ & $\checkmark$ & $\times$     & $\times$     & 28.34 & 32.26 \\
        Variant B            & $\checkmark$ & $\times$     & $\checkmark$ & $\times$     & 38.26 & 38.00 \\
        Variant C            & $\checkmark$ & $\checkmark$ & $\times$     & $\checkmark$ & 32.08 & 31.19 \\
        \midrule
        \textbf{DASL (ours)} & $\checkmark$ & $\times$     & $\checkmark$ & $\checkmark$ & \textbf{58.49} & \textbf{50.66} \\
        \bottomrule
    \end{tabular}
    }
    \vspace{-1em} %
\end{table}

\textbf{Effectiveness of Latent Skill Embedding.} To evaluate explicit skill representation, we compare the LISA baseline with variant A incorporating the embedding module. As Table~\ref{tab:ablation_res} shows, latent skill representation yields a substantial performance gain. This indicates raw observation spaces struggle to directly capture high-level semantic information. Introducing latent skill embeddings effectively aligns skills with trajectories within a structured latent space, achieving a unified and robust skill representation.

\textbf{Effectiveness of Asynchronous Decoupling.} Although effective skill representations improve performance, synchronous architectures risk codebook collapse. We therefore evaluate decoupling high-level decision-making and low-level control frequencies. As Table~\ref{tab:ablation_res} presents, the asynchronous variant B achieves substantial improvement over the synchronous baseline A equipped only with skill embeddings. Specifically, under state observations, the success rate rises from $28.34\%$ to $38.26\%$. This demonstrates that breaking temporal coupling mitigates interference between high-level reasoning and low-level action generation. Allowing the high-level policy to reason at a coarser resolution ($1/n \times \text{step}$) avoids instability triggered by forcing semantic updates at every high-frequency control step.

\textbf{Effectiveness of Latent Trajectory Diffusion.} While asynchronous execution enhances stability, linear low-level embedding struggles to capture complex compositional trajectory distributions. Table~\ref{tab:ablation_res} compares synchronous baseline A with variant C using latent diffusion. This indicates that applying diffusion without resolving frequency mismatches is insufficient for complex distributions, especially under high-frequency action noise. Conversely, integrating latent diffusion into asynchronous baseline B yields the DASL model with a profound performance leap, reaching $58.49\%$ and $50.66\%$ success rates in state and image modalities. This improvement exceeds the sum of individual gains from the asynchronous architecture and diffusion module. Data suggests a synergistic relationship where the asynchronous framework establishes a stable semantic latent space. The training-only diffusion mechanism forces fused state-skill embeddings onto the manifold of valid expert trajectories, mitigating inherent covariate shift in behavioral cloning. These results confirm synergy among latent skill embedding, asynchronous coordination, and diffusion-enhanced representation is crucial to DASL's superior performance.

\subsection{How does DASL perform in terms of inference frequency and resource consumption? }
\label{sec:resource}

Real-world experment requires not only high success rates but also low latency and efficient resource utilization. To answer the question, we compare the computational overhead of DASL against state-of-the-art baselines. As detailed in Table~\ref{tab:efficiency_comparison}. In terms of Inference Speed, DASL achieves an inference frequency of 13.61 Hz (State), which is comparable to the Decision Transformer-based LISA (15.68 Hz) and significantly faster than SkillDiffuser (2.02 Hz). This speed advantage stems from our design choice: unlike SkillDiffuser, which requires computationally expensive iterative denoising steps during inference, DASL utilizes the diffusion model solely as an auxiliary objective during training. At test time, the low-level policy operates as a fast, deterministic Decision Transformer.
Regarding Resource Utilization, DASL is highly efficient. With only 67.9M parameters and 768 MB of GPU memory usage, it consumes approximately 3x less memory than SkillDiffuser (2422 MB).
In conclusion, DASL achieves state-of-the-art task success rates without incurring the high computational costs usually associated with diffusion policies, making it highly suitable for real-time robotic control.

\begin{table}[hbtp]
    \centering
    \small 
    \vspace{-1em}
    \caption{Computational and inference efficiency on the LOReL Sawyer dataset.}
    \label{tab:efficiency_comparison}
    \vspace{4pt}

    \renewcommand{\arraystretch}{1.25} 

    \setlength{\tabcolsep}{5pt} 

    \begin{tabular}{@{} l cc@{\hspace{15pt}} cc@{\hspace{15pt}} cc@{\hspace{15pt}} cc @{}}
        \toprule
        \multirow{2}{*}{\textbf{Method}} & \multicolumn{2}{c}{\textbf{Parameters (M)}} & \multicolumn{2}{c}{\textbf{Memory (MB)}} & \multicolumn{2}{c}{\textbf{Inference (Hz)}} & \multicolumn{2}{c}{\textbf{Success Rate (\%)}} \\
        \cmidrule(lr){2-3} \cmidrule(lr){4-5} \cmidrule(lr){6-7} \cmidrule(lr){8-9}
        & \textbf{State} & \textbf{Image} & \textbf{State} & \textbf{Image} & \textbf{State} & \textbf{Image} & \textbf{State} & \textbf{Image} \\
        \midrule
        LISA                 & 67.00  & 73.90  & 764  & 798  & 15.68 & 15.01 & 19.08 & 30.14 \\
        SkillDiffuser        & 186.62 & 190.08 & 2390 & 2422 & 2.32  & 2.02  & 33.77 & 39.71 \\
        LADS                 & 67.00  & 141.30  & 738  & 1028 & 12.44 & 13.03 & 49.91 & 33.30 \\
        \midrule
        \textbf{DASL (ours)} & \textbf{67.90} & \textbf{74.80} & \textbf{768} & \textbf{804} & \textbf{13.61} & \textbf{13.97} & \textbf{58.49} & \textbf{50.66} \\
        \bottomrule
    \end{tabular}
    \vspace{-1em}
\end{table}

\subsection{How does DASL perform compared to baselines in real-world experiment?}

\begin{figure*}[hbtp]
  \centering
  \vspace{-1em}
  \begin{subfigure}[b]{0.3\textwidth}
    \centering
    \includegraphics[width=\linewidth]{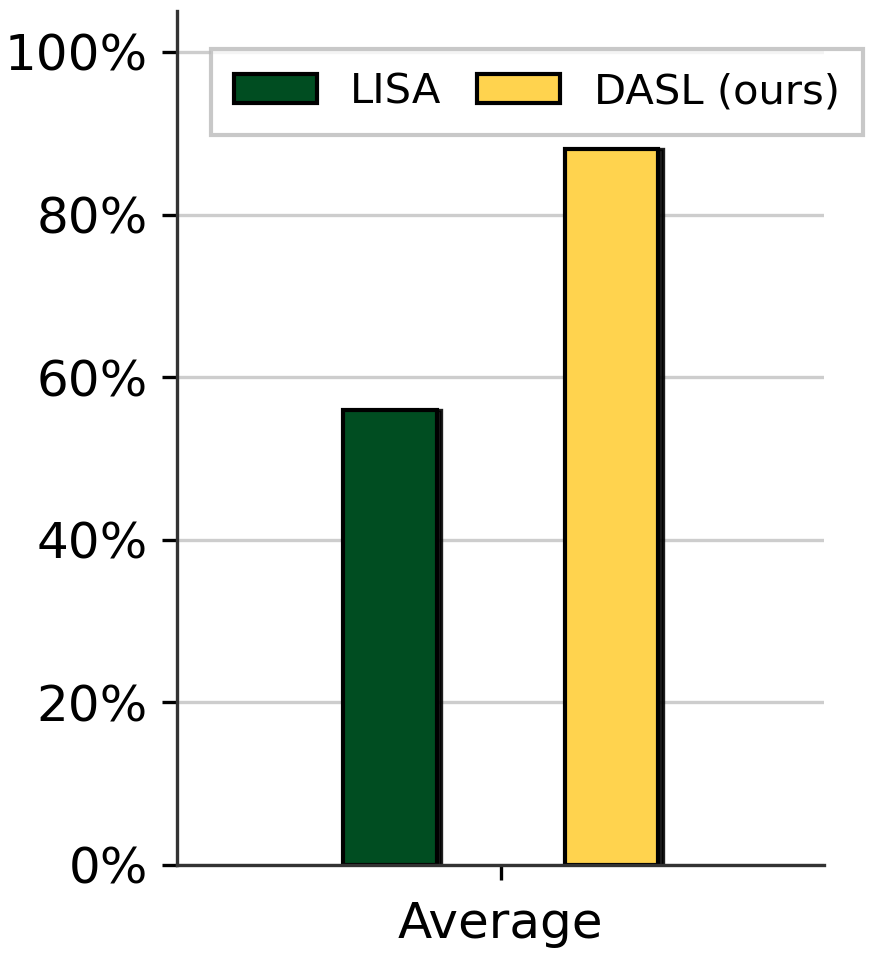}
    \caption{Average success rate}
    \label{fig:real_world_avg_bar}
  \end{subfigure}
  \hfill
  \begin{minipage}[b]{0.68\textwidth}
    \centering

    \begin{subfigure}{\linewidth}
      \centering
      \includegraphics[width=0.19\linewidth]{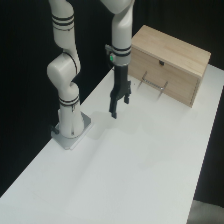}\hfill
      \includegraphics[width=0.19\linewidth]{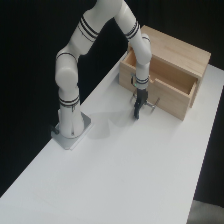}\hfill
      \includegraphics[width=0.19\linewidth]{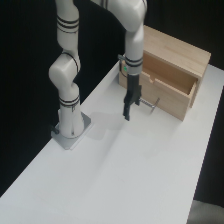}\hfill
      \includegraphics[width=0.19\linewidth]{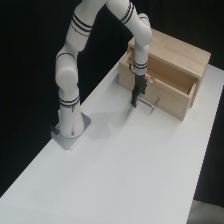}\hfill
      \includegraphics[width=0.19\linewidth]{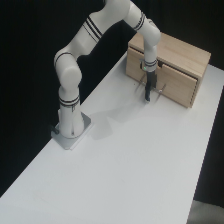}
      \caption{Compositional Task: open drawer and close drawer}
      \vspace{6pt}
    \end{subfigure}

    \begin{subfigure}{\linewidth}
      \centering
      \includegraphics[width=0.19\linewidth]{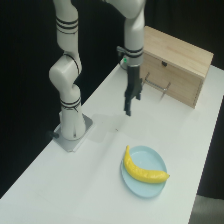}\hfill
      \includegraphics[width=0.19\linewidth]{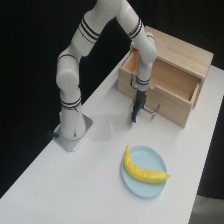}\hfill
      \includegraphics[width=0.19\linewidth]{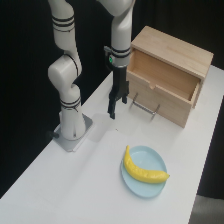}\hfill
      \includegraphics[width=0.19\linewidth]{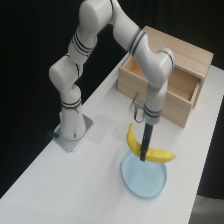}\hfill
      \includegraphics[width=0.19\linewidth]{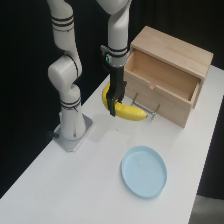}
      \caption{Compositional Task: open drawer and pick up banana from the plate}
    \end{subfigure}
  \end{minipage}

  \caption{Real-world quantitative and qualitative results for DASL. (a) compares average success rates. (b)-(c) display successful key-frame sequences for complex compositional tasks.}
  \label{fig:realman_combined_results}
  \vspace{-1em}
\end{figure*}

To evaluate the practical viability of DASL, we deploy it with a single-arm Realman manipulator and a fixed RealSense camera, as detailed in Appendix~\ref{app_Experiment}. We design five tasks: three atomic tasks (opening a drawer, picking a banana from a plate, and closing a drawer) using 60, 40, and 20 trajectories, respectively, alongside two compositional tasks (opening then closing a drawer, and opening a drawer before picking a banana) with 10 trajectories each. This deliberately imbalanced distribution assesses the capacity of the model to generalize to complex instructions. Models are trained using combined state and image inputs, with 10 test trials per category. As Figure~\ref{fig:realman_combined_results} demonstrates, DASL effectively executes both atomic and complex sequential instructions. Furthermore, under identical conditions, it significantly outperforms the LISA baseline, highlighting the capability of the asynchronous architecture to generalize despite imbalanced training demonstrations.

\section{Conclusion}
\label{sec:conclusion}
In this paper, we propose DASL, an asynchronous hierarchical imitation learning framework that addresses training instability and skill collapse in compositional, language-conditioned tasks. Inspired by the duality of System~1 and System~2 cognition, DASL decouples high-level semantic reasoning from low-level action execution by combining a slow-frequency skill planner with a high-frequency, diffusion-enhanced policy. Evaluations across both simulated benchmarks and real-world experment demonstrate that DASL outperforms state-of-the-art methods, particularly in compositional generalization and robustness to unseen instructions.

\textbf{Limitations and Future Work.} DASL currently relies on offline demonstrations, limiting its ability to adapt through online interaction. Future work will explore online fine-tuning via reinforcement learning and the integration of large-scale vision-language foundation models to support more open-ended instructions and diverse manipulation scenarios.

\newpage
\bibliographystyle{plainnat}
\bibliography{arxiv_2026}

\clearpage
\appendix
\section*{Appendix}
\section{Pseudo-code of Training DASL}
\label{app:pseudo_code}
Algorithm~\ref{alg:dasl_train} illustrates the training pseudocode of the DASL algorithm, which describes the main training procedure and the essential steps of the proposed method.

\begin{algorithm}[H]
\caption{Training Procedure of DASL (Dual-Process Atomic Skill Learning)}
\label{alg:dasl_train}
\small
\begin{algorithmic}[1]
\Require Expert dataset $\mathcal{D}=\{(l, \mathbf{o}, \mathbf{a}, \mathbf{t}, \mathbf{m})\}$, where $l$ is language, $\mathbf{o}$ observations, $\mathbf{a}$ actions, $\mathbf{t}$ timesteps, and $\mathbf{m}$ binary padding masks; batch size $B$; slow-frequency interval $I$; diffusion horizon $H$; diffusion steps $T_d$; codebook size $M$; weights $\lambda_{\mathrm{vq}}, \lambda_{\mathrm{diff}}$.
\Statex \textbf{Initialize:} Option Transformer $f_{\mathrm{slow}}$, Codebook $\mathcal{C}$, Decision Transformer $\pi_{\mathrm{fast}}$, Latent Diffusion $\epsilon_\theta$.
\State Set AdamW optimizer, learning rate scheduler, and gradient clipping threshold.

\While{not converged}
    \State Sample minibatch $\{(l^i, \mathbf{o}^i, \mathbf{a}^i, \mathbf{t}^i, \mathbf{m}^i)\}_{i=1}^{B} \sim \mathcal{D}$
    \State $\mathbf{E}^{\mathrm{lang}} \gets \phi_{\mathrm{lang}}(l)$ \Comment{Extract language embeddings for goal conditioning}

    \Statex \vspace{-0.2em}
    \Statex \hspace{-1em} \textit{\# 1. High-Level Semantic Skill Planning}
    \State $\mathbf{o}^{\mathrm{slow}}, \mathbf{t}^{\mathrm{slow}}, \mathbf{m}^{\mathrm{slow}} \gets \mathbf{o}_{::I},\ \mathbf{t}_{::I},\ \mathbf{m}_{::I}$ \Comment{Subsample at interval $I$}
    \State $\bar{\mathbf{Z}} \gets f_{\mathrm{slow}}(\mathbf{E}^{\mathrm{lang}}, \mathbf{o}^{\mathrm{slow}}, \mathbf{t}^{\mathrm{slow}}, \mathbf{m}^{\mathrm{slow}})$ \Comment{Predict skills conditioned on language}
    \State $\mathbf{Z}, \mathcal{L}_{\mathrm{vq}} \gets \operatorname{VQ}(\bar{\mathbf{Z}}; \mathcal{C})$ \Comment{Quantize skills \& compute commitment loss}

    \Statex \vspace{-0.2em}
    \Statex \hspace{-1em} \textit{\# 2. Low-Level Skill-Conditioned Decision Transformer}
    \State $\mathbf{E}^{\mathrm{obs}} \gets e_o(\mathbf{o})$ \textbf{and} $\mathbf{E}^{\mathrm{skill}} \gets e_z(\mathbf{Z})$ \Comment{Encode observations and discrete skills}
    \State $\tilde{\mathbf{E}}^{\mathrm{skill}} \gets \textsc{TrimOrPad}(\textsc{RepeatInterleave}(\mathbf{E}^{\mathrm{skill}}, I), |\mathbf{E}^{\mathrm{obs}}|)$ \Comment{Upsample skills to match obs}
    \State $\mathbf{E}^{\mathrm{fused}} \gets \mathbf{E}^{\mathrm{obs}} + \tilde{\mathbf{E}}^{\mathrm{skill}}$ \Comment{Element-wise embedding fusion}
    \State $\hat{\mathbf{a}} \gets \pi_{\mathrm{fast}}(\mathbf{E}^{\mathrm{fused}}, \mathbf{a}, \mathbf{t}, \mathbf{m})$ \Comment{Autoregressive action prediction}
    \State $\mathcal{L}_{\mathrm{action}} \gets \textsc{MSE}(\hat{\mathbf{a}}, \mathbf{a}; \mathbf{m})$ \Comment{Action prediction loss masked by $\mathbf{m}$}

    \Statex \vspace{-0.2em}
    \Statex \hspace{-1em} \textit{\# 3. Latent Trajectory Diffusion Regularization}
    \If{Latent Diffusion $\epsilon_\theta$ is enabled}
        \State $H' \gets \min(H, |\mathbf{E}^{\mathrm{fused}}|)$
        \State $\tau_0 \gets \mathbf{E}_{1:H'}^{\mathrm{fused}}$ \textbf{and} $\mathbf{Z}_{H'} \gets \textsc{TrimOrPad}(\textsc{RepeatInterleave}(\mathbf{E}^{\mathrm{skill}}, I), H')$
        \State Sample step $n \sim \mathcal{U}(1, T_d)$ and noise $\epsilon \sim \mathcal{N}(\mathbf{0}, \mathbf{I})$ \Comment{Bold $\mathbf{I}$ for Identity matrix}
        \State $\tau_n \gets \sqrt{\bar{\alpha}_n}\tau_0 + \sqrt{1 - \bar{\alpha}_n}\epsilon$ \Comment{Forward noise injection}
        \State $\mathcal{L}_{\mathrm{diff}} \gets \| \epsilon - \epsilon_\theta(\tau_n, n, \mathbf{Z}_{H'}) \|_2^2$ \Comment{Reverse denoising loss}
    \Else
        \State $\mathcal{L}_{\mathrm{diff}} \gets 0$
    \EndIf

    \Statex \vspace{-0.2em}
    \Statex \hspace{-1em} \textit{\# 4. Joint Optimization}
    \State $\mathcal{L}_{\mathrm{total}} \gets \mathcal{L}_{\mathrm{action}} + \lambda_{\mathrm{vq}}\mathcal{L}_{\mathrm{vq}} + \lambda_{\mathrm{diff}}\mathcal{L}_{\mathrm{diff}}$
    \State \textsc{Backprop}($\mathcal{L}_{\mathrm{total}}$);\ \textsc{ClipGradients}();\ \textsc{OptimizerStep}()
\EndWhile
\end{algorithmic}
\end{algorithm}

\section{Datasets Descriptions.}
\label{app:datasets_descriptions}

\subsection{LOReL Sawyer.}
\label{app:lorel_descriptions}

\textbf{LOReL Sawyer Dataset.} The LOReL (Language-conditioned Offline Reward Learning) Sawyer dataset comprises pseudo-expert trajectories or play data collected from arbitrary Reinforcement Learning (RL) policies, annotated with hindsight crowdsourced language instructions. Developed on the Meta-World benchmark, this dataset features a Sawyer robot interacting with a drawer, a faucet, and two mugs (see Figure \ref{fig:sub1}). It contains 50,000 trajectories---each with a horizon of 20 time steps---collected by random RL policies within the environment and labeled using procedurally generated hindsight language instructions. We utilize the same six tasks as the original paper to evaluate our method. Performance is assessed across five distinct scenarios: seen instructions, unseen verbs, unseen nouns, unseen verbs+nouns, and human-provided instructions. Examples of these linguistic variations are provided in Table \ref{tab:lorl-example-rephrasals}. Collectively, the six tasks encompass 77 instructions.

\begin{table}[htbp]
  \centering
  \small
  \caption{LOReL Example rephrasals for the instruction.} 
  \label{tab:lorl-example-rephrasals}
  \resizebox{\linewidth}{!}{ 
  \begin{tabular}{c|c|c|c|c} 
    \toprule
    \textbf{Seen} & \textbf{Unseen Verb} & \textbf{Unseen Noun} & \textbf{Unseen Verb + Noun} & \textbf{Human Provided} \\ 
    \midrule
    close drawer & shut drawer & close container & shut container & push the drawer shut \\
    open drawer & pull drawer & open container & pull container & pull the drawer open \\
    turn faucet left & rotate faucet left & turn tap left & rotate tap left & rotate nozzle left \\
    turn faucet right & rotate faucet right & turn tap right & rotate tap right & rotate nozzle right \\
    move black mug right & push black mug right & move dark cup right & push dark cup right & push black cup right \\
    move white mug down & push white mug down & move light cup down & push light cup down & bring white cup down \\
    \bottomrule
  \end{tabular}
  } 
\end{table}

\textbf{Composition Instructions.} For the composition instructions, we took these evaluation instructions from the original paper and combined them to form 12 new composition instructions, as detailed in Table \ref{tab:lorel-composite-instrs}. 

\begin{table}[htbp]
  \centering
  \small
  \caption{LOReL Composition tasks}  
  \begin{tabular}{c} 
    \toprule
    \textbf{Composite Instructions}\\
    \midrule
    open drawer and move black mug right \\
    pull the handle and move black mug down \\
    move white mug right \\
    move black mug down \\
    close drawer and turn faucet right \\
    close drawer and turn faucet left \\
    turn faucet left and move white mug down \\
    turn faucet right and close drawer \\
    move white mug down and turn faucet left \\
    close the drawer, turn the faucet left and move black mug right \\
    open drawer and turn faucet counterclockwise \\
    slide the drawer closed and then shift white mug down \\
    \bottomrule %
  \end{tabular}
  \label{tab:lorel-composite-instrs}
\end{table}

\subsection{Franka-Kitchen.}
\label{app:Franka-Kitchen-descriptions}
\textbf{Franka-Kitchen Dataset.} Kitchen benchmark \cite{gupta2019relay} is a physics-based simulation environment built upon the MuJoCo engine \cite{todorov2012mujoco}, where a Franka robotic manipulator operates in a kitchen scenario containing multiple interactive objects, including a microwave, a kettle, a sliding cabinet, a hinged cabinet, a switch, and two stove burners (see Figure \ref{fig:sub2}). The robot is actuated through joint position control, and the environment provides both low-dimensional state observations and RGB visual inputs. This dataset comprises 566 human demonstrations collected via VR teleoperation. In each trajectory, the demonstrator sequentially performs four manipulation actions on different kitchen components. 

\textbf{Language Labeling.} Given the absence of native linguistic labels in the base environment, we systematically synthesize natural language instructions for the demonstrations. Each trajectory is characterized by a sequence of four executed sub-tasks, which we directly map to their corresponding atomic phrases (detailed in Table \ref{tab:kitchen_atomic_instructions}). To construct the final composite instruction for an entire episode, these four atomic descriptions are chronologically chained together using the conjunction ``and''.

\begin{table}[htbp] 
  \centering
  \small 
  \caption{Atomic instructions labeled to sub-tasks.}
  \label{tab:kitchen_atomic_instructions}
  \begin{tabular}{l l}
    \toprule
    \textbf{Sub-tasks} & \textbf{Atomic Instructions} \\
    \midrule
    bottom burner & activate bottom burner \\
    top burner & activate top burner \\
    light switch & turn on light switch \\
    slide cabinet & open sliding cabinet \\
    hinge cabinet & open left hinge cabinet \\
    microwave & open microwave door \\
    kettle & move kettle to top left burner \\
    \bottomrule
  \end{tabular}
\end{table}

\textbf{Seen and Unseen Instructions.} To evaluate compositional generalization, we partition the dataset by holding out three specific composite instructions, leaving 509 demonstrations across 22 distinct instructions for training. Crucially, while the specific combinations are unseen during testing, every individual sub-task is thoroughly represented in the training distribution. We utilize the following three held-out instructions for evaluation:

\begin{itemize}
    \item \textit{Open microwave door and move kettle to top left burner and activate bottom burner and turn on light switch.}
    \item \textit{Move kettle to top left burner and activate bottom burner and open sliding cabinet and open left hinge cabinet.}
    \item \textit{Move kettle to top left burner and activate top burner and turn on light switch and open sliding cabinet.}
\end{itemize}

These evaluation tasks are strategically designed to present an escalating level of compositional difficulty. Specifically, the training corpus contains trajectories that perfectly match the \textit{first three} sub-tasks of the first instruction. For the second instruction, this overlap is reduced to only the \textit{first two} sub-tasks. The third instruction poses the most rigorous out-of-distribution challenge, as the training data lacks any sequences sharing even its two-task prefix. Finally, to systematically assess compositional capabilities, we quantify model performance by counting the number of sub-tasks successfully executed within a strict 280-step episode limit.

\subsection{CALVIN}
\label{app:CALVIN-descriptions}

\textbf{CALVIN Dataset.} CALVIN (Composing Actions from Language and Vision) \citep{mees2022calvin} is an open-source simulated benchmark designed to evaluate language-conditioned robotic manipulation. The environment features a 7-DOF Franka Emika Panda robot interacting with a diverse tabletop setting, as depicted in Figure~\ref{fig:sub3}. The proposed method is evaluated on a subset of the CALVIN benchmark, focusing strictly on six specific single tasks: Open Drawer, Close Drawer, Turn on Lightbulb, Turn off Lightbulb, Move Slider Left, and Move Slider Right. This targeted selection is fundamentally driven by the choice of state representation. Specifically, to eliminate the variance introduced by different RGB image encoders and camera perspectives, and to focus solely on evaluating the underlying policy, a 21-dimensional perspective state vector is utilized (consistent with LCSD \cite{ju2024rethinking}). Because this low-dimensional state space lacks the spatial coordinates of the unattached objects (e.g., the colored blocks), tasks requiring object grasping or displacement are explicitly excluded. Consequently, a constrained offline dataset is compiled by directly selecting relevant trajectories from the CALVIN-D split.

\textbf{Language Instructions.} The CALVIN dataset features significant linguistic diversity, utilizing distinct phrasing during training and evaluation to rigorously test the semantic understanding of the model. The specific linguistic settings, encompassing the diverse training and evaluation prompts for the dataset configuration, are detailed in Table~\ref{tab:calvin-dataset-language-settings}.

\begin{table}[htbp]
  \centering
  \vspace{-1em}
  \small
  \caption{CALVIN dataset language settings.} 
  \label{tab:calvin-dataset-language-settings}
  \resizebox{\textwidth}{!}{
  \begin{tabular}{llp{7.5cm}p{5.5cm}c} 
    \toprule
    \textbf{Category} & \textbf{Detail Tasks} & \textbf{Training Language} & \textbf{Evaluation Language} & \textbf{Num} \\ 
    \midrule
    lightbulb & turn on/off lightbulb & toggle the light switch to turn on/off the light bulb \newline turn on/off the light bulb \newline move the light switch to turn on/off the light bulb & use the switch to turn on/off the light bulb & 95 \\
    \midrule
    drawer & open/close drawer & open/close the cabinet drawer \newline grasp the drawer handle and open/close it \newline pull/push the handle of the drawer & pull/push the handle to open/close the drawer & 303 \\
    \midrule
    slider & move slider left/right & grasp the door handle, then slide the door to the left/right \newline move the door all the way to the left/right \newline slide/push the door to the left/right & push the sliding door to the left/right side & 382 \\
    \midrule
    light & push down the button to turn on/off the led & press the button to turn on/off the led light \newline turn on/off the led light \newline toggle the button to turn on/off the led light & push the sliding door to the left/right side & 382 \\
    \midrule
    Unknown tasks & \multicolumn{3}{c}{Turn on/off green/yellow lamp move/toggle the light switch to turn off the yellow/green light} & 227 \\
    \bottomrule
  \end{tabular}
  }
\end{table}

\subsection{BabyAI.}
\label{app:BabyAI-descriptions}
\textbf{BabyAI Dataset.} BabyAI presents an environment where an agent navigates and interacts within a grid world to achieve goals described via natural language. The dataset includes various environmental configurations where level difficulty and instruction complexity increase progressively. Each level is situated in a grid world where the agent observes a partially visible $7 \times 7$ square area from an egocentric perspective. While the dataset provides one million expert trajectories for each of the 19 levels, we evaluate DASL across varying data regimes, specifically utilizing subsets of 1k, 10k, and 100k trajectories. We evaluate our policy on 100 distinct instructions within the Gym environment for each level, covering unseen layouts and language instructions to assess generalization capabilities.

\begin{figure}[ht]
\centering
\vspace{-0.5em} %

\begin{subfigure}[b]{0.24\textwidth}
    \centering
    \includegraphics[width=\textwidth]{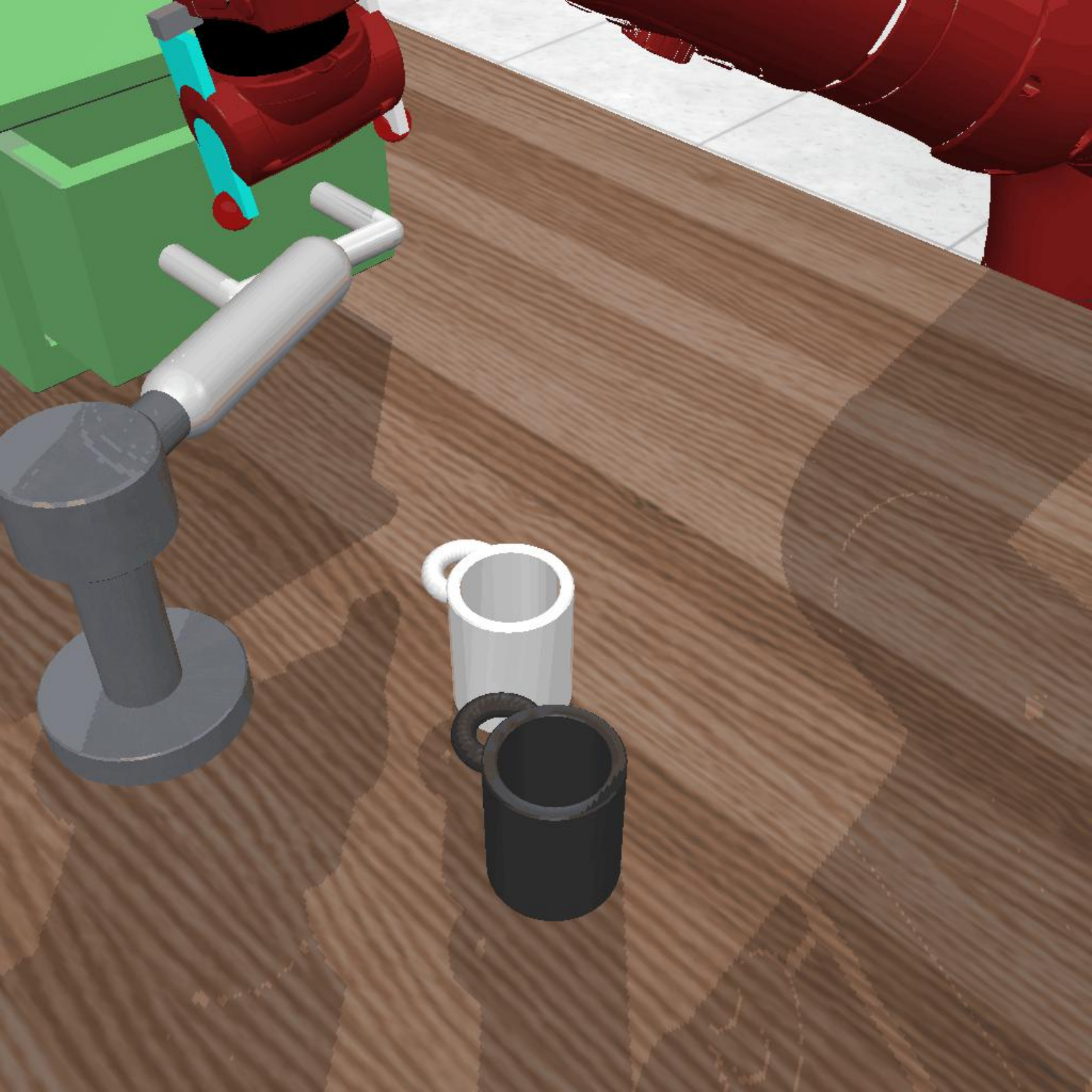} 
    \caption{LOReL-Sawyer}
    \label{fig:sub1}
\end{subfigure}
\hfill 
\begin{subfigure}[b]{0.24\textwidth}
    \centering
    \includegraphics[width=\textwidth]{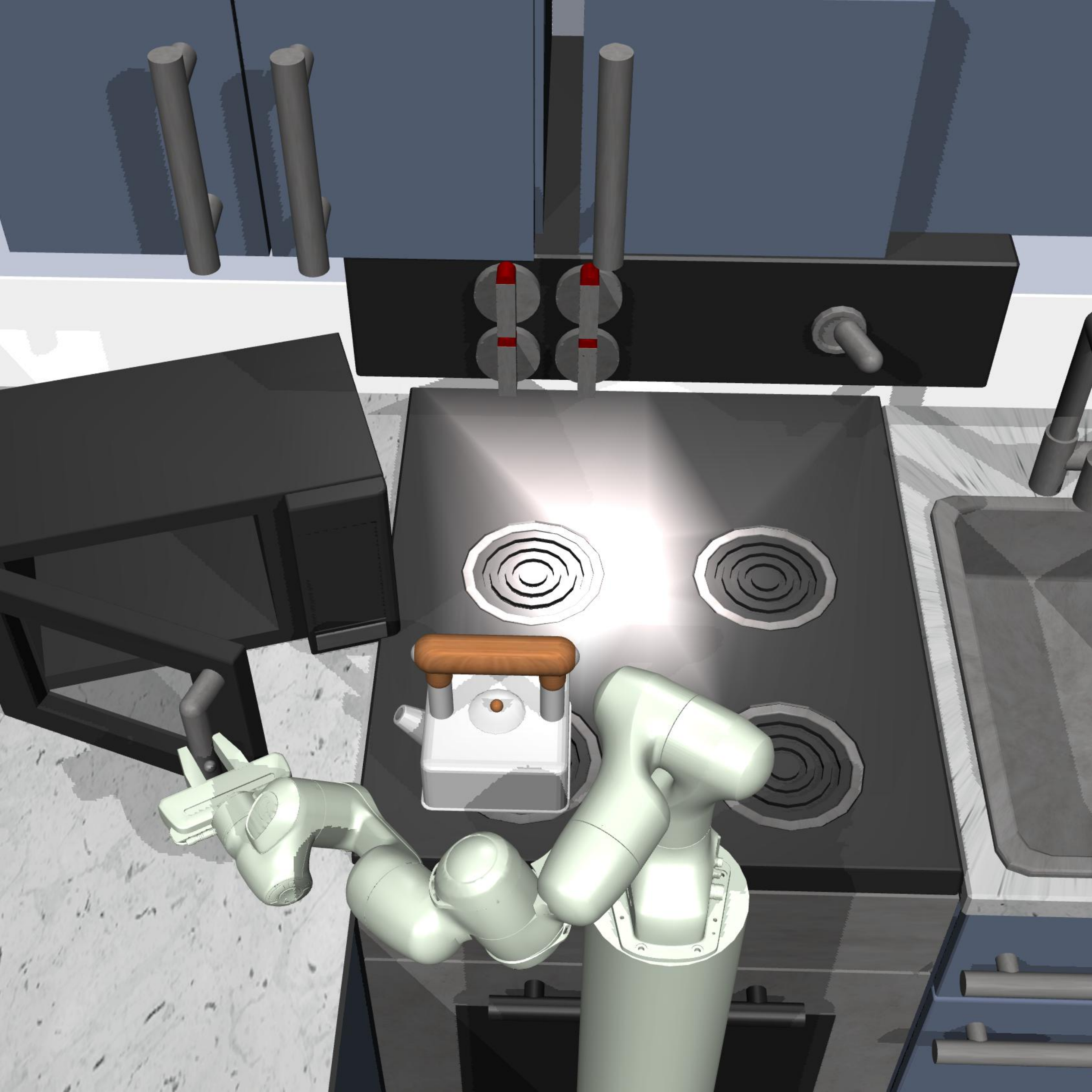}
    \caption{Franka Kitchen}
    \label{fig:sub2}
\end{subfigure}
\hfill
\begin{subfigure}[b]{0.24\textwidth}
    \centering
    \includegraphics[width=\textwidth]{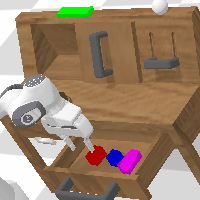}
    \caption{CALVIN}
    \label{fig:sub3}
\end{subfigure}
\hfill
\begin{subfigure}[b]{0.24\textwidth}
    \centering
    \includegraphics[width=\textwidth]{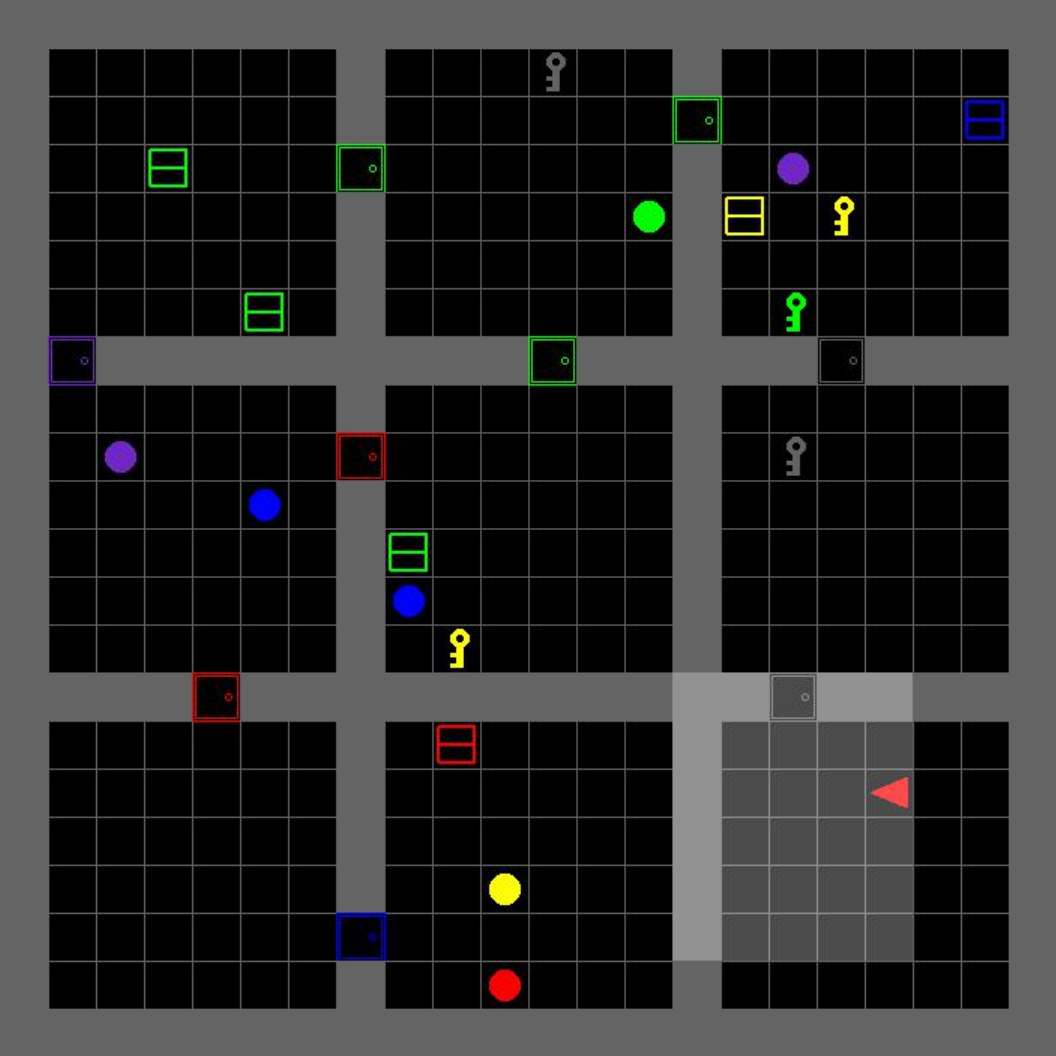}
    \caption{BabyAI GoToSeq}
    \label{fig:sub4}
\end{subfigure}

\vspace{-0.5em} %
\caption{Simulation environments for manipulation and navigation tasks.}
\label{fig:envs}

\vspace{-1.5em} %
\end{figure}

\textbf{GoToSeq Task.} This task requires the agent to execute a sequence of go-to-object commands (as illustrated in Figure \ref{fig:sub3}).
\begin{itemize}
    \item \textbf{Example command:} \textit{``go to a box and go to the purple door, then go to the grey door''}
    \item \textbf{Average demo length:} $72.7 \pm 52.2$ steps
\end{itemize}

\section{Baselines Descriptions.}
\label{app:baselines_descriptions}

\textbf{LCBC.} Language-Conditioned Behavioral Cloning (LCBC) serves as a supervised imitation learning baseline that directly maps language instructions and observations to actions using demonstration trajectories. As a straightforward language-conditioned policy, it relies neither on hierarchical planning nor reinforcement learning. We directly report the success rate of LCBC on the BabyAI GoToSeq task from \citet{garg2022lisa}.

\textbf{Lang DT.} Language Decision Transformer (langDT) extends the standard Decision Transformer by conditioning the transformer policy on language instructions alongside state and action histories. It models the sequential decision-making process in a flat manner, operating without explicit skill abstractions or hierarchical structures. For performance comparison, we directly report the success rates of Lang DT on the LOReL(image) tasks from \citet{liang2024skilldiffuser}, on the Franka-Kitchen benchmark from \citet{jiang2025discrete}, on the CALVIN benchmark from \citet{ju2024rethinking} and on the BabyAI GoToSeq task from \citet{garg2022lisa}. Since \citet{liang2024skilldiffuser} do not provide the results for the LOReL(state) tasks, we evaluate their official implementation\footnote{\label{fn:lisagithub}\url{https://github.com/Div99/LISA}} to obtain the corresponding metrics.

\textbf{LISA.} Learning Interpretable Skill Abstractions (LISA)  employs a Vector Quantization (VQ) bottleneck to derive a discrete library of latent skills from language-conditioned demonstrations. This hierarchical framework partitions long-term objectives into discrete skill representations, which subsequently guide a conditioned low-level control policy. For our benchmarking, we directly report LISA's success rates on the LOReL(image) tasks from \citet{liang2024skilldiffuser}, on Franka-Kitchen from \citet{jiang2025discrete}, on the CALVIN benchmark from \citet{ju2024rethinking} and on the BabyAI GoToSeq mission from \citet{garg2022lisa}. Due to the absence of LOReL(state) metrics in \citet{liang2024skilldiffuser}, we conducted an independent reproduction using the authorized implementation\,\textsuperscript{\ref{fn:lisagithub}}. Our empirical findings align with the issues highlighted by \citet{ju2024rethinking} and \citet{jiang2025discrete}: the VQ-based discovery mechanism frequently suffers from \textit{index collapse}, where only a fraction of the codebook is effectively utilized, leading to sub-optimal performance on LOReL(state) tasks. This observed instability in hierarchical skill discovery serves as a pivotal motivation for the design of our DASL framework.

\textbf{HULC.} Hierarchical Universal Language
Conditioned (HULC) serves as the official baseline introduced alongside the CALVIN benchmark \citep{mees2022calvin}, employing a multi-context imitation learning approach to execute language-conditioned manipulation tasks. We trained the HULC baseline from (Mees et al. 2022) on our constrained CALVIN setting by replacing its original vision encoder with a simple MLP, thereby adapting the model to directly process low-dimensional perspective state observations. We directly report the success rate of HULC on the CALVIN benchmark from \citet{ju2024rethinking}.

\textbf{LCSD.} Language-Conditioned Skill Discovery (LCSD) focuses on learning discrete skill representations by maximizing the mutual information between linguistic instructions and latent skills. This objective enables the agent to acquire a structured skill space, facilitating compositional generalization by associating each unique instruction with a corresponding learned skill embedding. The success rates of LCSD on both LOReL(state) and LOReL(image) are directly reported from the original study by \citet{ju2024rethinking}.

\textbf{SkillDiffuser.} SkillDiffuser is a diffusion-based hierarchical policy that generates latent skill trajectories conditioned on both language instructions and environmental observations. The framework utilizes a low-level controller to execute actions derived from predicted skill sequences, facilitating robust compositional planning through flexible skill composition. We directly report the success rates of SkillDiffuser on LOReL(image) from the original study by \citet{liang2024skilldiffuser}. As LOReL(state) results were not provided in \cite{liang2024skilldiffuser}, we conducted an independent evaluation using the official source code\,\footnote{\url{https://github.com/Liang-ZX/SkillDiffuser}} to obtain the corresponding metrics.

\textbf{LADS.} Language-Aware Discrete Skills (LADS) disentangles high-level language-conditioned planning from low-level skill execution by learning a discrete latent action space. This framework explicitly models task decomposition, exhibiting remarkable robustness in handling complex compositional tasks and unseen linguistic instructions. We report LADS's evaluation results on the Franka-Kitchen benchmark directly from \citet{jiang2025discrete}. However, as the original study does not provide detailed success rates for varied instruction rephrasals on the LOReL benchmark, we conducted an independent evaluation using their official implementation\,\footnote{\url{https://github.com/PKU-RL/LADS}} to obtain the corresponding metrics for both LOReL(state) and LOReL(image).

\section{Training Details}
\subsection{Network Architecture}
\label{app:Network_Architecture}

\textbf{Language Encoder.} 
We employ a pre-trained DistilBERT as the language encoder, maintaining a configuration consistent with LOReL and LISA, while freezing its parameters to ensure the stability of language understanding. Given a natural language instruction $l$, it is encoded into a word embedding sequence $\mathbf{E}^{\mathrm{lang}} = \text{DistilBERT}(l) \in \mathbb{R}^{L_{\mathrm{lang}} \times d}$, where $L_{\mathrm{lang}}$ is the token length and $d$ is the shared hidden dimension. This helps preserve the high-level semantic information of the language instructions.

\textbf{Observation Encoder.}
We consider both vector observations and pixel observations. For vector inputs (e.g., proprioceptive states or precomputed visual embeddings), we apply a linear projection to match the transformer hidden dimension $d$. For image inputs, we use a lightweight convolutional encoder $\phi$ (following prior LOReL-style implementations) to map each frame into a fixed-dimensional embedding. 

\textbf{Slow Frequency Policy.}
The slow policy acts as a semantic predictor that produces a high-level latent skill at a lower temporal frequency, updating once every $I$ environment steps (the fixed interval). Specifically, we employ an Option Transformer that predicts continuous skill embeddings by reasoning over both the language instruction and the subsampled observation sequence. These continuous skills are subsequently discretized via a vector-quantization (VQ) bottleneck, which is optimized using a standard commitment regularization.

\textbf{High Frequency Policy.}
The high frequency policy is a Decision Transformer that outputs low-level actions at every high-frequency time step. In our DASL framework, this transformer is explicitly conditioned on the selected discrete skill fused with the observation-action sequence. To align the high frequency policy's temporal receptive field with the slow semantic schedule, the retrieved skill embedding is repeated $I$ times to remain constant within each interval segment. Furthermore, to structure the latent manifold and ensure robust semantic alignment, we augment this low-level policy with an auxiliary latent diffusion module exclusively during training. This hybrid design ensures consistent low-level execution guided by stable high-level semantics, without compromising real-time inference efficiency.

\subsection{Training Hyperparameters.}

\label{app:Hyperparameters}
Table~\ref{tab:Hyperpar} details the comprehensive hyperparameter configurations for DASL across all environments, including LOReL, Franka Kitchen, BabyAI, and CALVIN. To ensure stable convergence, we consistently employ the Adam optimizer with a $1e^{-5}$ learning rate and $1e^{-4}$ weight decay across all tasks. Key structural parameters, such as the sequence horizon are scaled by environment complexity, ranging from 5 in LOReL (state) to 20 in Franka Kitchen. Notably, the diffusion module (configured with 100 timesteps and 4 blocks) is enabled across all settings to regularize the latent space. Batch sizes and iteration counts are tuned per dataset to balance computational efficiency with high-level semantic alignment.

\begin{table}[htbp]
    \centering
    \vspace{-1em}
    \small %
    \caption{\textbf{Hyperparameters on Different Datasets}}
    \label{tab:Hyperpar}
    \vspace{5pt} 

    \begin{tabular*}{\textwidth}{@{\extracolsep{\fill}} l c c c c c c @{}} 
        \toprule
        \textbf{Hyperparameter} & 
        \begin{tabular}{@{}c@{}}\textbf{LOReL}\\ \textbf{(state)}\end{tabular} & 
        \begin{tabular}{@{}c@{}}\textbf{LOReL}\\ \textbf{(image)}\end{tabular} & 
        \begin{tabular}{@{}c@{}}\textbf{Kitchen}\\ \textbf{(state)}\end{tabular} & 
        \begin{tabular}{@{}c@{}}\textbf{Kitchen}\\ \textbf{(image)}\end{tabular} & 
        \textbf{BabyAI} & \textbf{CALVIN} \\
        \midrule
        Batch Size              & 512       & 512       & 16        & 16        & 128/64/180 & 24 \\
        Learning Rate           & $1e^{-5}$ & $1e^{-5}$ & $1e^{-5}$ & $1e^{-5}$ & $1e^{-5}$  & $1e^{-5}$ \\
        Max Iterations          & 2500      & 2500      & 3000      & 3000      & 2500       & 5000 \\
        Horizon                 & 5         & 10        & 20        & 20        & 10         & 10 \\
        Slow Frequency Interval & 2         & 2         & 20        & 20        & 2          & 5 \\
        Option Dimension        & 128       & 128       & 128       & 128       & 128        & 128 \\
        Number of Options       & 20        & 20        & 20        & 30        & 20         & 20 \\
        Codebook Dimension      & 16        & 16        & 16        & 16        & 16         & 16 \\
        Diffusion Enabled       & True      & True      & True      & True      & True       & True \\
        Diffusion Loss Weight   & 0.5       & 0.01      & 0.1       & 0.1       & 0.1        & 0.1 \\
        Diffusion Timesteps     & 100       & 100       & 100       & 100       & 100        & 100 \\
        Diffusion Blocks        & 4         & 4         & 4         & 4         & 4          & 4 \\
        Warmup Steps            & 500      & 500      & 500      & 500      & 500       & 500 \\
        Evaluation Frequency    & 20        & 20        & 20        & 20        & 100        & 100 \\
        Evaluation Episodes     & 50         & 50         & 50         & 50         & 50        & 50 \\
        Weight Decay            & $1e^{-4}$ & $1e^{-4}$ & $1e^{-4}$ & $1e^{-4}$ & $1e^{-4}$  & $1e^{-4}$ \\
        Optimizer               & Adam      & Adam      & Adam      & Adam      & Adam       & Adam \\
        \bottomrule
    \end{tabular*}
    \vspace{-1.0em} 
\end{table}

\section{Supplementary Results}
\subsection{Additional Experimental Results on LOReL}
\label{app:lorel_result}
Table \ref{tab:lorel_res_compact} demonstrates that DASL establishes a new state-of-the-art performance across diverse manipulation tasks, achieving the highest average success rates in both state (53\%) and image (49\%) observation spaces. While certain baselines exhibit extreme sensitivity to the observation modality, such as Lang DT suffering a drastic performance degradation from 34\% to 15\% when transitioning from state to visual inputs, DASL maintains a highly stable performance footprint. This minimal performance gap between modalities indicates that the proposed architecture effectively mitigates the visual complexity and representation collapse typically inherent in high-dimensional image inputs.

\begin{table*}[htbp]
    \centering
    \vspace{-5pt}
    \small
    \setlength{\tabcolsep}{5pt} 
        
    \caption{\textbf{Task success rates on the LOReL Sawyer dataset.} We present the evaluated success rates (\%) of different algorithms across various tasks under both state and image observation settings. DASL outperforms all other methods in terms of average performance. The best results are highlighted in \textbf{bold}, and the second-best results are \underline{underlined}.}

    \newcommand{\res}[3][]{%
        \ifx\relax#1\relax%
            \makebox[1.25em][r]{#2} \scriptsize{\raisebox{.5pt}{$\pm$ \makebox[1.05em][l]{#3}}}%
        \else%
            #1{%
                \makebox[1.25em][r]{#2} %
                \scriptsize{\raisebox{.5pt}{$\pm$}} %
                \makebox[1.05em][l]{#3}%
            }%
        \fi%
    }
    \resizebox{\textwidth}{!}{
    \begin{tabular}{l c c c c c c c}
        \toprule
        \textbf{Task Instruction} & \multicolumn{1}{p{0.2cm}}{\centering \textbf{Obs}} & \textbf{LCBC} & \textbf{Lang DT} & \textbf{LISA} & \textbf{SkillDiffuser} & \textbf{LADS} & \textbf{DASL(ours)} \\
        \midrule
        \multirow{2}{*}{close drawer} & State & \res{45}{11} & \res{52}{14} & \res{52}{41} & \res{80}{6} & \res[\underline]{83}{19} & \res[\textbf]{96}{1}\\
        & Image & \res{50}{\textcolor{blue}{0}} & \res{10}{\textcolor{blue}{0}} & \res[\textbf]{100}{\textcolor{blue}{0}} & \res[\underline]{95}{3} &  \res{62}{18} & \res{50}{13}\\
        \multirow{2}{*}{open drawer} & State & \res{0}{0} & \res{33}{20} & \res{4}{6} & \res{31}{14} & \res[\textbf]{84}{12} & \res[\underline]{44}{12} \\
        & Image & \res{0}{\textcolor{blue}{0}} & \res[\textbf]{60}{\textcolor{blue}{0}} & \res{20}{\textcolor{blue}{0}} & \res[\underline]{55}{13} & \res{29}{19} & \res{46}{12}\\
        \multirow{2}{*}{turn faucet left} & State & \res{25}{18} & \res[\underline]{48}{22} & \res{7}{8} & \res{37}{10} & \res{27}{25} & \res[\textbf]{68}{2}\\
        & Image & \res{12}{\textcolor{blue}{0}} & \res{0}{\textcolor{blue}{0}} & \res{0}{\textcolor{blue}{0}} & \res[\textbf]{55}{9} & \res{27}{8} & \res[\underline]{40}{2}\\
        \multirow{2}{*}{turn faucet right} & State & \res{30}{7} & \res{12}{6} & \res{1}{1} & \res[\underline]{34}{6} & \res{25}{21} & \res[\textbf]{43}{17}\\
        & Image & \res{31}{\textcolor{blue}{0}} & \res{0}{\textcolor{blue}{0}} & \res{30}{\textcolor{blue}{0}} & \res{25}{4} & \res[\underline]{41}{17} & \res[\textbf]{50}{7}\\
        \multirow{2}{*}{move black mug right} & State & \res[\textbf]{65}{4} & \res{50}{13} & \res{0}{1} & \res{17}{2} & \res{55}{23} & \res[\underline]{55}{11}\\
        & Image & \res[\textbf]{73}{\textcolor{blue}{0}} & \res{20}{\textcolor{blue}{0}} & \res{60}{\textcolor{blue}{0}} & \res{18}{4} & \res{53}{8} & \res[\underline]{61}{26}\\
        \multirow{2}{*}{move white mug down} & State & \res{5}{8} & \res{10}{5} & \res{0}{0} & \res{10}{6} & \res[\textbf]{35}{3} & \res[\underline]{11}{10}\\
        & Image & \res{6}{\textcolor{blue}{0}} & \res{0}{\textcolor{blue}{0}} & \res[\underline]{30}{\textcolor{blue}{0}} & \res{10}{2} & \res{29}{7} & \res[\textbf]{33}{3}\\
          \midrule
        \multirow{2}{*}{\textbf{Average}} & State & \res{28}{5} & \res{34}{7} & \res{11}{7} & \res{35}{4} & \res[\underline]{52}{2} & \res[\textbf]{53}{2}\\
        & Image & \res{29}{\textcolor{blue}{0}} & \res{15}{\textcolor{blue}{0}}& \res{40}{\textcolor{blue}{0}} & \res[\underline]{43}{1} & \res{40}{4} & \res[\textbf]{47}{2}\\
        \bottomrule
    \end{tabular}
    }
    \label{tab:lorel_res_compact}
\end{table*}

A fine-grained analysis of individual tasks reveals that DASL excels in maneuvers requiring sustained directional control and precise spatial execution. For instance, in the state modality of the turn faucet left and turn faucet right tasks, DASL achieves 68\% and 43\% success rates respectively, whereas baselines like LISA completely fail, often scoring near zero. Furthermore, in tasks demanding decisive endpoint manipulation, such as close drawer, DASL reaches a near-perfect 96\% success rate under state observations. This consistency across diverse instruction types underscores the framework's adaptability to both continuous control semantics and discrete state changes.

\subsection{Qualitative Results in LOReL}
\label{app:Qualitative_Results_in_LOReL}

\begin{figure*}[htbp]
  \centering
  \begin{subfigure}[t]{0.19\textwidth}
    \includegraphics[width=\linewidth]{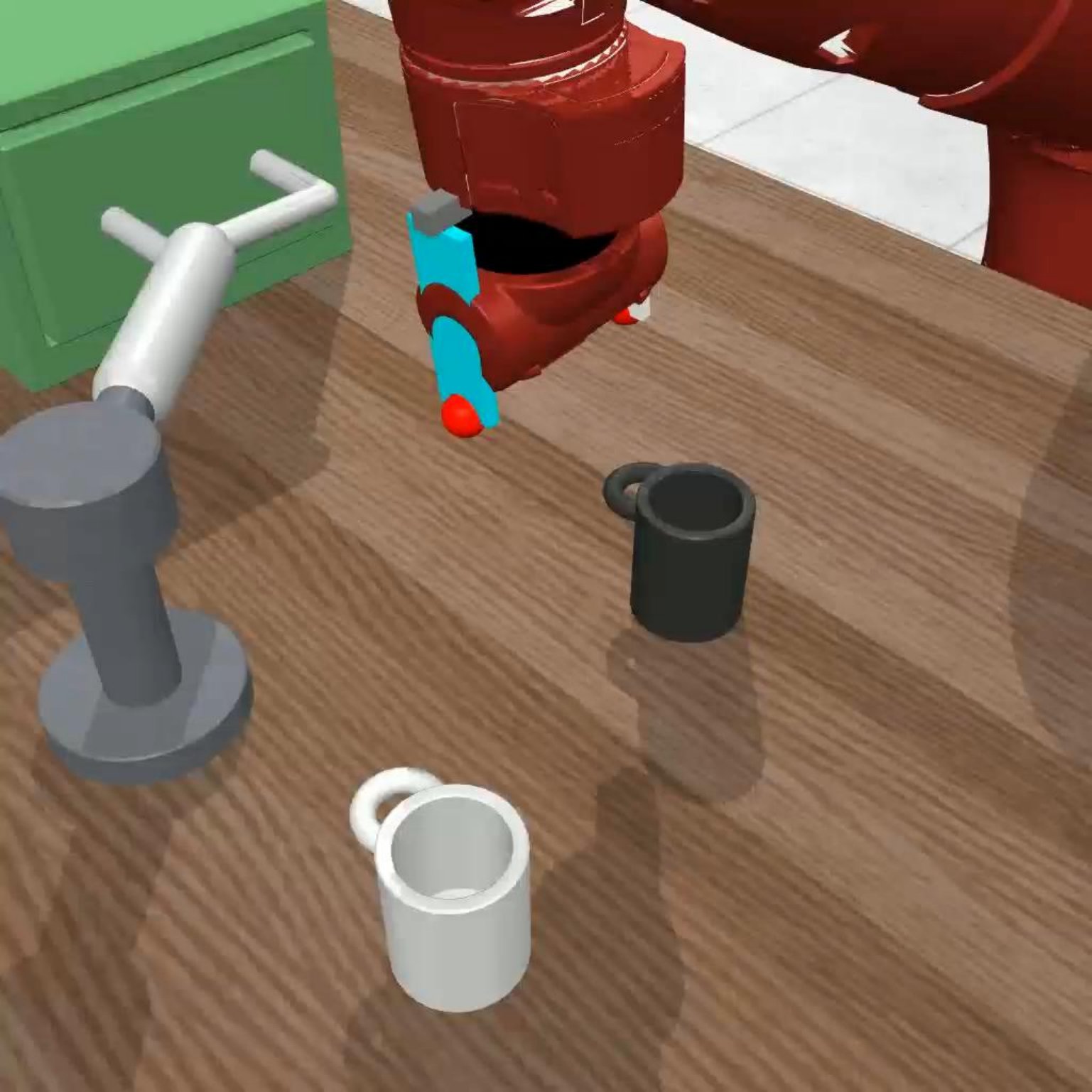}
    \caption{}
  \end{subfigure}\hfill
  \begin{subfigure}[t]{0.19\textwidth}
    \includegraphics[width=\linewidth]{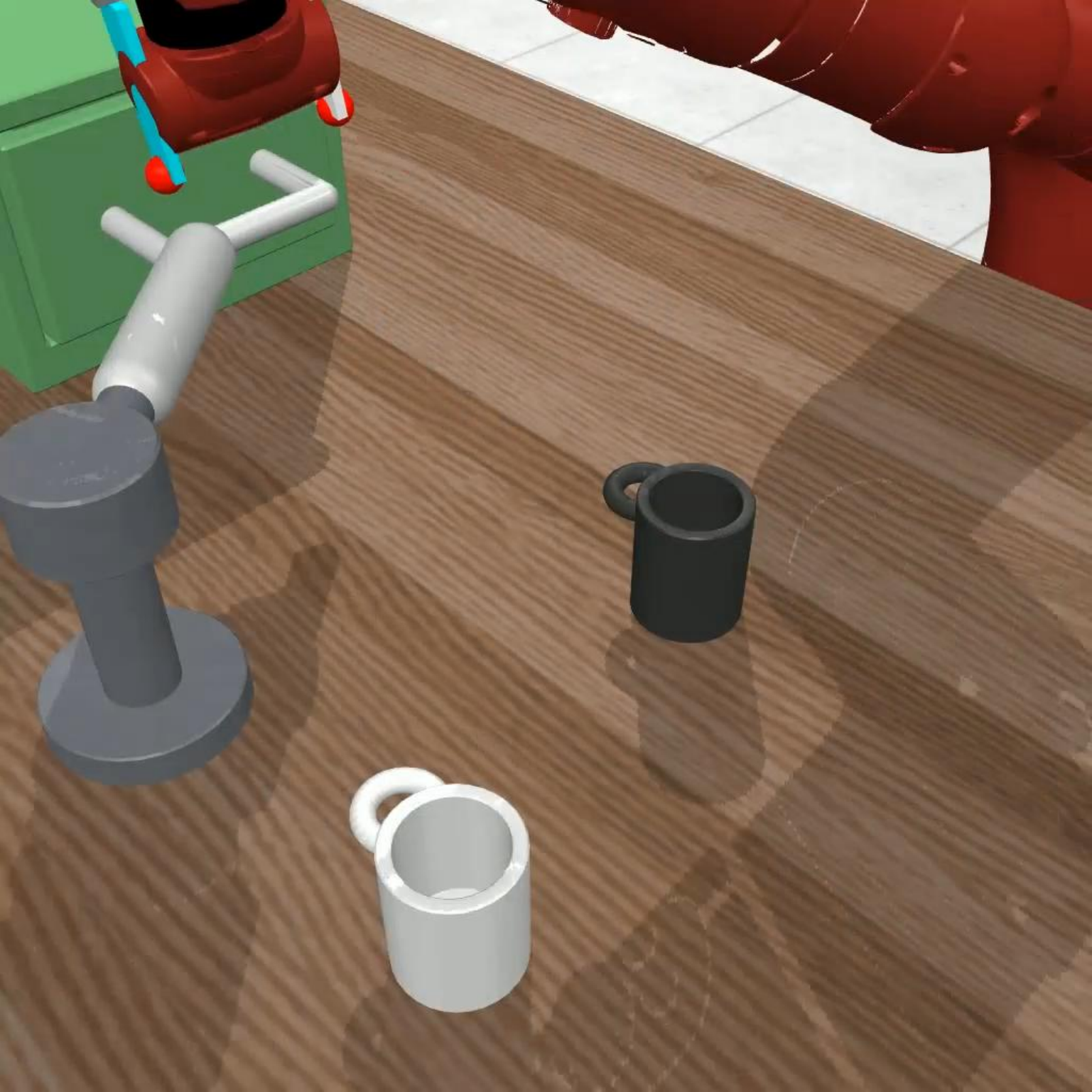}
    \caption{}
  \end{subfigure}\hfill
  \begin{subfigure}[t]{0.19\textwidth}
    \includegraphics[width=\linewidth]{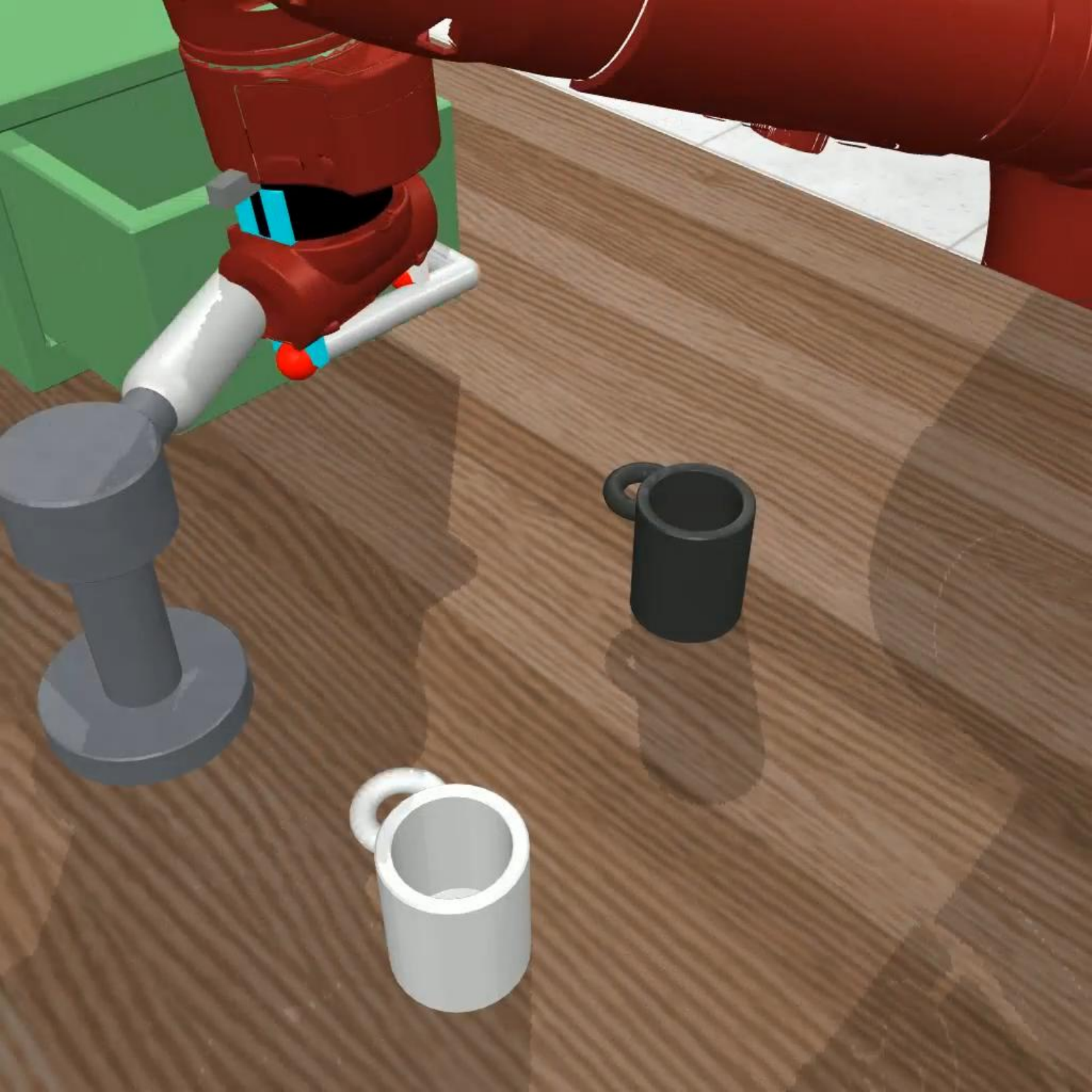}
    \caption{}
  \end{subfigure}\hfill
  \begin{subfigure}[t]{0.19\textwidth}
    \includegraphics[width=\linewidth]{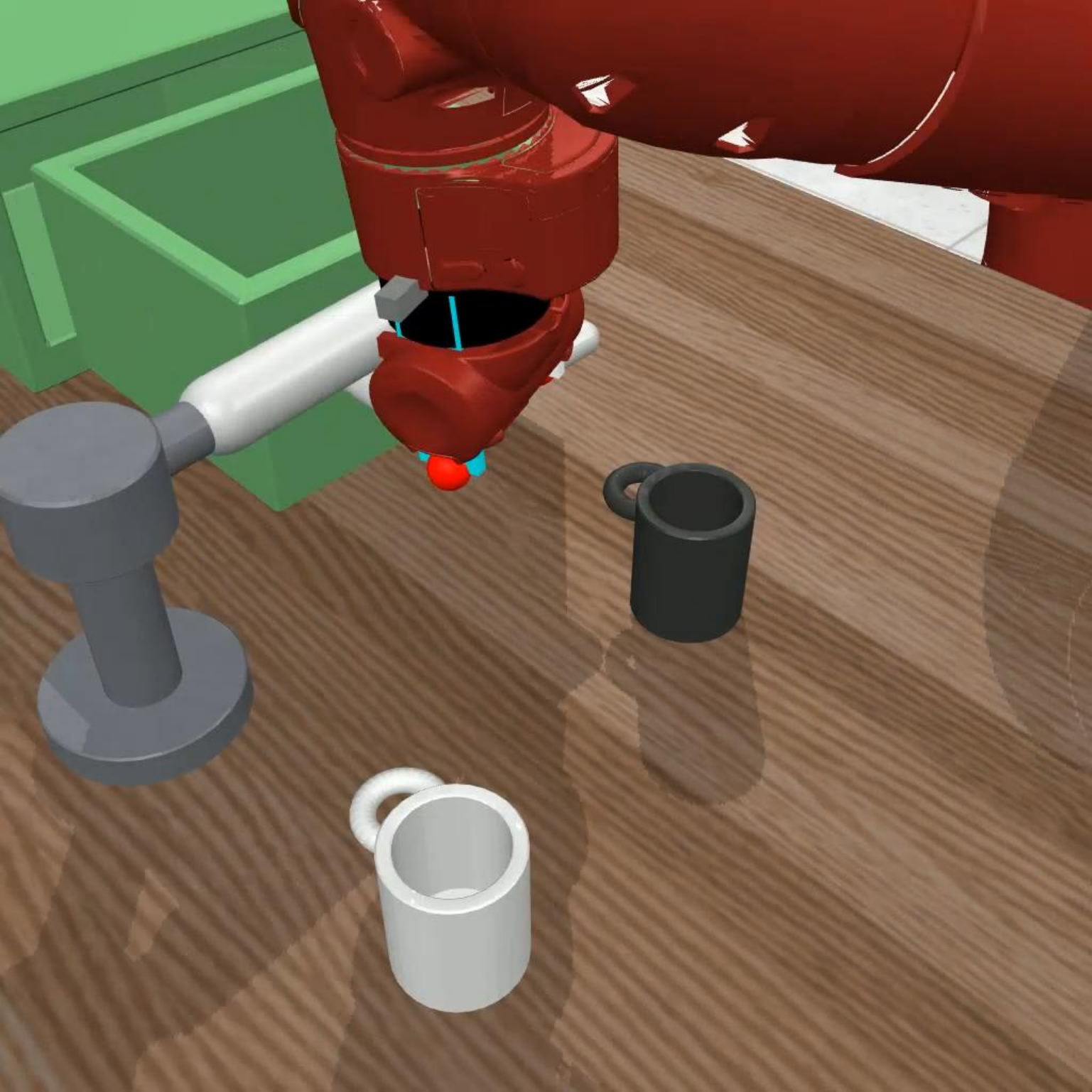}
    \caption{}
  \end{subfigure}\hfill
  \begin{subfigure}[t]{0.19\textwidth}
    \includegraphics[width=\linewidth]{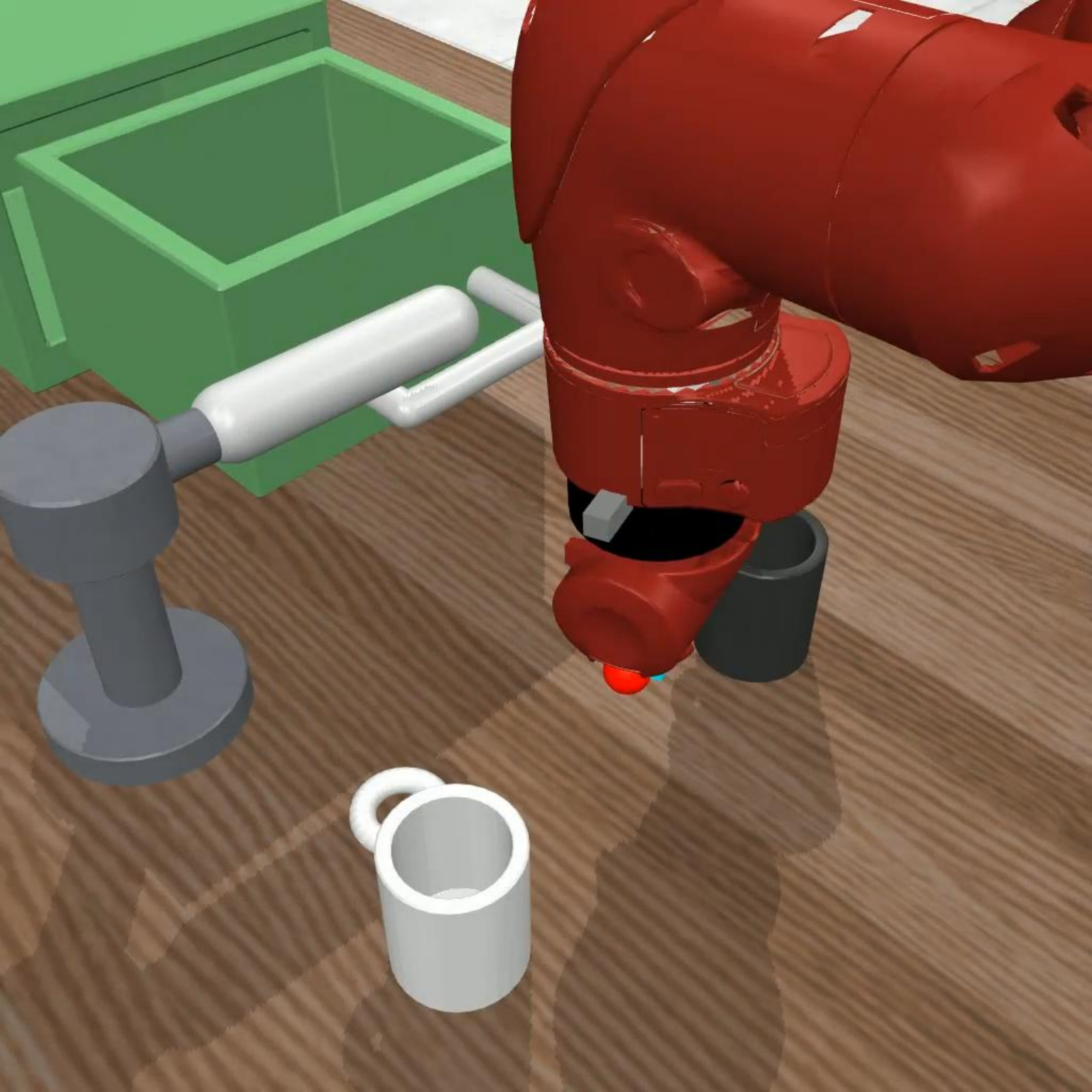}
    \caption{}
  \end{subfigure}
  \caption{LOReL State: open drawer and move black mug right}
  \label{fig:open drawer and move black mug right}
\end{figure*}

Figure~\ref{fig:open drawer and move black mug right} illustrates a qualitative execution sequence of the DASL framework performing a multi-step, compositional instruction---``open drawer and move black mug right''---within the LOReL State environment. The sequence demonstrates the capability of the model to parse and execute complex hierarchical commands. Panels (a) through (c) depict the robot arm locating and fully opening the drawer; subsequently, panels (d) and (e) show the agent transitioning to the second sub-task to manipulate the black mug. This visual trajectory validates the effectiveness of the proposed hierarchical architecture and the latent skill representations in maintaining semantic consistency and execution stability across multi-stage manipulation tasks.

\begin{figure*}[htbp]
  \centering
  \begin{subfigure}[t]{0.19\textwidth}
    \includegraphics[width=\linewidth]{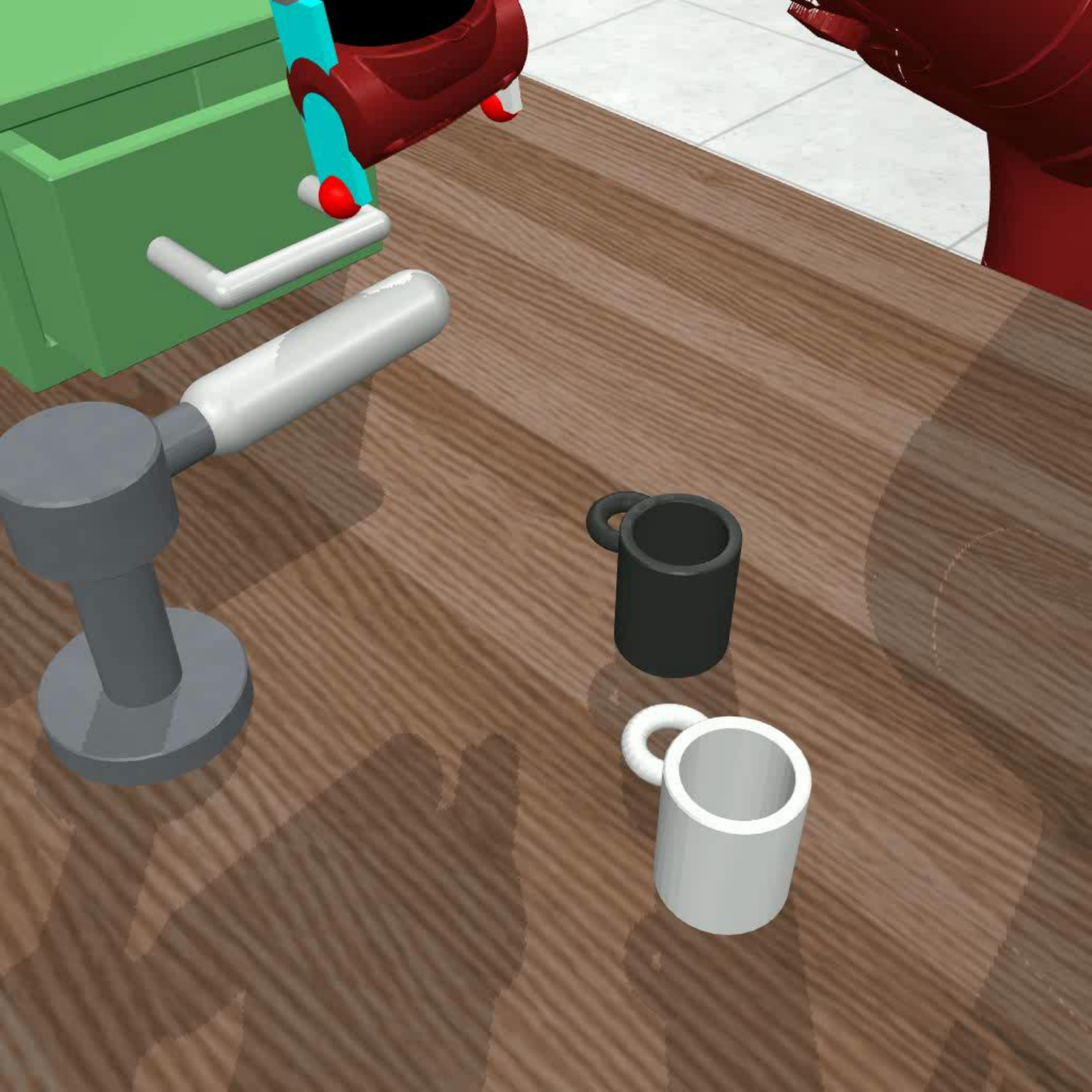}
    \caption{}
  \end{subfigure}\hfill
  \begin{subfigure}[t]{0.19\textwidth}
    \includegraphics[width=\linewidth]{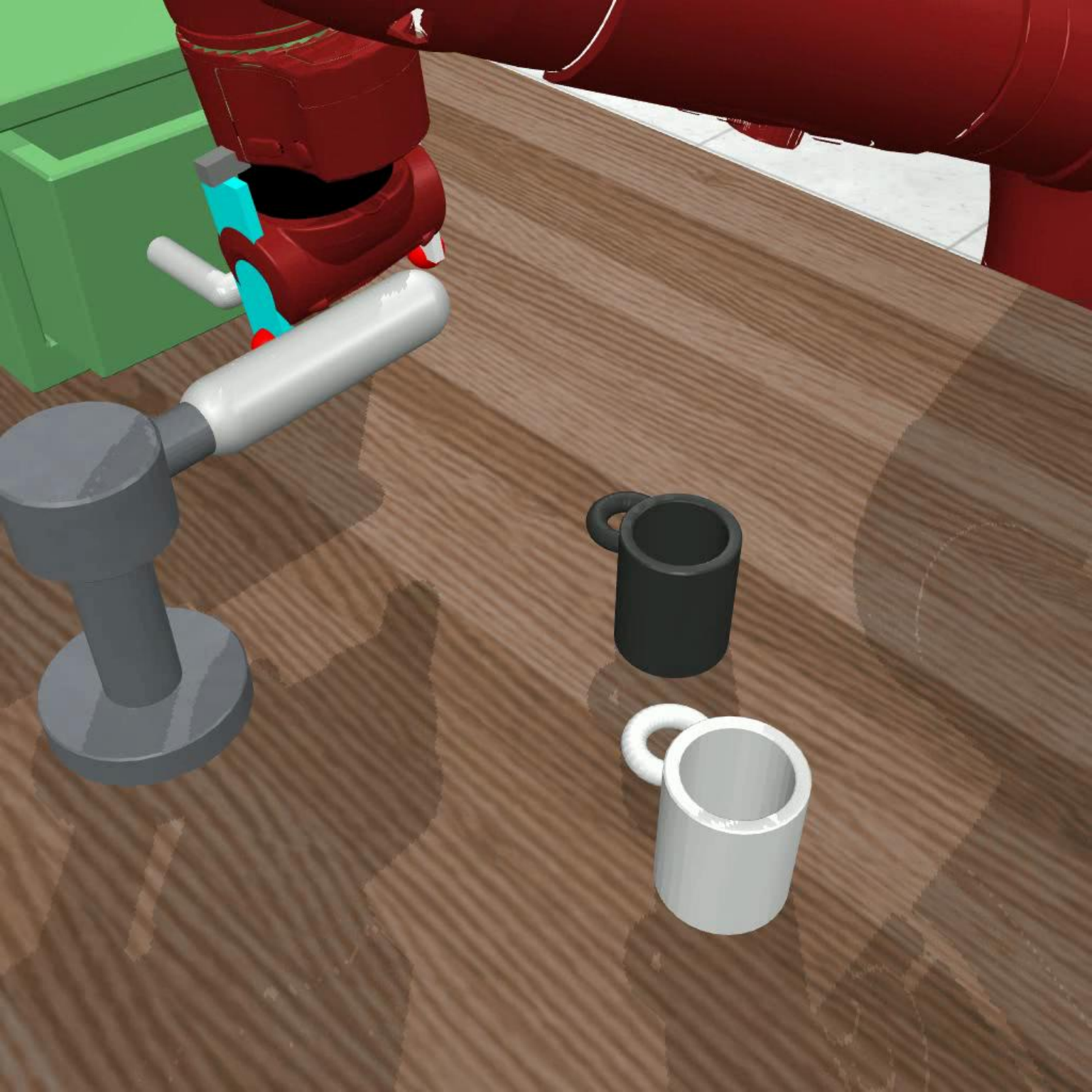}
    \caption{}
  \end{subfigure}\hfill
  \begin{subfigure}[t]{0.19\textwidth}
    \includegraphics[width=\linewidth]{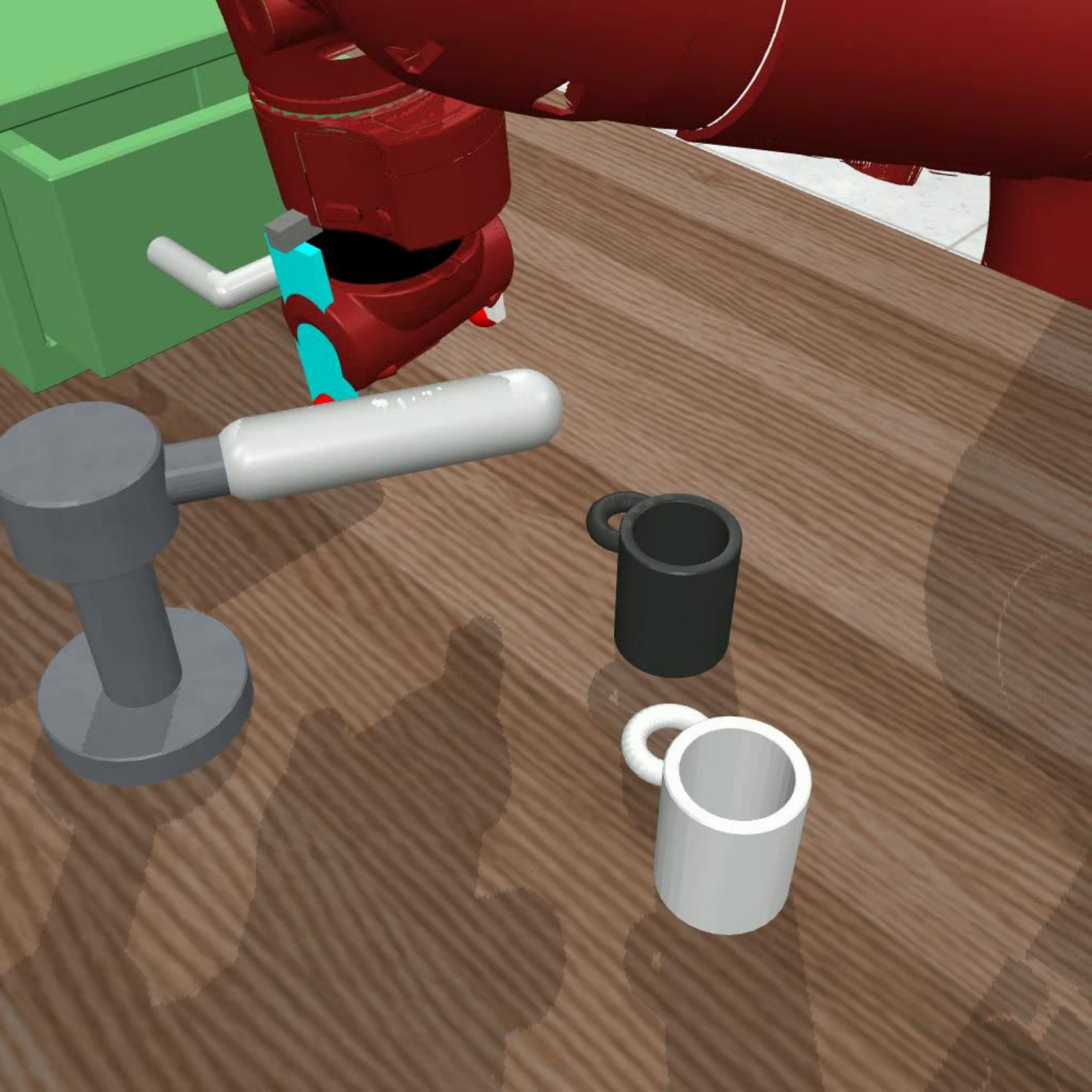}
    \caption{}
  \end{subfigure}\hfill
  \begin{subfigure}[t]{0.19\textwidth}
    \includegraphics[width=\linewidth]{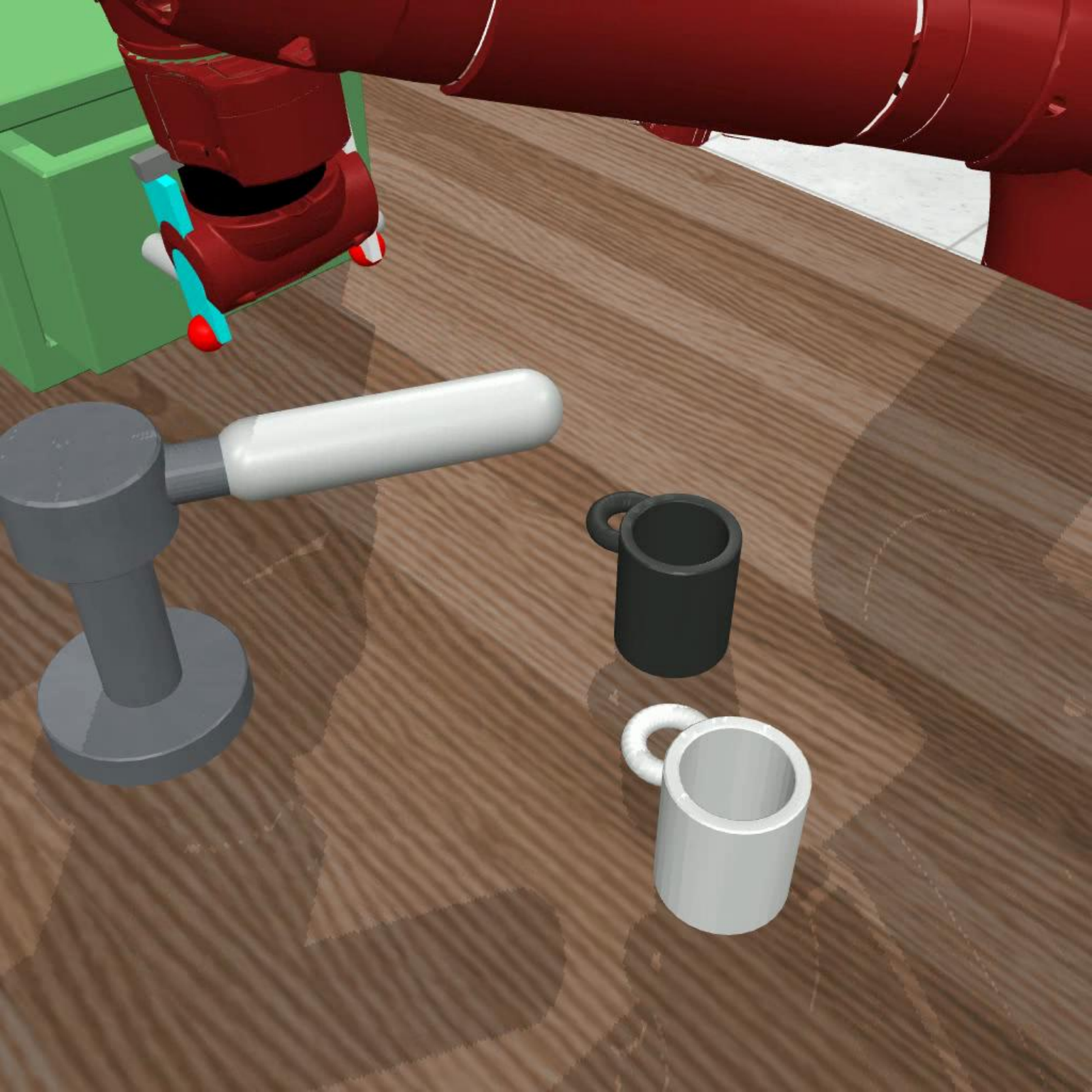}
    \caption{}
  \end{subfigure}\hfill
  \begin{subfigure}[t]{0.19\textwidth}
    \includegraphics[width=\linewidth]{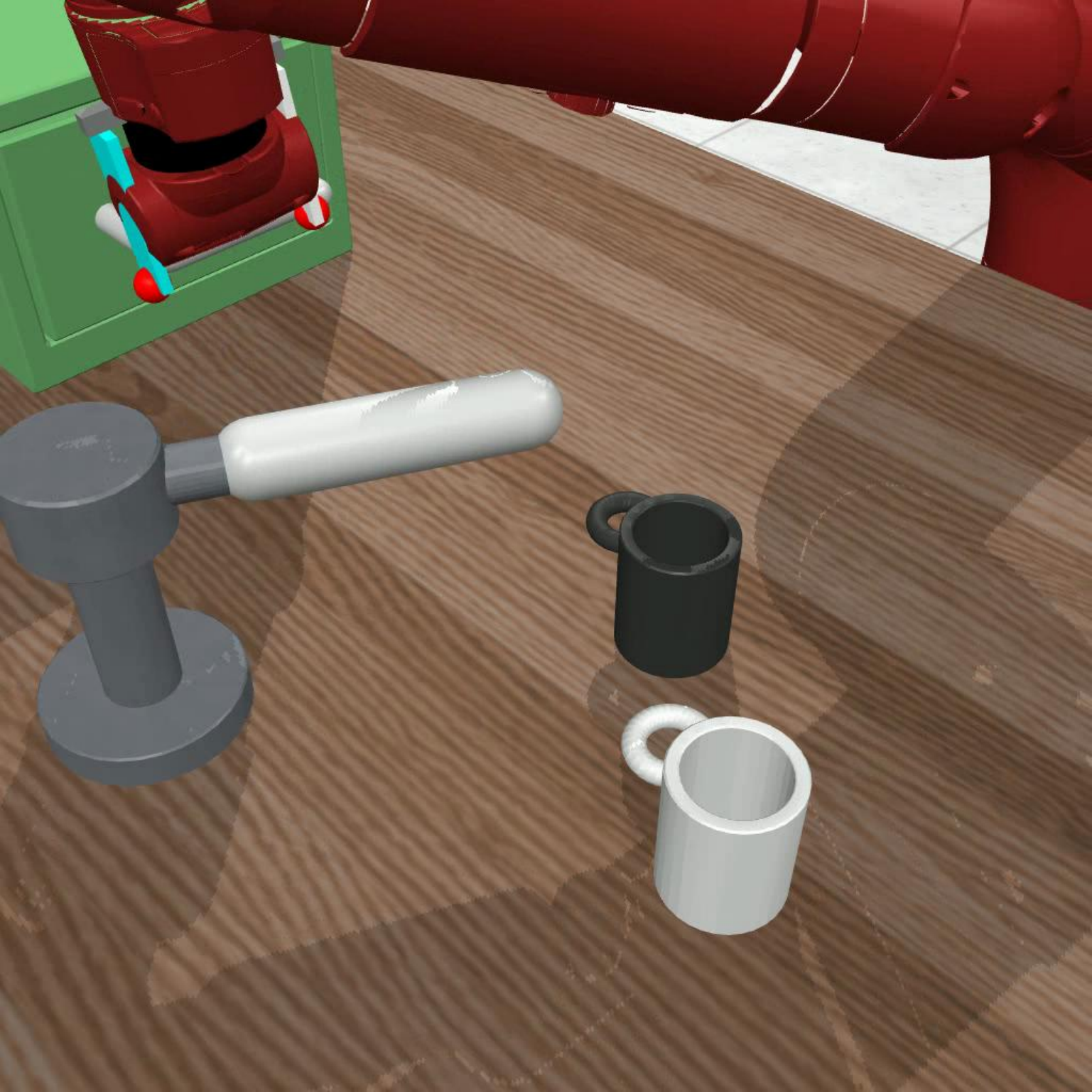}
    \caption{}
  \end{subfigure}
  \caption{LOReL Image: turn faucet right and close drawer}
  \label{fig:turn faucet right and close drawer}
\end{figure*}

Figure~\ref{fig:turn faucet right and close drawer} presents another qualitative execution sequence of the DASL framework on the LOReL Image dataset for the multi-step instruction---``turn faucet right and close drawer''. The panels depict the transition between distinct sub-tasks: panels (a) through (c) show the robot arm interacting with the faucet handle, followed by panels (d) and (e), which display the closing of the drawer. This visual trajectory confirms the ability of DASL to maintain semantic consistency and execution stability across compositional manipulation tasks using purely image-based observations.

\subsection{Additional Experimental Results on Kitchen}
\label{app:Kitchen_result}

Tables  ~\ref{tab:results_kitchen_state} and  ~\ref{tab:results_kitchen_image} present the K-rate evaluations on the LOReL Sawyer dataset, measuring the success rates for completing sequences of up to four sub-tasks under state and image modalities, respectively. DASL demonstrates strong sequential execution capabilities, particularly in the image-based setting where it achieves the highest success rates for the initial stages ($K=1, 2$) of both seen and unseen instructions. These results highlight the robustness of our hierarchical architecture in maintaining semantic alignment throughout multi-step task trajectories.

\begin{table*}[htbp]
    \centering
    \vspace{-0.5em}
    \small 
    \caption{K-rates of different methods on seen and unseen instructions in Kitchen (state)}
    \label{tab:results_kitchen_state}
    
    \newcommand{\hpad}{\hspace{2pt}} 
    \newcommand{\res}[3][]{%
        \def\corecontent{%
            \makebox[1.6em][r]{#2} 
            \scriptsize{\raisebox{.5pt}{$\pm$}} 
            \makebox[1.6em][l]{#3}%
        }%
        \ifx\relax#1\relax \hpad\corecontent\hpad \else #1{\hpad\corecontent\hpad} \fi%
    }
    
    \setlength{\tabcolsep}{3pt}
    \resizebox{\textwidth}{!}{%
    \begin{tabular}{l c c c c c c c c }
        \toprule
        \multirow{2}{*}{\textbf{Methods}} & \multicolumn{4}{c}{\textbf{seen}} & \multicolumn{4}{c}{\textbf{unseen}} \\
        \cmidrule(lr){2-5} \cmidrule(lr){6-9}
         & 1 & 2 & 3 & 4 & 1 & 2 & 3 & 4 \\
        \midrule
        Lang DT & \res{83.1}{3.2} & \res{40.0}{7.4} & \res{12.6}{4.7} & \res{3.0}{1.6} & \res{83.3}{6.4} & \res{40.9}{20.8} & \res{1.1}{0.8} & \res{0.0}{0.0} \\
        LISA    & \res{52.5}{28.1}  & \res{16.9}{14.4}  & \res{1.6}{2.1}  & \res{0.0}{0.0} & \res{51.1}{33.0}  & \res{19.8}{19.4} & \res{0.0}{0.0} & \res{0.0}{0.0} \\
        LADS    & \res[\textbf]{98.4}{1.1} & \res[\textbf]{79.9}{5.8} & \res[\textbf]{39.3}{5.9} & \res[\textbf]{6.9}{1.1} & \res[\textbf]{99.8}{0.4}  & \res[\textbf]{80.9}{7.4} & \res[\textbf]{41.8}{7.7} & \res[\textbf]{0.9}{1.0} \\
        DASL    & \res{97.3}{1.8}  & \res{58.7}{1.1}  & \res{9.7}{1.9}  & \res{0.0}{0.0} & \res{97.8}{3.9}  & \res{77.8}{15.4} & \res{0.0}{0.0} & \res{0.0}{0.0} \\
        \bottomrule
    \end{tabular}%
    }
\end{table*}

\begin{table*}[htbp]
    \centering
    \vspace{-1em}
    \small 
    \caption{K-rates of different methods on seen and unseen instructions in Kitchen (image)}
    \label{tab:results_kitchen_image}
    
    \newcommand{\hpad}{\hspace{2pt}} 
    \newcommand{\res}[3][]{%
        \def\corecontent{%
            \makebox[1.6em][r]{#2} 
            \scriptsize{\raisebox{.5pt}{$\pm$}} 
            \makebox[1.6em][l]{#3}%
        }%
        \ifx\relax#1\relax \hpad\corecontent\hpad \else #1{\hpad\corecontent\hpad} \fi%
    }
    
    \setlength{\tabcolsep}{3pt}
    \resizebox{\textwidth}{!}{%
    \begin{tabular}{l c c c c c c c c }
        \toprule
        \multirow{2}{*}{\textbf{Methods}} & \multicolumn{4}{c}{\textbf{seen}} & \multicolumn{4}{c}{\textbf{unseen}} \\
        \cmidrule(lr){2-5} \cmidrule(lr){6-9}
         & 1 & 2 & 3 & 4 & 1 & 2 & 3 & 4 \\
        \midrule
        Lang DT & \res{73.9}{20.4} & \res{42.7}{14.5} & \res{12.9}{5.1} & \res{2.0}{0.9} & \res{78.0}{12.7} & \res{25.6}{18.0} & \res{2.9}{2.1} & \res{0.0}{0.0} \\
        LISA    & \res{62.3}{8.6}  & \res{20.8}{3.2}  & \res{3.0}{0.9}  & \res{0.0}{0.0} & \res{88.9}{7.4}  & \res{34.4}{11.7} & \res{5.8}{2.1} & \res{0.2}{0.4} \\
        LADS    & \res{90.6}{10.3} & \res{55.5}{17.5} & \res{14.3}{4.9} & \res[\textbf]{4.1}{4.4} & \res{91.6}{7.8}  & \res{47.3}{18.3} & \res[\textbf]{4.9}{1.5} & \res[\textbf]{0.4}{0.4} \\
        DASL    & \res[\textbf]{94.6}{1.8}  & \res[\textbf]{60.0}{5.1}  & \res[\textbf]{17.9}{4.2}  & \res{0.30}{0.53} & \res[\textbf]{95.6}{7.7}  & \res[\textbf]{71.1}{3.9} & \res{4.44}{7.7} & \res{0.0}{0.0} \\
        \bottomrule
    \end{tabular}%
    }
\end{table*}

\begin{figure}[H]
  \centering
  \begin{subfigure}[t]{0.14\textwidth}
    \includegraphics[width=\linewidth]{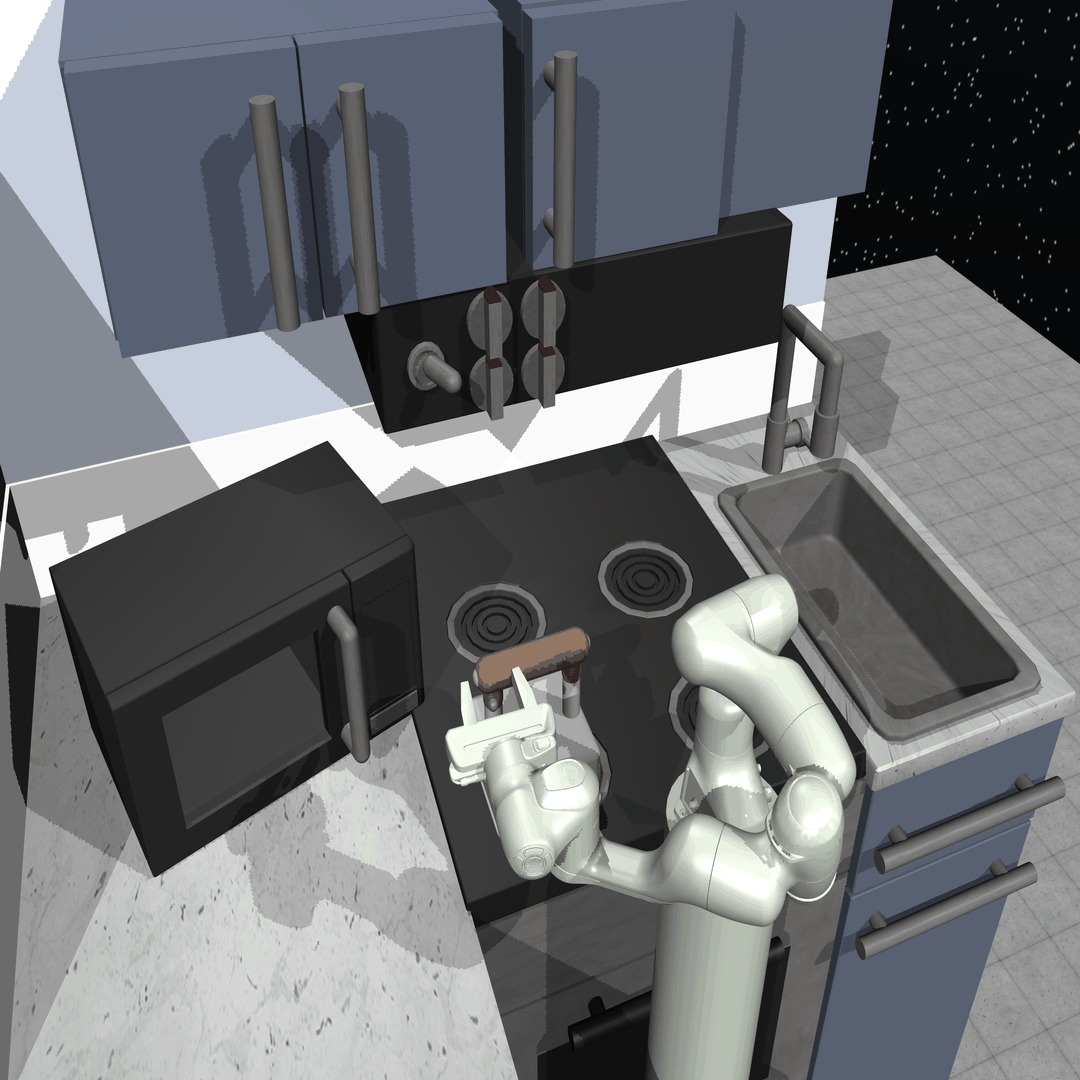}
    \caption{}
  \end{subfigure}\hfill
  \begin{subfigure}[t]{0.14\textwidth}
    \includegraphics[width=\linewidth]{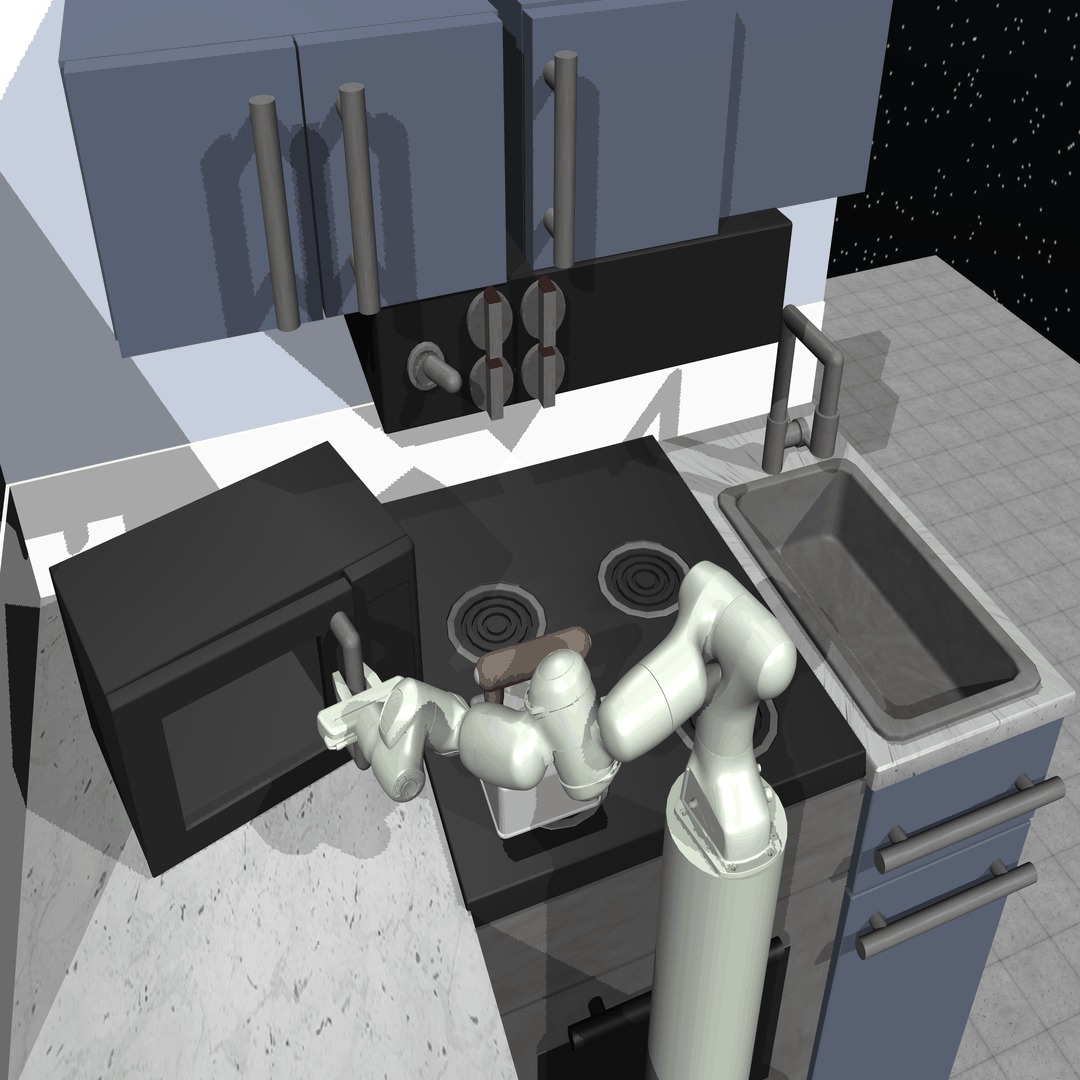}
    \caption{}
  \end{subfigure}\hfill
  \begin{subfigure}[t]{0.14\textwidth}
    \includegraphics[width=\linewidth]{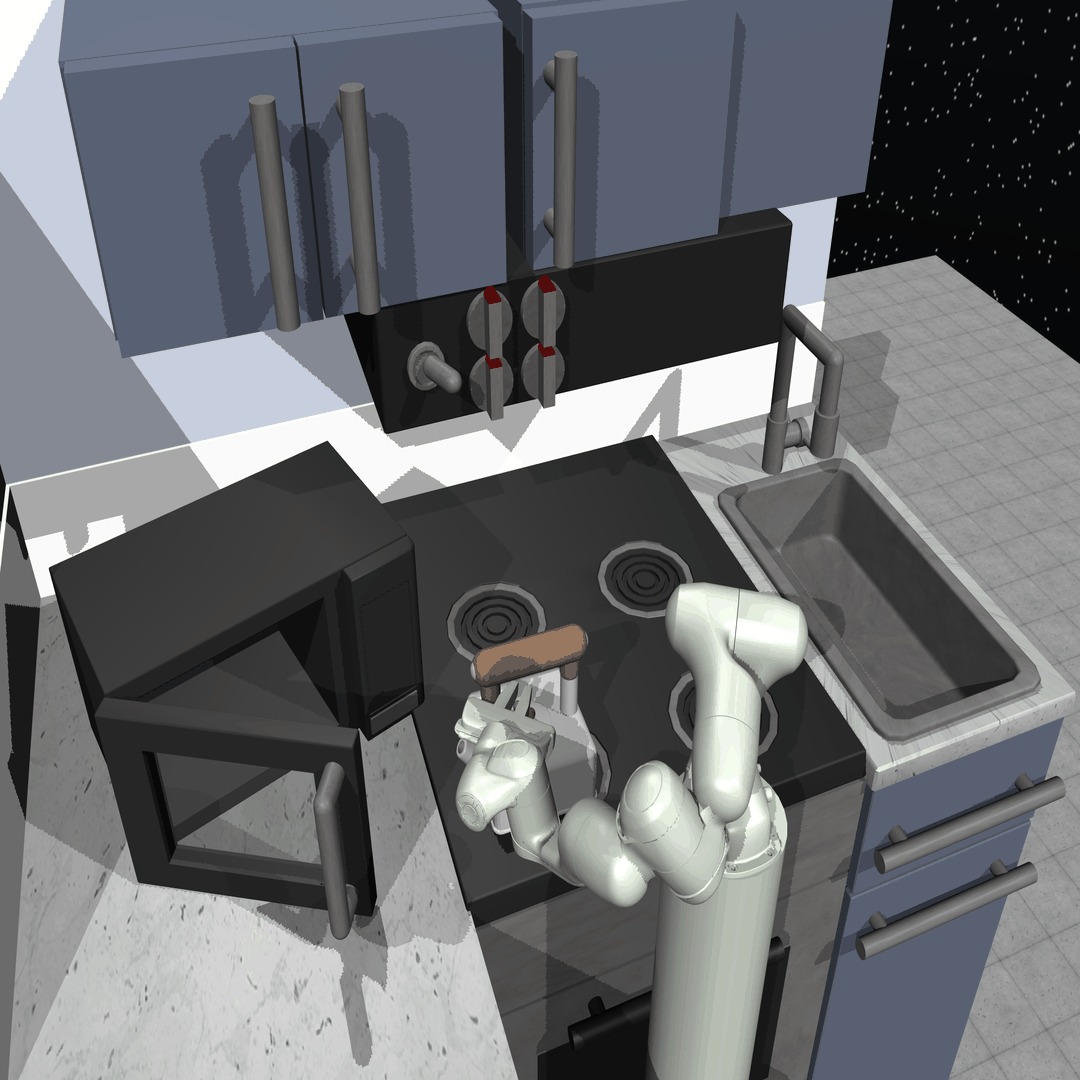}
    \caption{}
  \end{subfigure}\hfill
  \begin{subfigure}[t]{0.14\textwidth}
    \includegraphics[width=\linewidth]{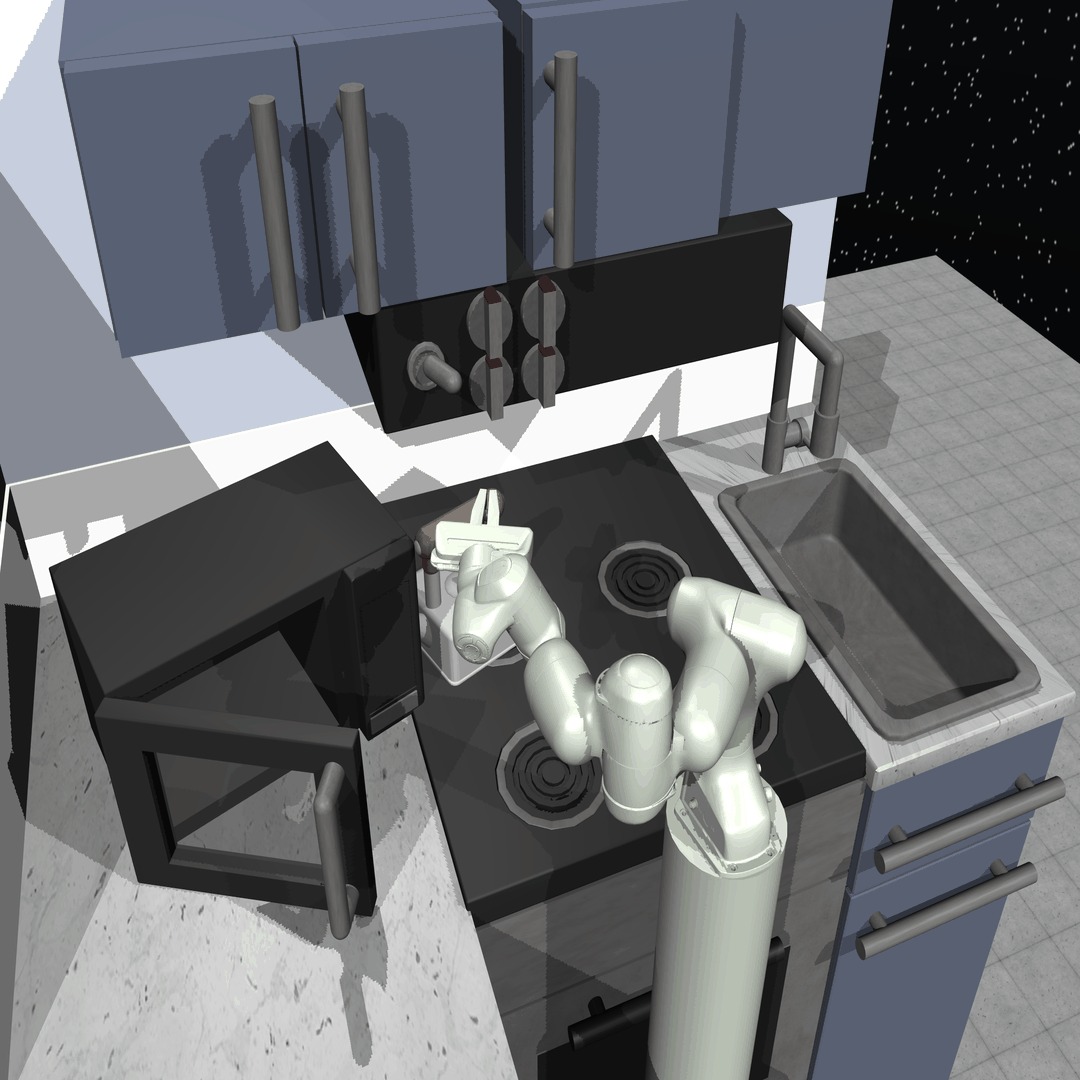}
    \caption{}
  \end{subfigure}\hfill
  \begin{subfigure}[t]{0.14\textwidth}
    \includegraphics[width=\linewidth]{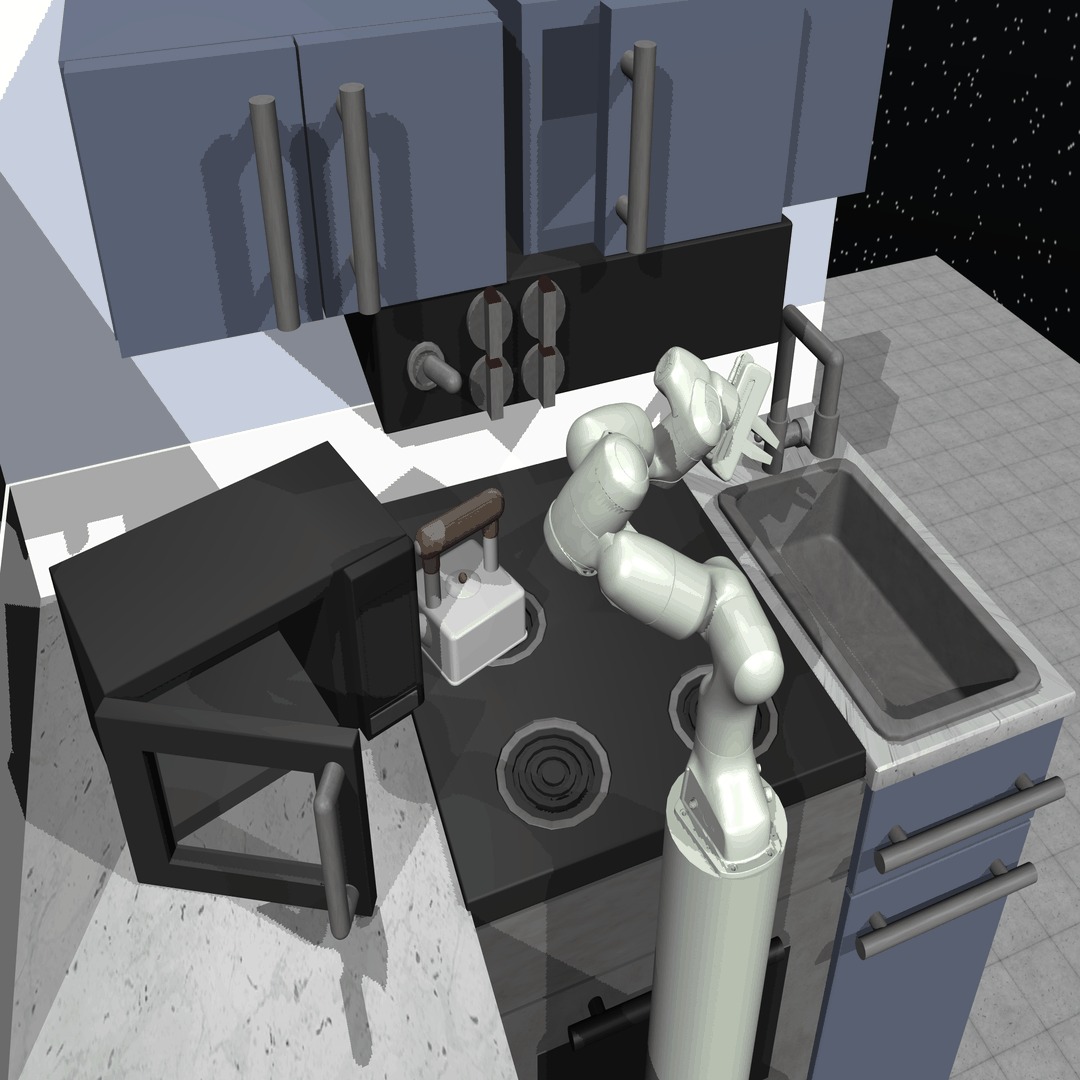}
    \caption{}
  \end{subfigure}\hfill
  \begin{subfigure}[t]{0.14\textwidth}
    \includegraphics[width=\linewidth]{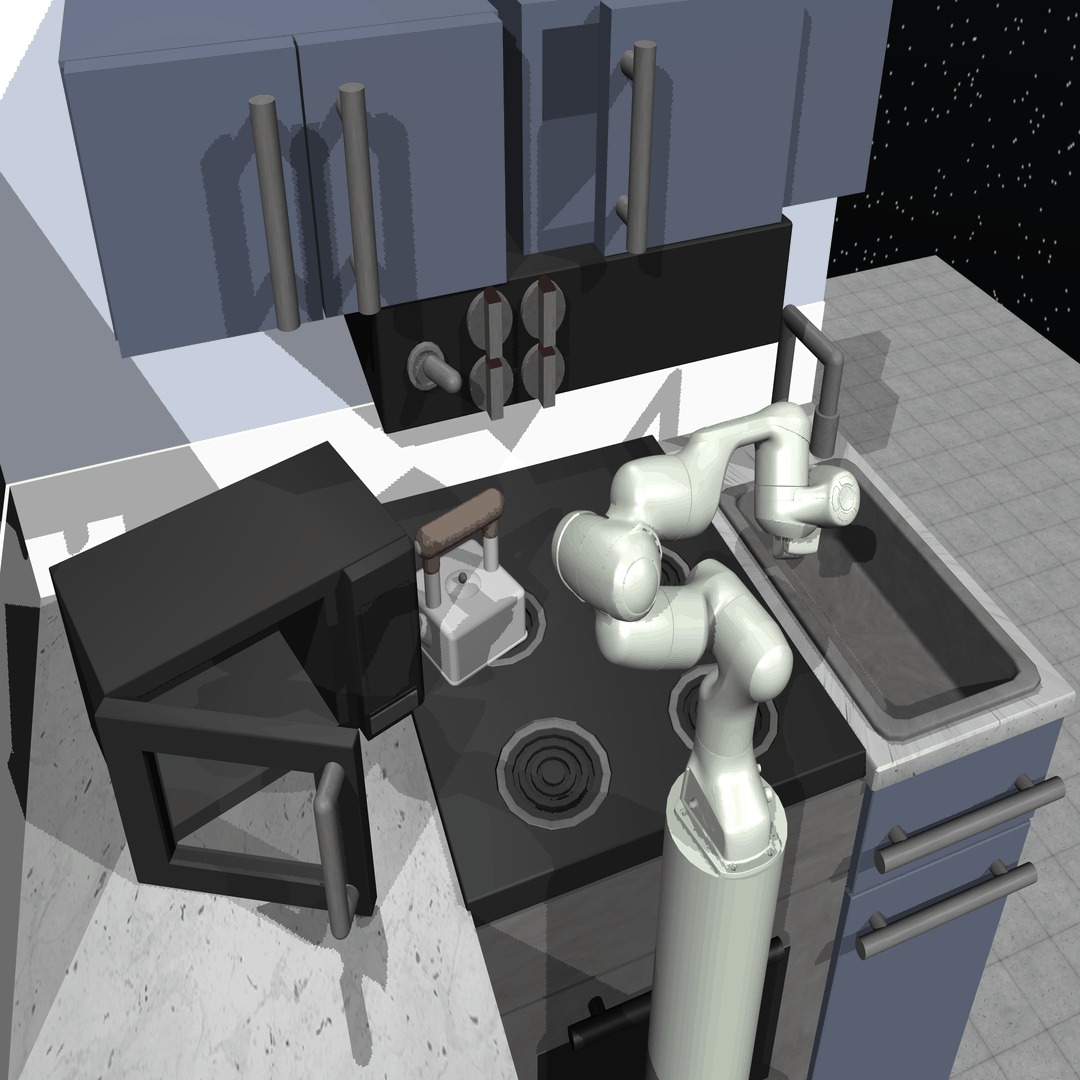}
    \caption{}
  \end{subfigure}\hfill
  \begin{subfigure}[t]{0.14\textwidth}
    \includegraphics[width=\linewidth]{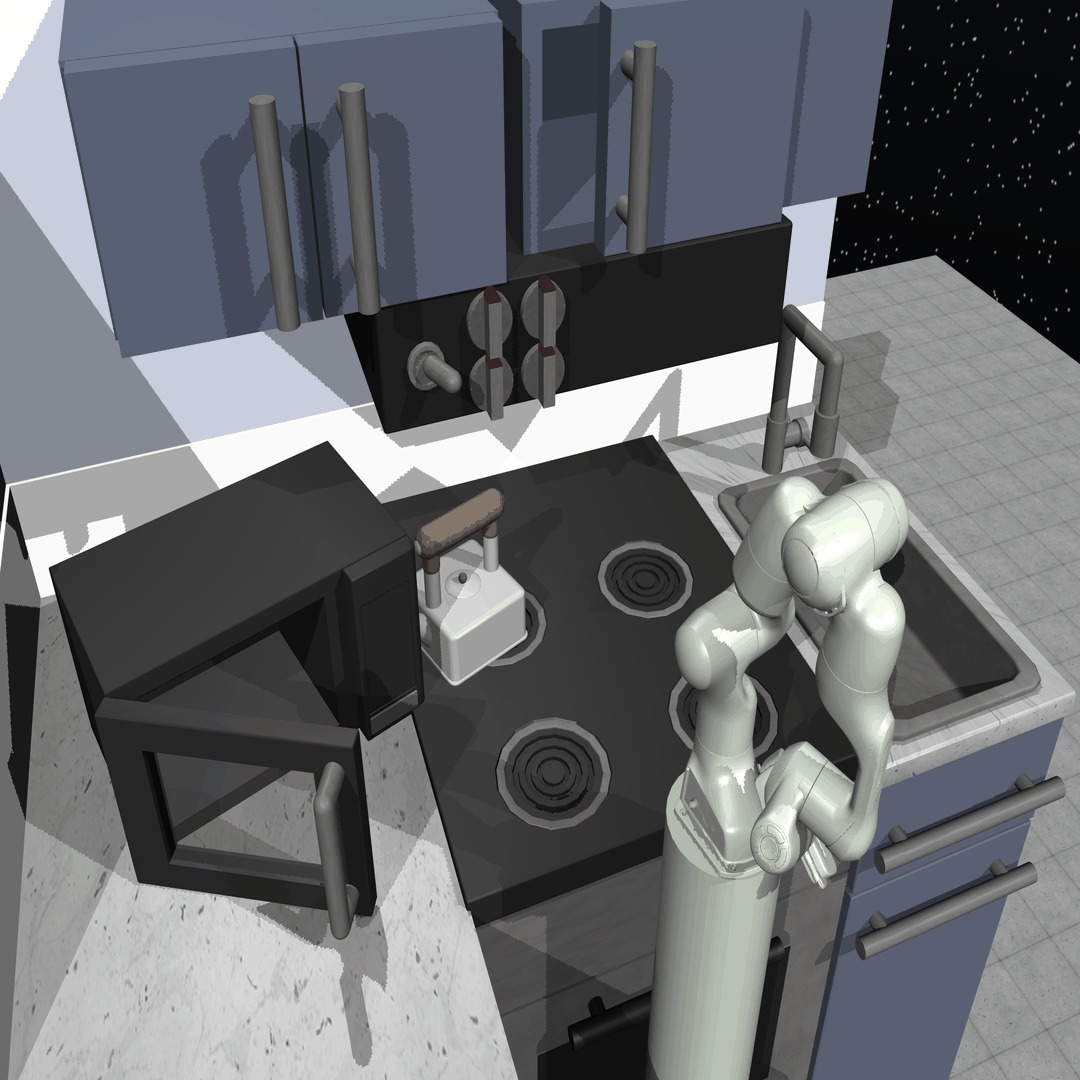}
    \caption{}
  \end{subfigure}
  \caption{Kitchen State: activate bottom burner and activate top burner and turn on light switch and open sliding cabinet}
  \label{fig:activate bottom burner and activate top burner and turn on light switch and open sliding cabinet_state}
  \vspace{-1em} 
\end{figure}

Figure ~\ref{fig:activate bottom burner and activate top burner and turn on light switch and open sliding cabinet_state} presents a qualitative execution sequence in the Kitchen (state), where the DASL agent successfully performs a complex, compositional task comprising four sequential sub-tasks: activating the bottom burner, activating the top burner, turning on the light switch, and opening the sliding cabinet. Panels (a) through (g) illustrate the robot arm's precise manipulation and smooth transitions between diverse interaction points within the high-dimensional kitchen workspace. This successful execution of a multi-stage trajectory highlights the framework's robustness in maintaining long-term semantic alignment and stable control in intricate, -inspired environments.

\begin{figure*}[htbp]
  \centering
  \begin{subfigure}[t]{0.14\textwidth}
    \includegraphics[width=\linewidth]{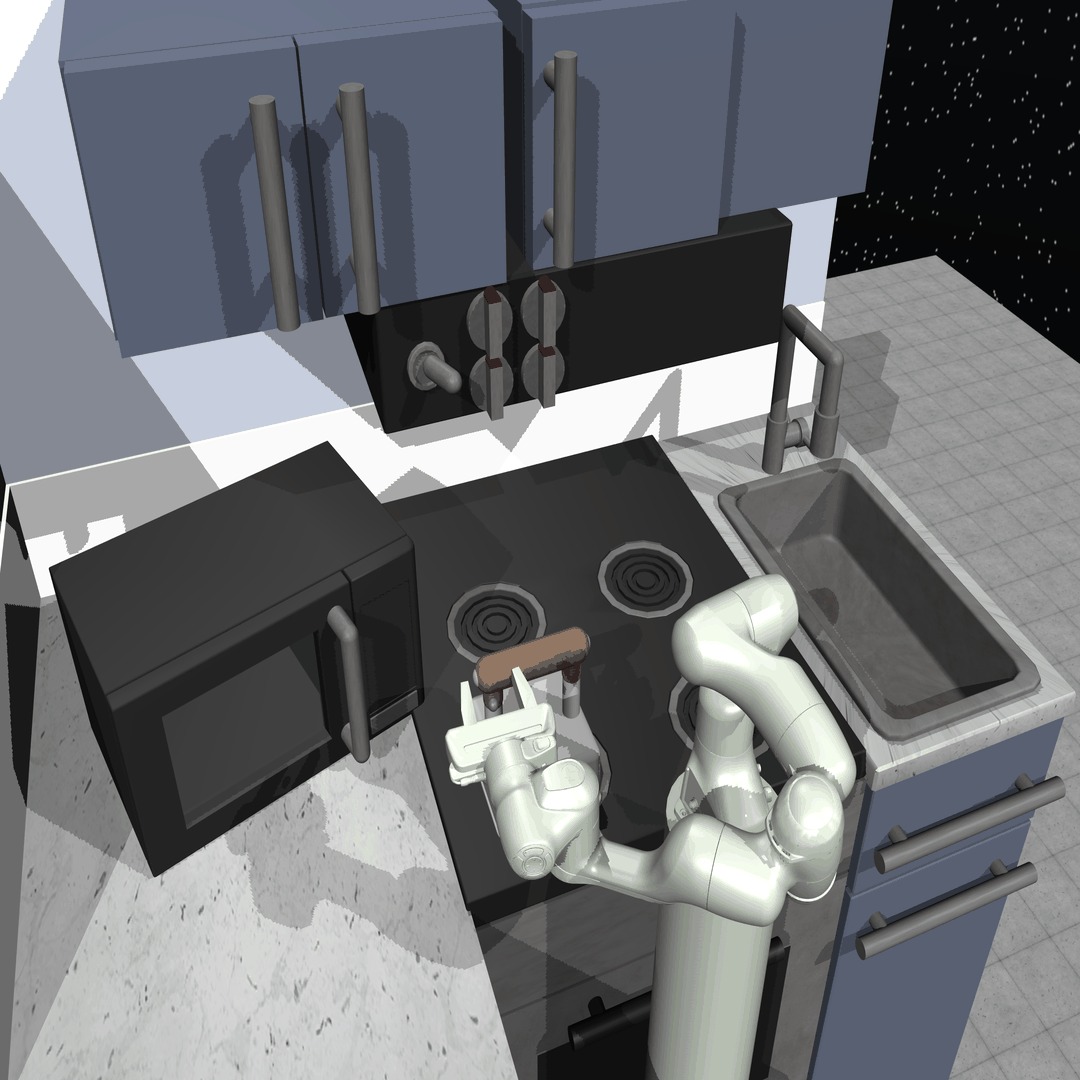}
    \caption{}
  \end{subfigure}\hfill
  \begin{subfigure}[t]{0.14\textwidth}
    \includegraphics[width=\linewidth]{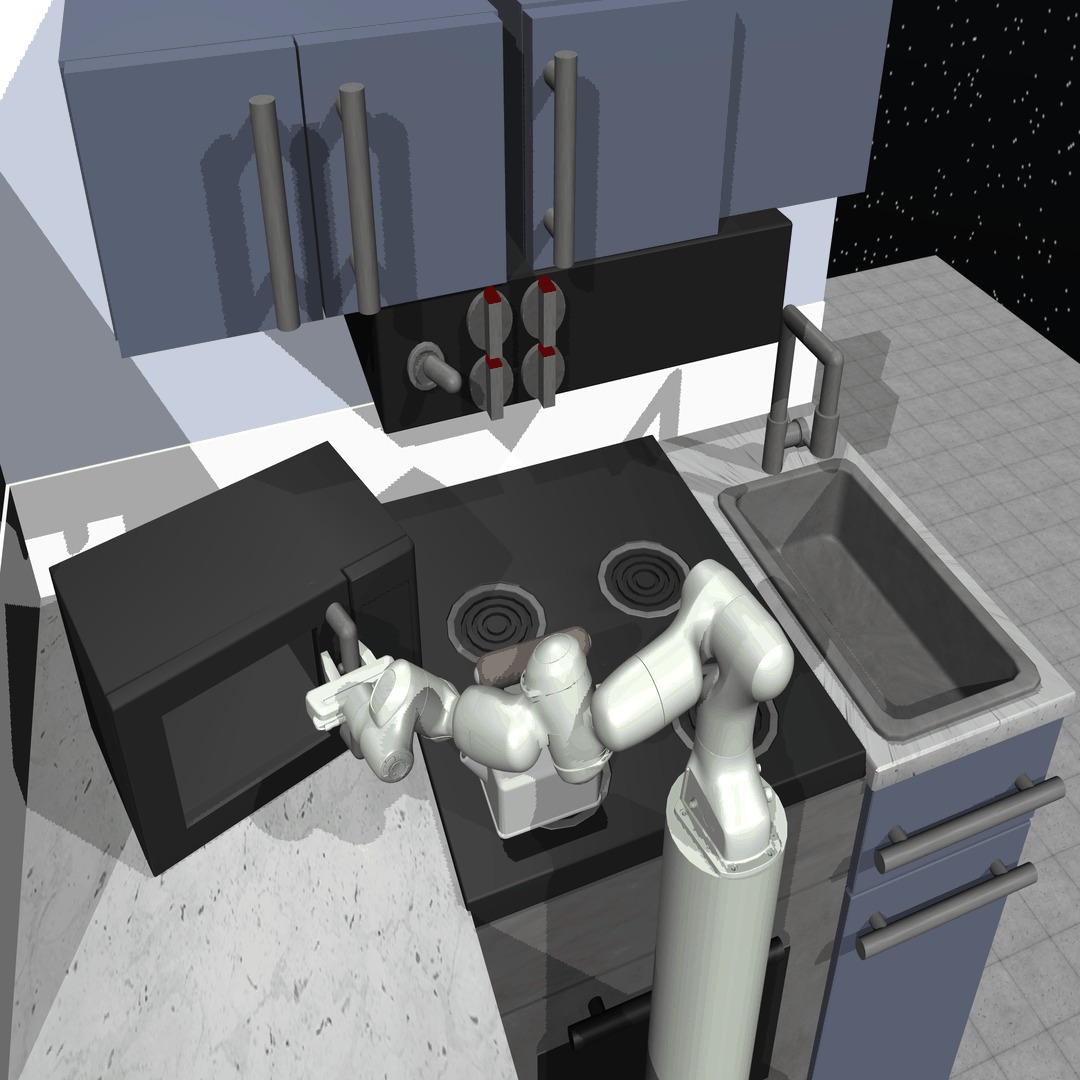}
    \caption{}
  \end{subfigure}\hfill
  \begin{subfigure}[t]{0.14\textwidth}
    \includegraphics[width=\linewidth]{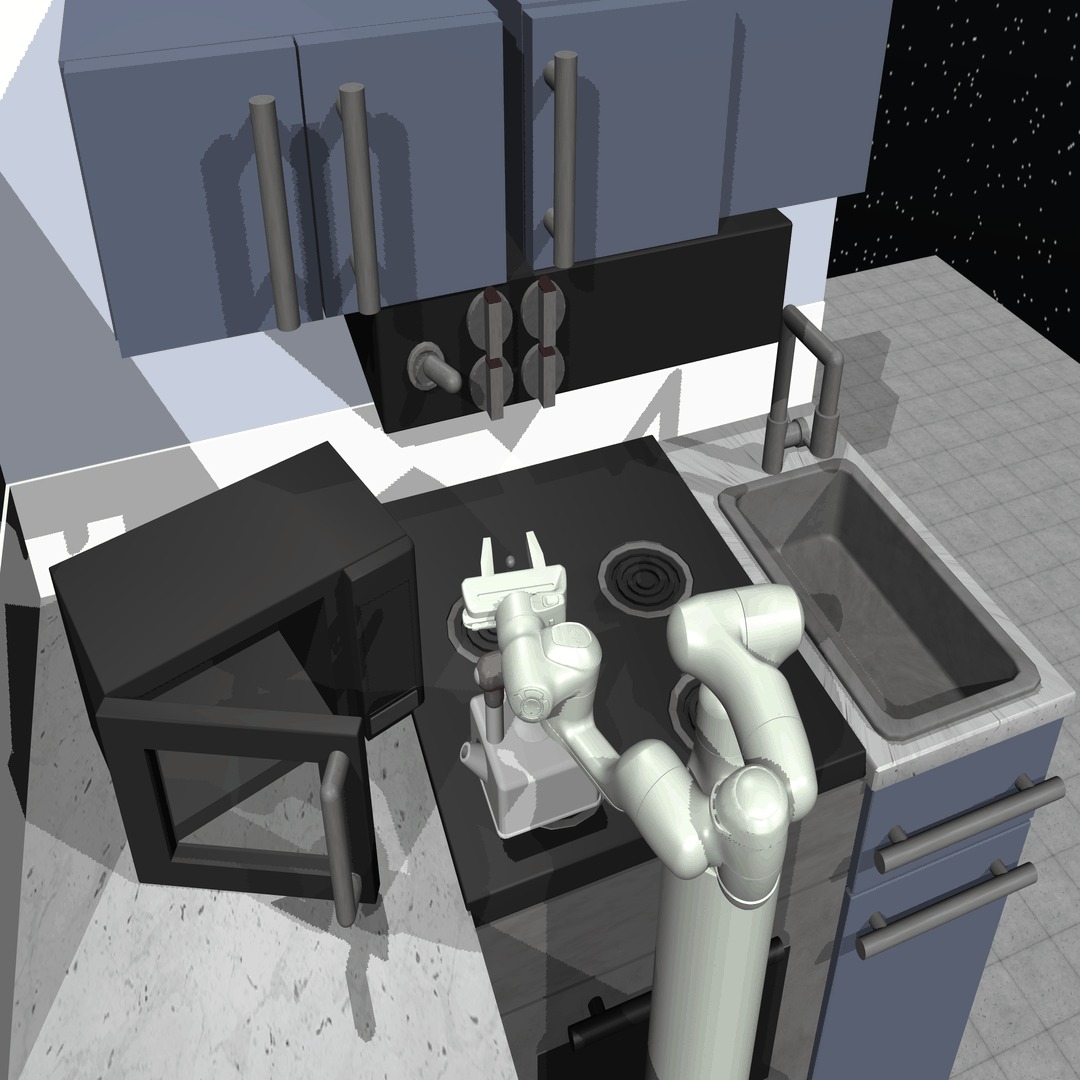}
    \caption{}
  \end{subfigure}\hfill
  \begin{subfigure}[t]{0.14\textwidth}
    \includegraphics[width=\linewidth]{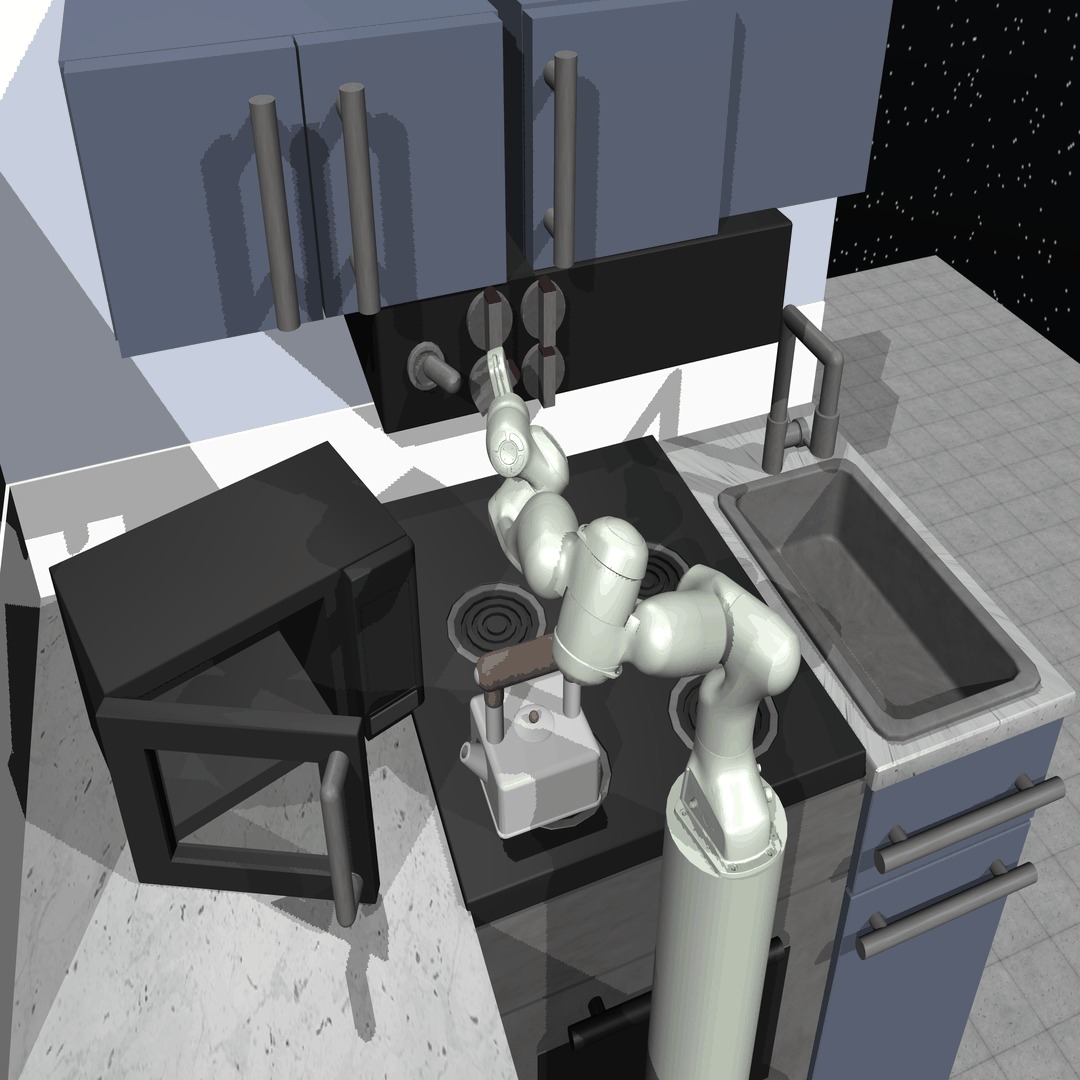}
    \caption{}
  \end{subfigure}\hfill
  \begin{subfigure}[t]{0.14\textwidth}
    \includegraphics[width=\linewidth]{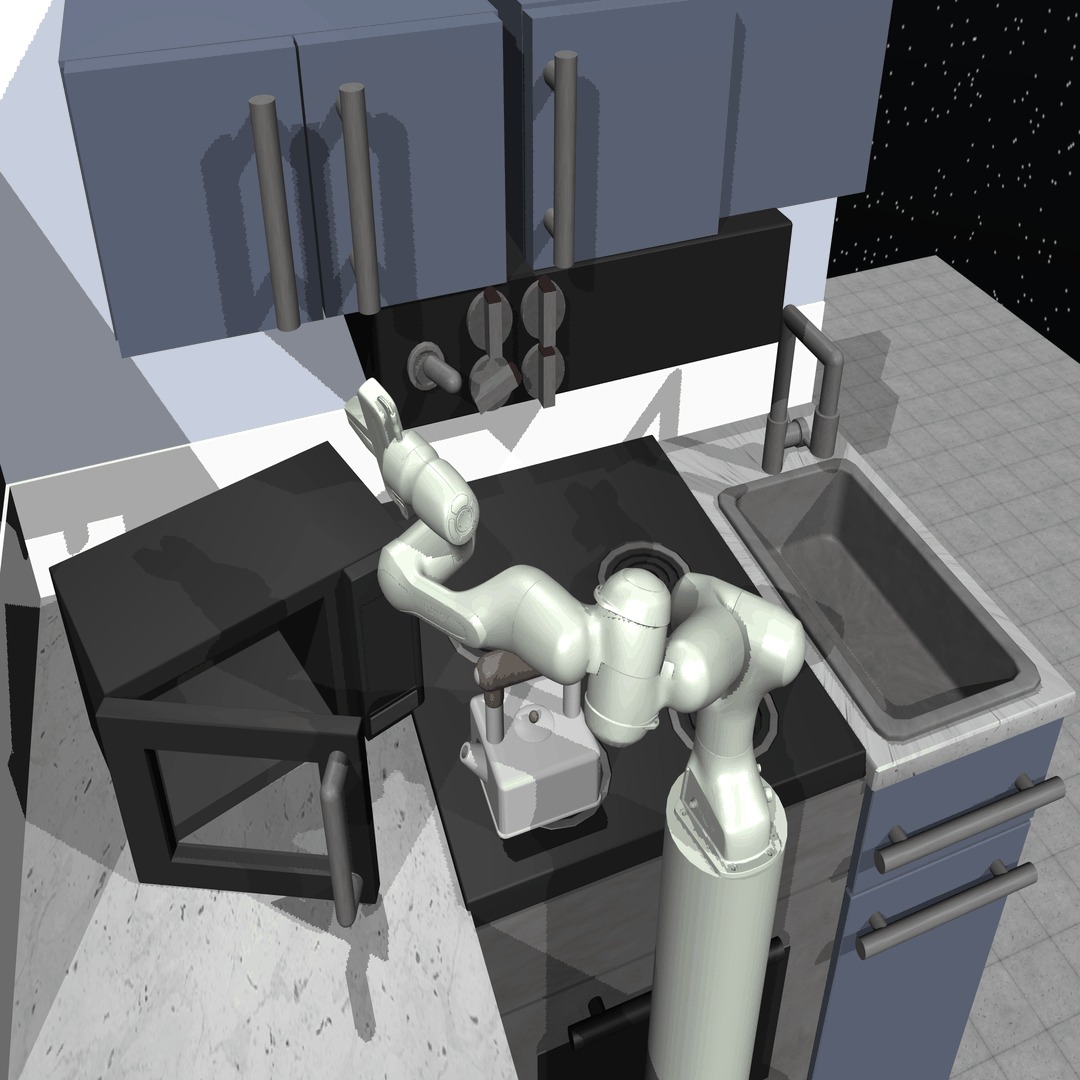}
    \caption{}
  \end{subfigure}\hfill
  \begin{subfigure}[t]{0.14\textwidth}
    \includegraphics[width=\linewidth]{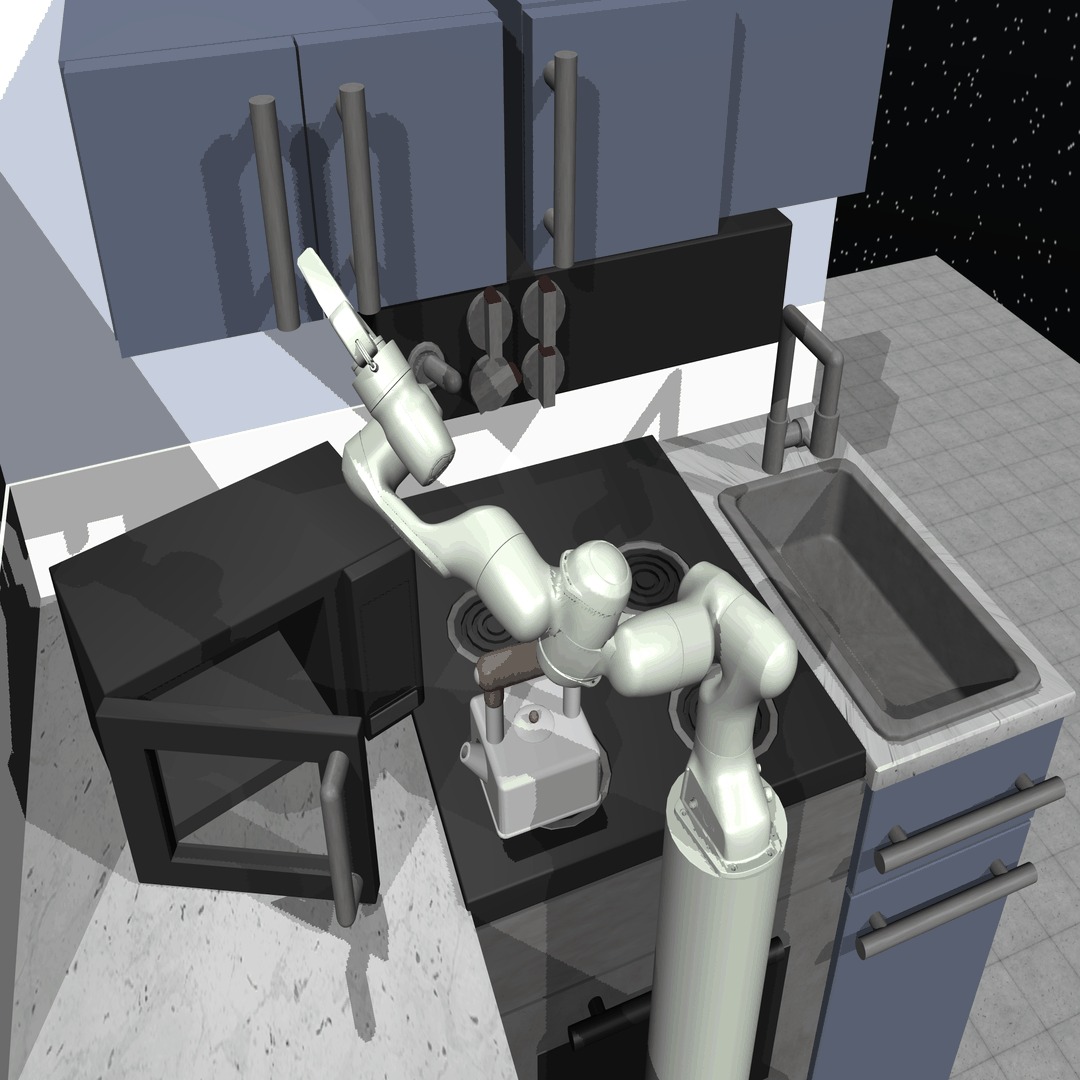}
    \caption{}
  \end{subfigure}\hfill
  \begin{subfigure}[t]{0.14\textwidth}
    \includegraphics[width=\linewidth]{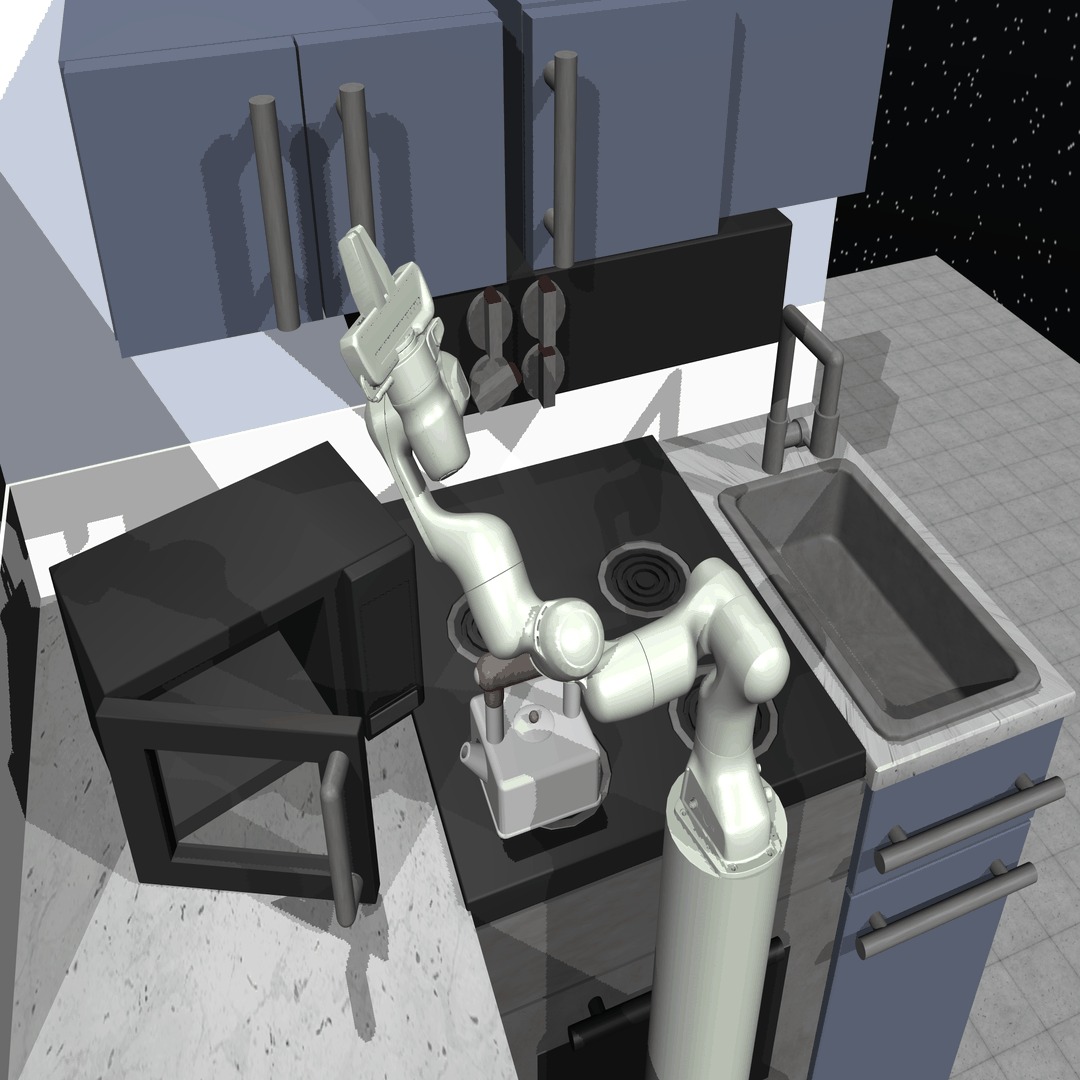}
    \caption{}
  \end{subfigure}
  \caption{Kitchen Image: activate bottom burner and activate top burner and turn on light switch and open sliding cabinet}
  \label{fig:activate bottom burner and activate top burner and turn on light switch and open sliding cabinet_image}
\end{figure*}

Figure ~\ref{fig:activate bottom burner and activate top burner and turn on light switch and open sliding cabinet_image} illustrates a qualitative execution sequence of the DASL framework in the Kitchen (image). The sequence showcases the agent's successful completion of a complex compositional task consisting of four sub-tasks: activating the bottom burner, activating the top burner, turning on the light switch, and opening the sliding cabinet. Panels (a) through (g) highlight the agent's ability to maintain semantic alignment and precise control across multiple manipulation points while operating directly from visual inputs in a high-dimensional workspace.

\subsection{Additional Experimental Results on CALVIN}
\label{app:CALVIN_result}

\begin{figure}[H]
  \centering
  \begin{subfigure}[t]{0.14\textwidth}
    \includegraphics[width=\linewidth]{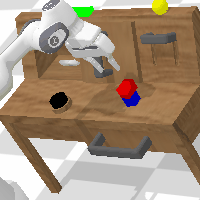}
    \caption{}
  \end{subfigure}\hfill
  \begin{subfigure}[t]{0.14\textwidth}
    \includegraphics[width=\linewidth]{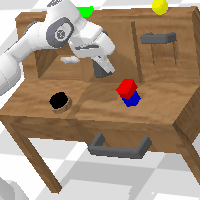}
    \caption{}
  \end{subfigure}\hfill
  \begin{subfigure}[t]{0.14\textwidth}
    \includegraphics[width=\linewidth]{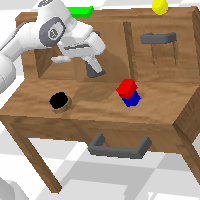}
    \caption{}
  \end{subfigure}\hfill
  \begin{subfigure}[t]{0.14\textwidth}
    \includegraphics[width=\linewidth]{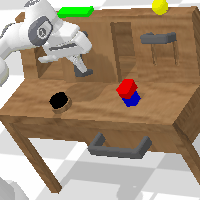}
    \caption{}
  \end{subfigure}\hfill
  \begin{subfigure}[t]{0.14\textwidth}
    \includegraphics[width=\linewidth]{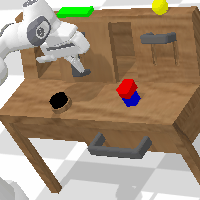}
    \caption{}
  \end{subfigure}\hfill
  \begin{subfigure}[t]{0.14\textwidth}
    \includegraphics[width=\linewidth]{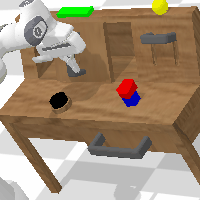}
    \caption{}
  \end{subfigure}\hfill
  \begin{subfigure}[t]{0.14\textwidth}
    \includegraphics[width=\linewidth]{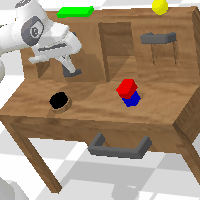}
    \caption{}
  \end{subfigure}
  \caption{CALVIN: push the sliding door to the left side}
  \label{fig:calvin_move_slider_left}
  \vspace{-1em} 
\end{figure}

Figure~\ref{fig:calvin_move_slider_left} shows a qualitative execution sequence in the CALVIN environment, where the DASL completes a language-conditioned task: moving the green slider to the left. Panels (a) through (g) capture the continuous motion of the robotic arm as it interacts with the articulated object on the tabletop. The successful completion of this fine-grained trajectory demonstrates the capability of the framework to maintain accurate spatial alignment and stable low-level control in contact-rich simulations.

\begin{figure}[H]
  \centering
  \begin{subfigure}[t]{0.14\textwidth}
    \includegraphics[width=\linewidth]{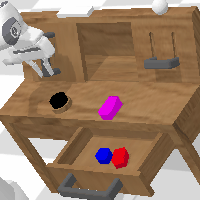}
    \caption{}
  \end{subfigure}\hfill
  \begin{subfigure}[t]{0.14\textwidth}
    \includegraphics[width=\linewidth]{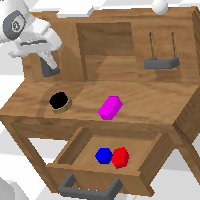}
    \caption{}
  \end{subfigure}\hfill
  \begin{subfigure}[t]{0.14\textwidth}
    \includegraphics[width=\linewidth]{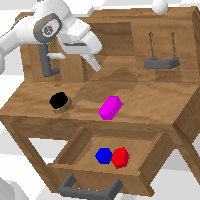}
    \caption{}
  \end{subfigure}\hfill
  \begin{subfigure}[t]{0.14\textwidth}
    \includegraphics[width=\linewidth]{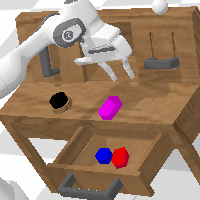}
    \caption{}
  \end{subfigure}\hfill
  \begin{subfigure}[t]{0.14\textwidth}
    \includegraphics[width=\linewidth]{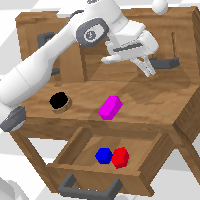}
    \caption{}
  \end{subfigure}\hfill
  \begin{subfigure}[t]{0.14\textwidth}
    \includegraphics[width=\linewidth]{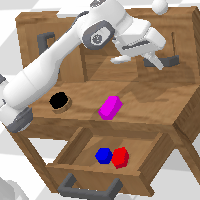}
    \caption{}
  \end{subfigure}\hfill
  \begin{subfigure}[t]{0.14\textwidth}
    \includegraphics[width=\linewidth]{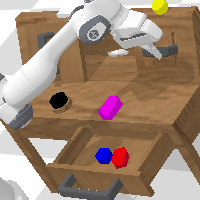}
    \caption{}
  \end{subfigure}
  \caption{CALVIN: use the switch to turn on the light bulb}
  \label{fig:calvin_turn_on_lightbulb}
  \vspace{-1em} 
\end{figure}

Figure~\ref{fig:calvin_turn_on_lightbulb} shows a qualitative execution sequence in the CALVIN environment, where the DASL agent completes a language-conditioned task: turning on the lightbulb. Panels (a) through (g) capture the continuous motion of the robotic arm as it interacts with the light switch on the tabletop. The successful completion of this trajectory demonstrates the capability of the framework to maintain accurate spatial alignment and stable low-level control during contact-rich physical interactions.

\subsection{Visualization.}
\label{app:skill_visuals}

\textbf{Skill Heat Map and Codebook Collapse.} Figure \ref{fig:codebook_collapse} reveals a clear contrast in how DASL and LISA utilize the discrete latent space. LISA exhibits severe index collapse, with word-skill activations concentrated on a few indices like 1 and 7, while the rest of the codebook remains dormant. Conversely, DASL produces a more distributed and sparse activation pattern across the 20 initialized indices. By successfully identifying 14 distinct atomic skills, DASL demonstrates much higher codebook efficiency. This sparsity suggests that the model effectively disentangles latent representations, grounding high-level instructions into semantically explicit atomic skills. Such representational diversity is essential for an interpretable mapping between linguistic tokens and robot actions.

\textbf{Word Clouds.} Figure~\ref{fig:word_clouds} further presents the word clouds of the atomic skills learned by LOReL on the Sawyer (state) compositional tasks. The results demonstrate that the model successfully acquired 14 distinct atomic skills, each exhibiting a strong correlation with specific task semantics. Notably, these skills display significant robustness to ambiguous language instructions. For instance, \textbf{Skill 5} shows high consistency with the token `mug', effectively encoding the action of 'move white mug right'. Similarly, \textbf{Skill 1} corresponds to the skill of `turn faucet right'; it not only aligns with the tokens `turn' and `right' but also demonstrates the ability to successfully reconstruct fragmented sub-word tokens (e.g., `fa', `\#\#uce', `\#\#t') into a coherent, physically meaningful operation. This highlights our method's capability to accurately interpret and execute complex language instructions.

\begin{figure}[htbp]
  \centering
  \includegraphics[width=\textwidth]{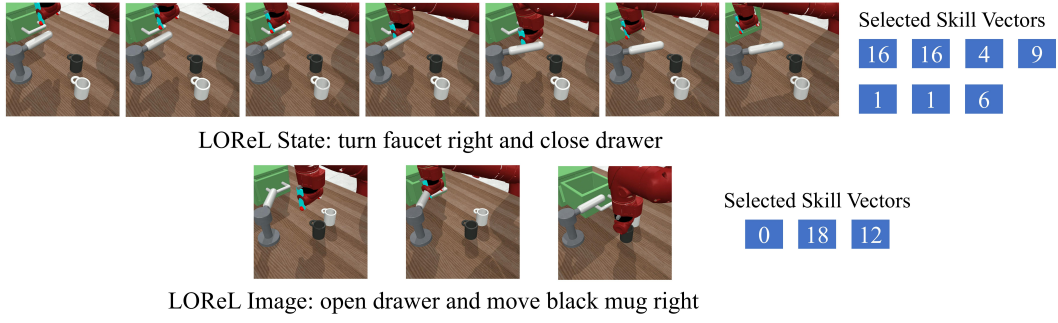}
  \caption{Visualize two LOReL episodes with aligned keyframes and the discrete skill index selected at each timestep. }
  \label{fig:choose_skill}
\end{figure}

\textbf{Skill Selection.} Visual analysis of Figure \ref{fig:choose_skill} confirms the capability of the hierarchical framework to perform autonomous task decomposition across varying time horizons. For the instruction turn faucet right and close drawer, the model invokes a sequence of discrete skill indices [16, 16, 4, 9, 1, 1, 6], where transitions in the latent indices directly correspond to the physical shift from faucet manipulation to drawer closure. A similar adaptive behavior is observed in the second episode with the sequence [0, 18, 12] for the task of opening the drawer and moving the mug. This precise mapping between temporal sub-goals and discrete skill vectors suggests that the learned codebook effectively encapsulates the prerequisite atomic actions required for multi-stage manipulation, ensuring semantic consistency throughout the trajectory.

\begin{figure*}[htbp]
  \centering
  \setlength{\tabcolsep}{0pt} %
  \renewcommand{\arraystretch}{0.5} 

  \resizebox{\textwidth}{!}{%
  \begin{tabular}{@{}cc@{}} %

    \includegraphics[width=0.495\linewidth]{fig/eval_comp_0329_235408_correlation_matrix.pdf} &
    \includegraphics[width=0.495\linewidth]{fig/eval_comp_correlation_matrix.pdf} \\

    \includegraphics[width=0.495\linewidth]{fig/eval_comp_0329_235408_option_freq_matrix.pdf} &
    \includegraphics[width=0.495\linewidth]{fig/eval_comp_option_freq_matrix.pdf} \\

    \includegraphics[width=0.495\linewidth]{fig/eval_comp_0329_235408_word_freq_matrix.pdf} &
    \includegraphics[width=0.495\linewidth]{fig/eval_comp_word_freq_matrix.pdf} \\

    \multicolumn{1}{c}{\small \textbf{(a) LISA}} &
    \multicolumn{1}{c}{\small \textbf{(b) DASL (Ours)}} \\
    
  \end{tabular}%
  }

\caption{Qualitative comparison of learned skill representations for LISA and DASL on LOReL Sawyer tasks. The rows display code correlation (Word $\times$ Skill), option frequency (normalized row-wise), and word frequency (normalized column-wise) for a size-20 skill set. Column (a) illustrates significant index collapse in LISA, where activations are limited to a small subset of indices like 1 and 7. In contrast, the DASL heatmaps in column (b) show marked sparsity and distinct activation regions across the codebook. These disentangled associations between linguistic tokens and atomic skills highlight the superior interpretability and utilization of our framework. (Zoom in for best view).}
  \label{fig:codebook_collapse}
\end{figure*}

\begin{figure}[htbp]  
    \begin{center}
    \includegraphics[width=1\linewidth]{fig/eval_comp_wordclouds_combined.pdf}
    \caption{Word cloud of skills learned in LOReL Sawyer (state) compositional tasks.}
    \label{fig:word_clouds}
    \end{center}
\end{figure}

\begin{figure*}[htbp]
  \centering
  \setlength{\tabcolsep}{0pt} %
  \renewcommand{\arraystretch}{1.0} %

  \resizebox{\textwidth}{!}{%
  \begin{tabular}{@{}ccc@{}} %

    \includegraphics[height=0.28\textwidth]{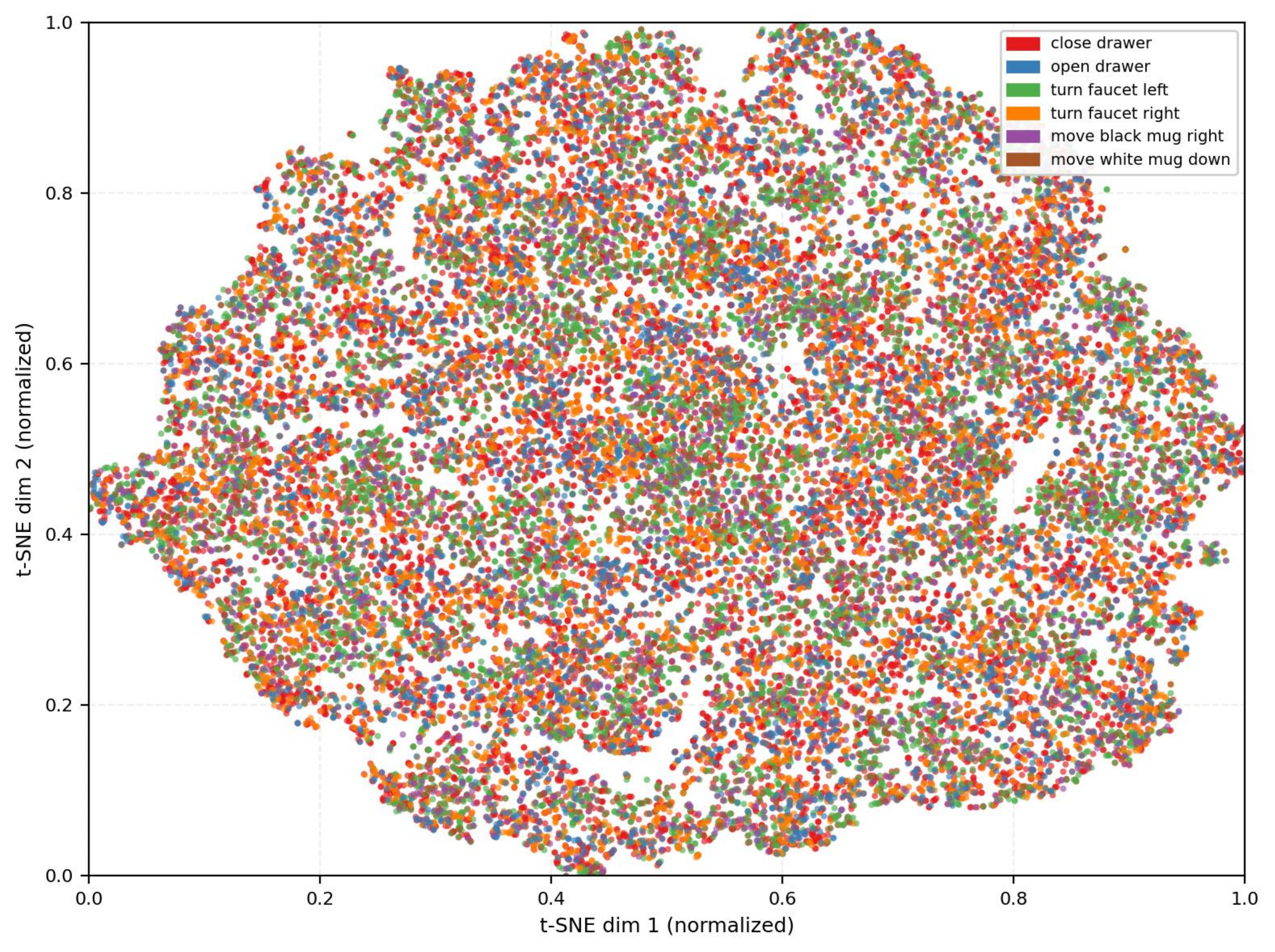} &
    \includegraphics[height=0.28\textwidth]{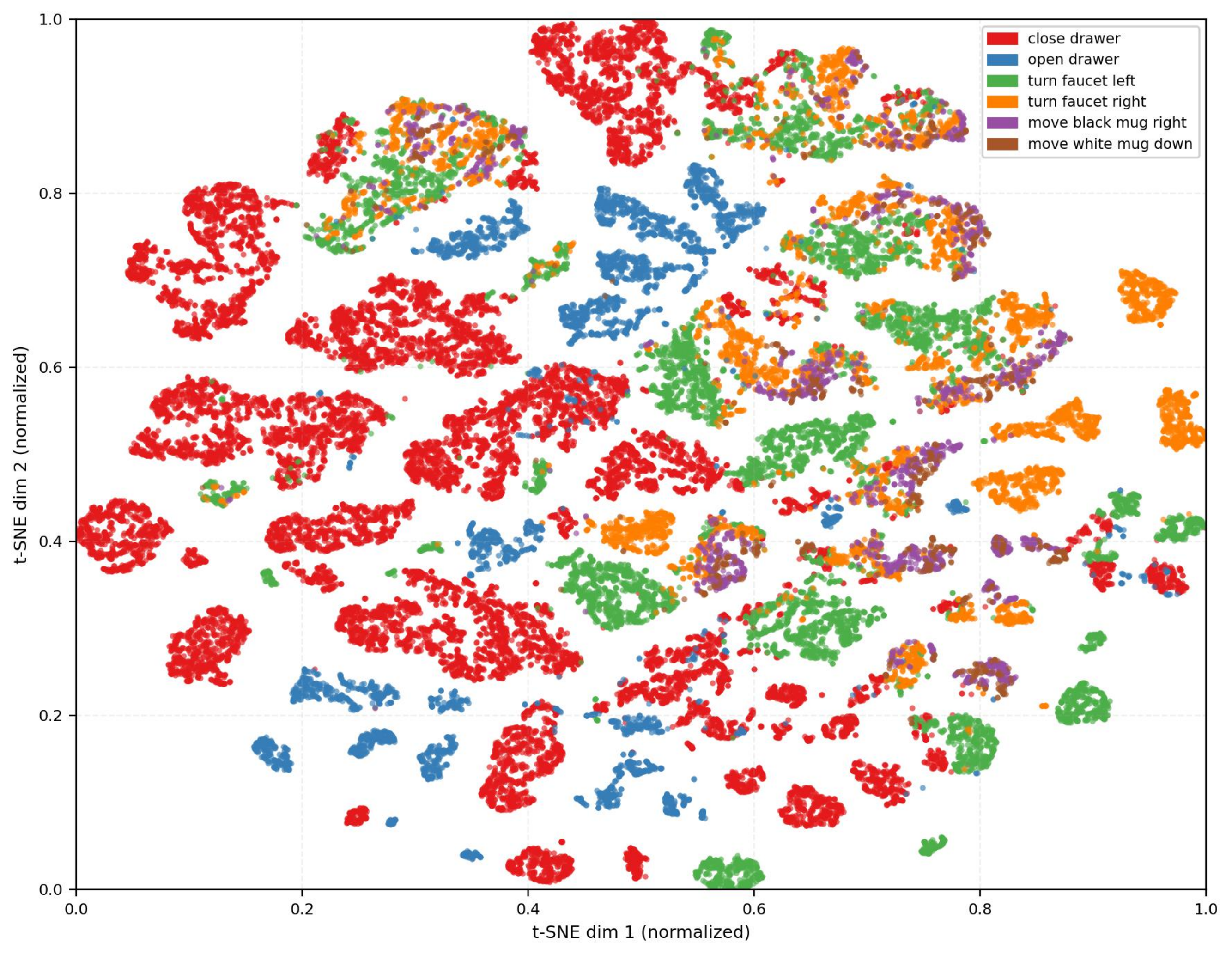} &
    \includegraphics[height=0.28\textwidth]{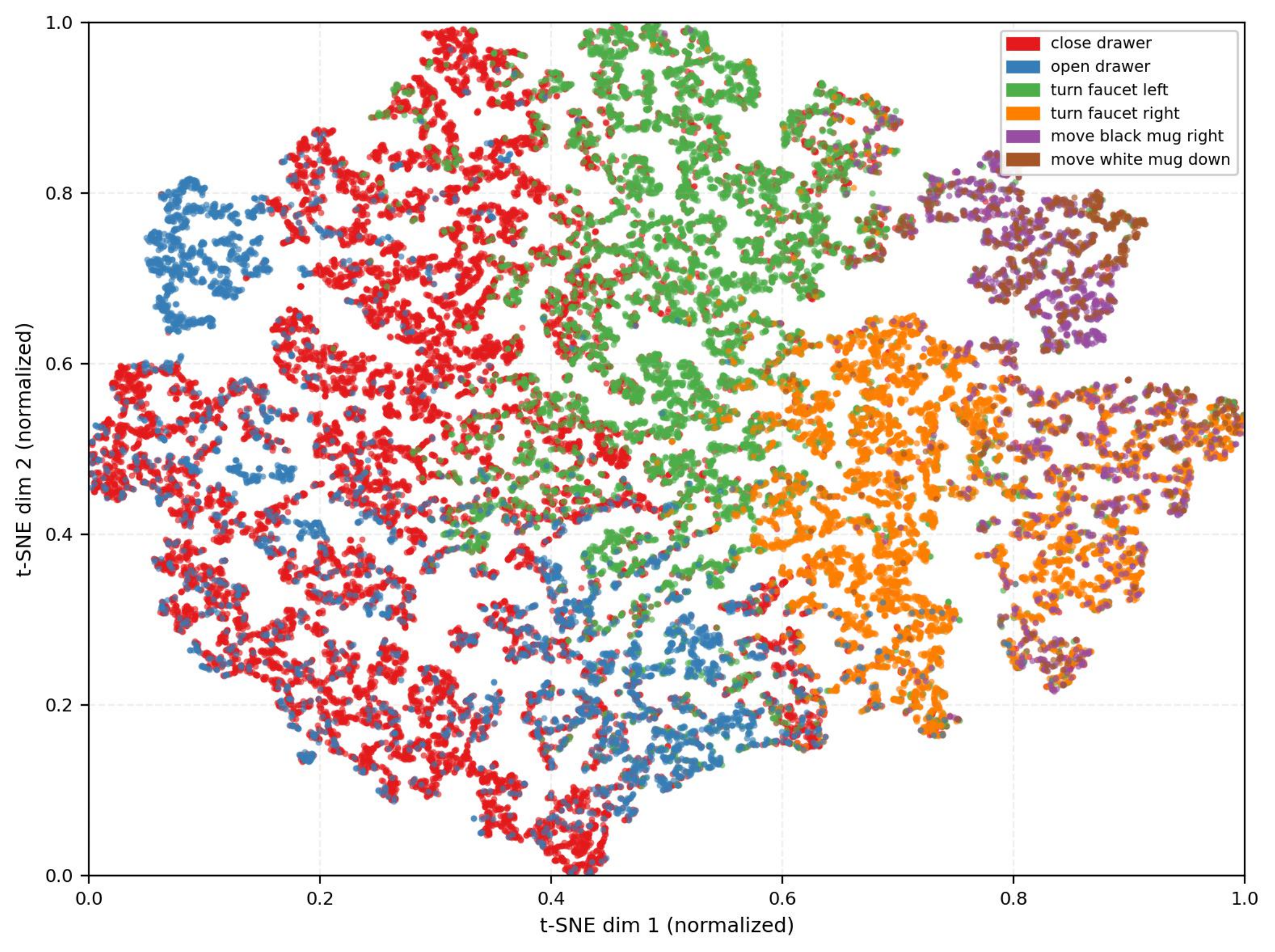} \\

    \small (a) Raw Expert States &
    \small (b) Continuous Skills (Pre-VQ) &
    \small (c) Fused Trajectories (Ours) \\

  \end{tabular}%
  }

  \caption{Evolution of latent space representations across different stages. t-SNE visualizations illustrate the progression from (a) raw expert states with high entanglement to (b) continuous skill primitives prior to vector quantization (Pre-VQ), and finally to (c) our fused trajectories. The transition demonstrates how the DASL framework effectively transforms diffuse continuous embeddings into semantically distinct and well-separated clusters, significantly enhancing representational discriminability.}
  \label{fig:tsne_skill}
\end{figure*}

\textbf{Latent Skill Analysis.} Figure \ref{fig:tsne_skill} illustrates the evolution of the latent space structure during hierarchical learning. The raw expert states in panel (a) are heavily entangled, making various task distributions nearly indistinguishable. Continuous skill modeling leads to the emergent clustering seen in panel (b), where broad task categories begin to diverge. Despite this, these pre-quantization embeddings still have diffuse boundaries and significant overlap, which limits the precision of semantic grounding. In contrast, the fused trajectories in panel (c) show a sharp increase in cluster separation and density. By using a vector quantization bottleneck and the DASL fusion mechanism, the framework compresses diffuse embeddings into compact, semantically explicit clusters. This structural refinement ensures that each atomic skill occupies a unique region in the latent manifold.

\begin{figure}[htbp]
    \centering
    \includegraphics[width=0.6\linewidth]{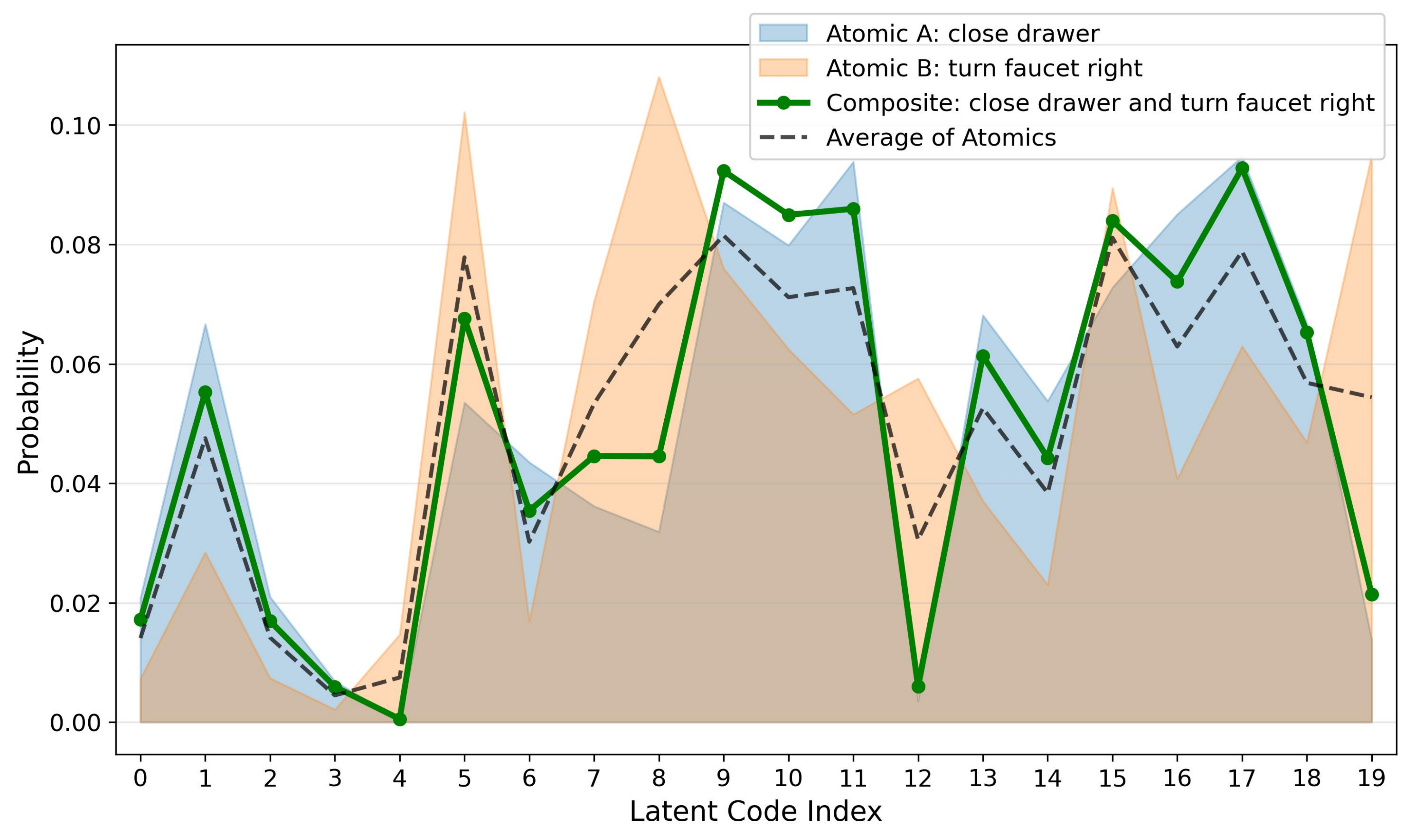}
    \caption{Visualization of Latent Skill Distributions. Comparing skill codebook usage between composite and atomic skills.}
    \label{app_latent_skill_dist}
\end{figure}

To validate DASL's capability to compose atomic skills for solving compositional manipulation tasks, we visualized the distributions of composite and atomic instructions on the LOReL Sawyer (state-based) dataset. Figure~\ref{app_latent_skill_dist} illustrates the frequency distribution of skill codebook usage for the composite task ``close drawer and turn faucet right'' alongside its constituent atomic tasks. We observe a significant distributional alignment between the composite task distribution (green solid line) and the arithmetic mean of the atomic task distributions (black dashed line). Notably, the high-probability modes corresponding to Atomic Task A (indices 9--11) and Atomic Task B (indices 5--8) are explicitly preserved in the composite distribution, while task-irrelevant skill codes in low-probability regions (indices 0--4 and 12) are successfully suppressed. This indicates that DASL captures reusable latent atomic skills and effectively solves compositional tasks by composing these atomic primitives. 

\section{Additional Ablation Studies}
\label{app_Additional_Ablation_Studies}

\subsection{Sensitivity Analysis of Frequency Ratios}
\label{app_frequency_ratio}

The sensitivity analysis in Table \ref{tab:freq_ratio_res} highlights the critical impact of the temporal interaction between the high-level policy and the low-level control. Rather than treating the update frequency ratio $I$ as an ad hoc parameter requiring exhaustive task-specific grid searches, we conceptualize it as an explicit temporal scale inherent to sequence policies. In practice, we initialize $I$ using a straightforward heuristic based on the episode budget and context length. This standard environment-dependent tuning process, identical to setting the chunk horizon $K$ or codebook size, aligns the update stride with the natural timescale of the target environment. For the LOReL environment, the chosen 1:2 ratio emerges as a highly stable optimum across the suite, achieving average success rates of 54.51\% for state inputs and 50.96\% for image inputs. Operating with a perfectly synchronous 1:1 ratio significantly impairs performance, reducing the average state success rate to 37.05\%. This degradation indicates that forcing the high-level semantic model to intervene at every discrete step disrupts atomic skill continuity and introduces unnecessary reasoning overhead.

\begin{table*}[htbp]
    \centering
    \vspace{-5pt}
    \footnotesize

    \setlength{\tabcolsep}{12pt} 
        
    \caption{Performance comparison under different frequency ratios (State)}
    \label{tab:freq_ratio_res}
    
    {
    \begin{tabular}{l c c c c c}
        \toprule
        \textbf{Task Instruction} & \textbf{Obs} & \textbf{1:1} & \textbf{1:2(ours)} & \textbf{1:4} & \textbf{1:8} \\
        \midrule
        
        \multirow{2}{*}{close drawer} & State & 70.77 & 96.92 & 78.46 & 86.15 \\
        & Image & 13.85 & 64.00& 30.77 & 26.15 \\
        
        \multirow{2}{*}{open drawer} & State & 30.00 & 31.67 & 28.33 & 0 \\
        & Image & 8.33 & 58.67 & 41.67 & 5.00 \\
        
        \multirow{2}{*}{turn faucet left} & State & 53.85 & 66.15 & 41.53 & 80 \\
        & Image & 35.38 & 41.69 & 13.85 & 58.46 \\
        
        \multirow{2}{*}{turn faucet right} & State & 1.54 & 52.31 & 16.92 & 0 \\
        & Image & 18.46 & 41.69 & 4.62 & 0.00 \\
        
        \multirow{2}{*}{move black mug right} & State & 43.08 & 66.15 & 32.31 & 0 \\
        & Image & 50.77 & 70.46 & 23.08 & 16.92 \\
        
        \multirow{2}{*}{move white mug down} & State & 23.08 & 13.85 & 1.54 & 16.92 \\
        & Image & 27.69 & 29.23 & 21.54 & 10.77 \\
        
        \midrule
        \multirow{2}{*}{\textbf{Average}} & State & 37.05 & 54.51 & 33.18 & 30.51 \\
        & Image & 25.75 & 50.96 & 22.59 & 19.55 \\
    \bottomrule
    \end{tabular}}
\end{table*}

\subsection{Additional Ablation on Action Generation Architectures}
\label{app:action_generation}

To establish a rigorous comparative baseline, we construct a pure generative dual-diffusion variant of our framework. Specifically, retaining the latent diffusion model for trajectory regularization, we replace the deterministic Decision Transformer with a secondary diffusion model dedicated exclusively to low-level action generation. Motivated by real-time control and optimization stability requirements, this section evaluates employing a Decision Transformer over such pure generative policies. While generative baselines relying on iterative denoising for action execution (such as SkillDiffuser) suffer from substantial latency (achieving merely 2.02 Hz), integrating a Decision Transformer enables the DASL framework to bypass this bottleneck, securing an inference rate of 13.61 Hz. Beyond computational efficiency, empirical results in Table \ref{tab:dt_vs_diffusion} demonstrate that embedding a deterministic Decision Transformer mitigates the optimization instability typical of pure diffusion-based hierarchical approaches. Specifically, our hybrid architecture substantially outperforms the pure generative dual-diffusion configuration, nearly doubling the success rate on the LOReL state evaluation (54.51\% vs. 27.53\%) and maintaining clear superiority in the observation setting (50.96\% vs. 42.33\%). These findings confirm that the deterministic mapping of the Decision Transformer provides a crucial stabilizing effect without compromising operational fluidity.

\begin{table}[htbp]
  \centering
  \vspace{-1em} 
  \small
  \setlength{\tabcolsep}{8pt}
  \caption{Success rate comparison across different architectures.}
  \vspace{0.5em} 
  \label{tab:dt_vs_diffusion}
  \begin{tabular}{l c c}
    \toprule
    \textbf{Architecture} & \textbf{LOReL (State)} & \textbf{LOReL (Obs)} \\
    \midrule
    DT w/ Diffusion (Ours) & \textbf{54.51} & \textbf{50.96} \\
    Dual-Diffusion (Pure Generative) & 27.53 & 42.33 \\ 
    \bottomrule
  \end{tabular}
\end{table}

\subsection{Additional Ablation on Latent Space Regularization}
\label{app:latent_regularization}

This section investigates the impact of latent space regularization by comparing the proposed diffusion module against parameter-matched baselines, as detailed in Table \ref{tab:diffusion_ablation}. Empirical results indicate that introducing basic regularization via an MLP Autoencoder or a VAE yields limited improvements over the unregularized baseline (38.26\%). In contrast, the Latent Diffusion module achieves a substantially higher success rate of 54.51\% under an identical parameter budget. These findings indicate that regularizing the latent space is inherently beneficial for policy learning. Furthermore, the performance discrepancy across parameter-matched models suggests that the performance leap stems not from increased model capacity, but from the diffusion formulation's specific capability to effectively model complex, multimodal trajectory distributions.

\begin{table}[htbp]
  \centering
  \vspace{-1em} 
  \small
  \setlength{\tabcolsep}{12pt} 
  \caption{Success rate comparison of parameter-matched regularization methods on LOReL (State).}
  \vspace{0.5em} 
  \label{tab:diffusion_ablation}
  \begin{tabular}{l c}
    \toprule
    \textbf{Regularization Method} & \textbf{LOReL (State)} \\
    \midrule
    None (Baseline) & 38.26 \\
    VAE (KL Regularization) & 41.30 \\
    MLP Autoencoder & 44.15 \\
    Latent Diffusion (Ours) & \textbf{54.51} \\
    \bottomrule
  \end{tabular}
\end{table}

\subsection{Additional Ablation on Skill Discretization}
\label{app:skill_discretization}

This section further analyzes the architectural design of discretizing continuous skills. Beyond providing enhanced human interpretability, discrete representations introduce several structural advantages for hierarchical control. Primarily, Vector Quantization (VQ) imposes an information bottleneck that prevents the entanglement of high-level semantic intents with low-level continuous control dynamics, facilitating the compositional reuse of atomic skills. Empirical evaluations in Table \ref{tab:discrete_vs_continuous} demonstrate the effectiveness of discrete representations in complex manipulation scenarios. On the LOReL state-based evaluation, the discrete approach achieves a success rate of 54.51\%, yielding a 22.56\% absolute improvement over continuous latent skills (31.95\%). This observation aligns with established findings in hierarchical control literature (e.g., LISA \cite{garg2022lisa}). While continuous skills exhibit a marginally higher success rate on simpler navigation tasks such as BabyAI GoToSeq (65.26\% vs. 63.93\%), the significant gains in complex manipulation tasks highlight the utility of the discrete bottleneck. Additionally, aligning with the discrete VQ plans utilized by primary baselines ensures a consistent framework for comparative evaluation.

\begin{table}[htbp]
  \centering
  \vspace{-1em} 
  \small
  \setlength{\tabcolsep}{8pt}
  \caption{Performance comparison between continuous and discrete skill spaces.}
  \vspace{0.5em} 
  \label{tab:discrete_vs_continuous}
  \begin{tabular}{l c c}
    \toprule
    \textbf{Skill Space} & \textbf{LOReL (State)} & \textbf{BabyAI GoToSeq 1k} \\
    \midrule
    Continuous & 31.95 & \textbf{65.26} \\
    Discrete VQ (Ours) & \textbf{54.51} & 63.93 \\
    \bottomrule
  \end{tabular}
\end{table}

\subsection{Additional Ablation on the Skill Codebook Size}
\label{app:codebook_size}

This section investigates the sensitivity of the proposed framework to the skill codebook size, denoted as $K$. Table \ref{tab:codebook_size} presents the empirical performance across varying values of $K$ on the LOReL evaluation. The results indicate that a restricted codebook capacity ($K \le 10$) underfits the inherent skill diversity of complex tasks, yielding sub-optimal performance. Conversely, excessively enlarging the codebook size (e.g., $K=40$) hinders optimization and diminishes code utilization, leading to a noticeable performance degradation (34.80\% on the state-based evaluation). The optimal balance between representational capacity and optimization stability is achieved at $K=20$, which secures the highest success rates in both state (54.51\%) and observation-based (50.96\%) settings. In practice, without prior knowledge of the optimal task decomposition, determining the initial $K$ through a small-scale validation sweep. During early training, the ideal codebook size can be empirically adjusted by monitoring the codebook entropy: $K$ should be increased if the capacity saturates, or decreased if a large portion of the codes remains inactive.

\begin{table}[htbp]
  \centering
  \vspace{-1em} 
  \small
  \setlength{\tabcolsep}{8pt}
  \caption{Performance comparison across different skill codebook sizes ($K$) on LOReL tasks.}
  \vspace{0.5em} 
  \label{tab:codebook_size}
  \begin{tabular}{l c c}
    \toprule
    \textbf{Codebook Size ($K$)} & \textbf{LOReL (State)} & \textbf{LOReL (Obs) } \\
    \midrule
    5 & 47.53 & 36.36 \\
    10 & 45.97 & 32.58 \\
    20 (Ours) & \textbf{54.51} & \textbf{50.96} \\
    40 & 34.80 & 37.40 \\
    \bottomrule
  \end{tabular}
    \vspace{-1em}
\end{table}

\subsection{Additional Ablation on Temporal Embeddings and OOD Horizons}
\label{app:temporal_ood}

This section investigates the role of temporal embeddings and the model's robustness to out-of-distribution (OOD) episode lengths. A potential concern when utilizing temporal embeddings is the risk of the policy overfitting to specific time intervals observed during training, leading to brittle performance when evaluated over longer horizons. To systematically assess this, empirical evaluations were conducted on the LOReL state-based benchmark using extended episode horizons up to 3.0$\times$ the maximum training length ($T_{\max}$). The analysis compares four distinct temporal extrapolation strategies to process the time step $t$:

\begin{itemize}
    \item \textbf{Clamp (Ours):} $t = \min(t, T_{\max})$. This strategy caps the temporal signal at the maximum training length, signaling to the model that the task has entered a sustained terminal phase.
    \item \textbf{Modular:} $t = t \pmod{T_{\max}}$. This approach cyclically resets the temporal signal based on the original training horizon.
    \item \textbf{Relative:} $t = t \times \frac{T_{\max}}{\text{horizon}}$. This method proportionally scales the temporal signal relative to the new, extended evaluation horizon, stretching the embeddings to match the original distribution.
    \item \textbf{Zero:} $t = 0$. This serves as a pure ablation baseline by completely removing the temporal signal.
\end{itemize}

Table \ref{tab:temporal_ood} demonstrates substantial robustness against extended horizons. The proposed Clamp strategy yields the highest success rates, achieving 69.3\% at a 3.0$\times$ horizon, notably exceeding its in-distribution performance (61.9\%). This stable and improving performance across OOD lengths effectively rules out temporal overfitting. It indicates the model leverages temporal embeddings to infer macroscopic task phases (e.g., beginning, execution, and endgame) rather than memorizing rigid step counts. Furthermore, ablating the temporal signal entirely (Zero strategy) degrades performance by approximately 20\% across all horizons, confirming temporal conditioning remains an essential global phase indicator. Ultimately, these results validate temporal embeddings as flexible, stage-aware indicators rather than brittle per-step triggers.

\begin{table}[htbp]
  \centering
  \vspace{-1em}
  \small
  \setlength{\tabcolsep}{8pt}
  \caption{Performance evaluation on LOReL (State) across out-of-distribution (OOD) episode horizons using different temporal embedding extrapolation strategies. Success rates (SR (\%)) are reported.}
  \vspace{0.5em}
  \label{tab:temporal_ood}
  \begin{tabular}{l c c c c}
    \toprule
    \textbf{Episode Length} & \textbf{Clamp (Ours)} & \textbf{Modular} & \textbf{Relative} & \textbf{Zero} \\
    \midrule
    1.0$\times$ (In-dist) & \textbf{61.9} & 60.2 & 57.6 & 40.7 \\
    2.0$\times$ (OOD) & \textbf{66.7} & 62.8 & 54.5 & 38.1 \\
    3.0$\times$ (OOD) & \textbf{69.3} & 64.9 & 63.2 & 45.9 \\
    \bottomrule
  \end{tabular}
    \vspace{-1em}
\end{table}

\section{Real-world Experiment}
\label{app_Experiment}

\begin{figure}[htbp]
    \centering
    \includegraphics[width=0.6\linewidth]{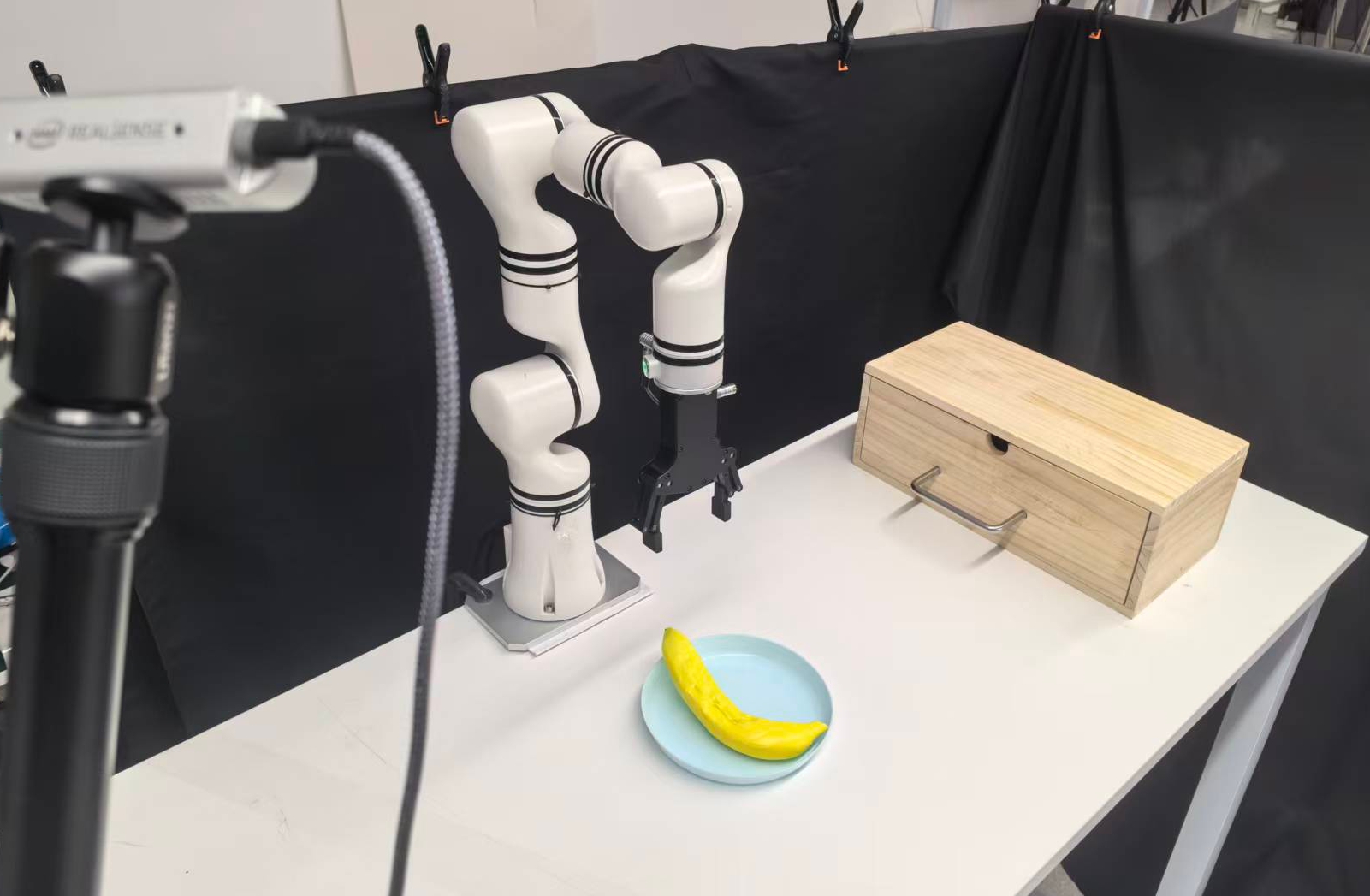}
    \caption{\textbf{Real-world experimental setup.} We evaluate the proposed DASL framework on a Realman robot. The setup involves various interactive objects, such as a drawer and a banana on a plate, to validate the performance of the model across atomic and composite manipulation tasks.} 
    \label{fig:real_world_setup}
\end{figure}

To evaluate the practical viability of DASL in real-world scenarios, we deploy the framework in a physical setup equipped with a single-arm Realman manipulator and a statically mounted third-person RealSense RGB camera. As illustrated in Figure~\ref{fig:real_world_setup}, all models are trained and evaluated using combined state and image inputs. We design five distinct manipulation tasks, categorized into fundamental atomic skills and complex compositional tasks. 
The atomic tasks comprise \textit{open drawer}, \textit{pick up the banana from the plate}, and \textit{close drawer}, for which we collect 60, 40, and 20 demonstration trajectories, respectively. 
The compositional tasks include \textit{open drawer and close drawer} and \textit{open drawer and pick up banana from the plate}, with training data restricted to exactly 10 trajectories each. This deliberately imbalanced data distribution explicitly evaluates the capacity of the model to generalize to complex compositional tasks despite limited demonstrations. 

To assess the robustness of the proposed method, we execute each task for 20 trials to compute the average success rate. As shown in Table~\ref{tab:real_world_results}, empirical results demonstrate that DASL generates stable predictions and effectively completes both atomic and complex sequential instructions in the physical environment. Furthermore, under identical observation conditions, DASL significantly outperforms the LISA baseline, achieving an overall average success rate of 88\% compared to 56\% for LISA.

\begin{table}[htbp]
    \centering
    \small
    \setlength{\tabcolsep}{12pt} 
    
    \caption{\textbf{Success rates of real-world robot experiments.} We evaluate LISA and DASL on a Realman robot across five manipulation tasks, categorized into atomic and compositional instructions. The results report the success rate (\%) over 20 trials for each task. The best results are highlighted in \textbf{bold}.}
    \vspace{10pt} 
    
    \begin{tabular}{l c c}
        \toprule
        \textbf{Task Instruction} & \textbf{LISA} & \textbf{DASL (ours)} \\
        \midrule
        close drawer & 70 & \textbf{95} \\
        open drawer & 55 & \textbf{90} \\
        pick up the banana from the plate & 60 & \textbf{100} \\
        open drawer and close drawer & 50 & \textbf{75} \\
        open drawer and pick up banana from the plate & 45 & \textbf{80} \\
        \midrule
        \textbf{Average} & 56 & \textbf{88} \\
        \bottomrule
    \end{tabular}
    \label{tab:real_world_results}
\end{table}

\begin{figure}[p] %
  \centering

  \newcommand{\frameimg}[1]{\includegraphics[width=0.135\textwidth, height=0.11\textwidth]{#1}}

  \begin{subfigure}{\textwidth}
    \centering
    \frameimg{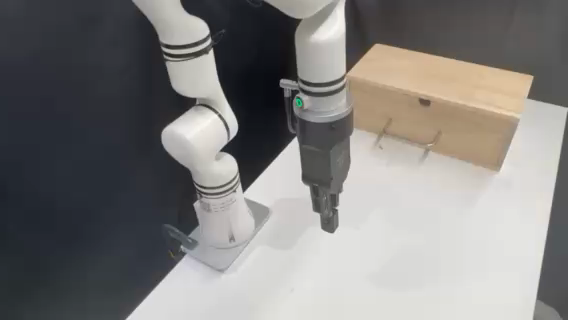}\hfill
    \frameimg{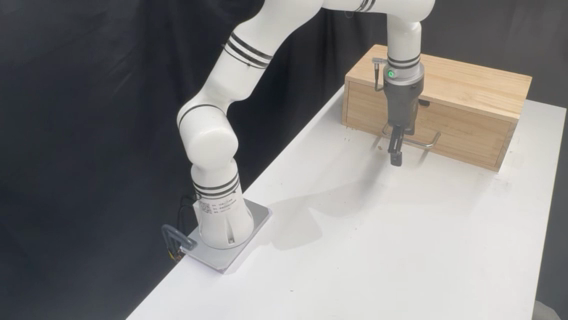}\hfill
    \frameimg{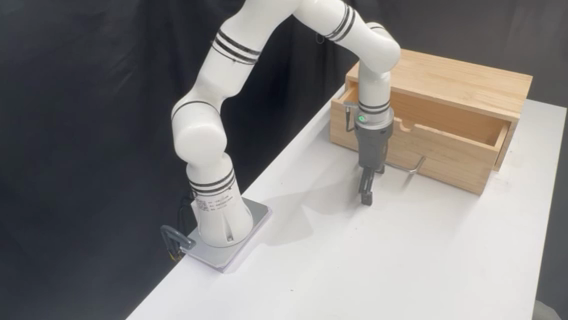}\hfill
    \frameimg{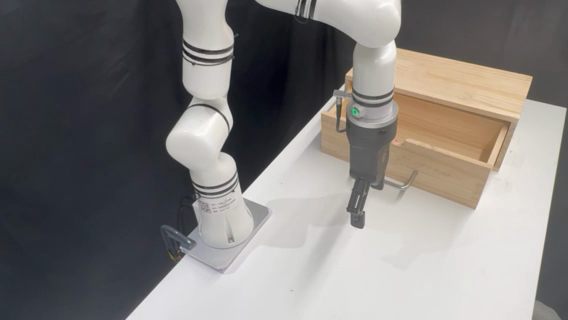}\hfill
    \frameimg{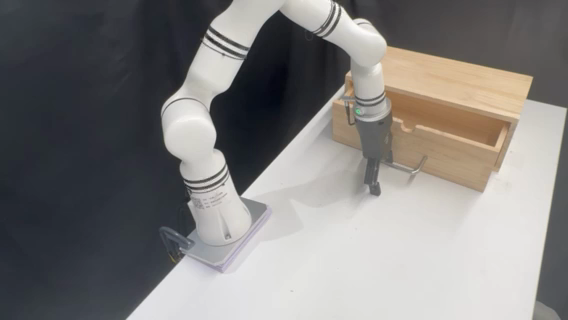}\hfill
    \frameimg{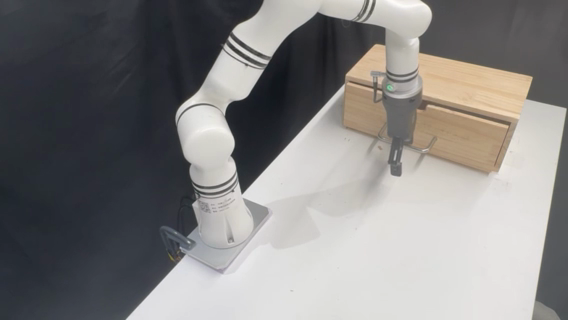}\hfill
    \frameimg{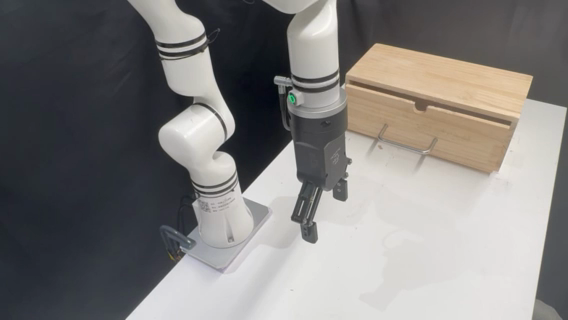}
    \caption{LISA Baseline: open drawer and close drawer}
    \vspace{4pt} %
  \end{subfigure}

  \begin{subfigure}{\textwidth}
    \centering
    \frameimg{fig/realman/open-close/frames_top/episode_0/frame_000000.png}\hfill
    \frameimg{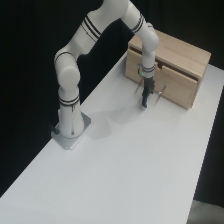}\hfill
    \frameimg{fig/realman/open-close/frames_top/episode_0/frame_000027.png}\hfill
    \frameimg{fig/realman/open-close/frames_top/episode_0/frame_000049.png}\hfill
    \frameimg{fig/realman/open-close/frames_top/episode_0/frame_000055.png}\hfill
    \frameimg{fig/realman/open-close/frames_top/episode_0/frame_000067.png}\hfill
    \frameimg{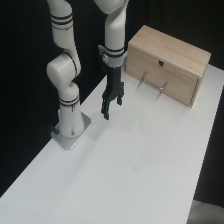}
    \caption{DASL (Ours): open drawer and close drawer}
    \vspace{8pt}
  \end{subfigure}

  \begin{subfigure}{\textwidth}
    \centering
    \frameimg{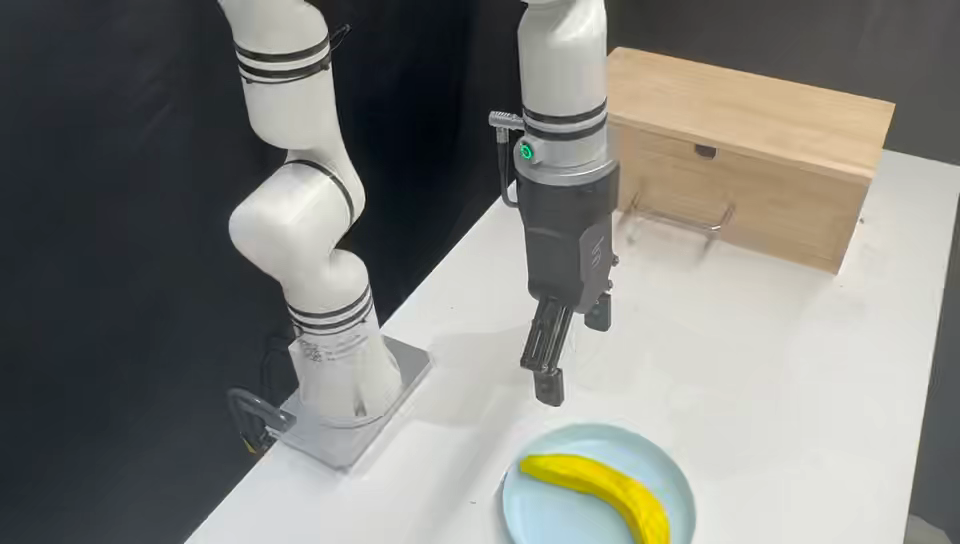}\hfill
    \frameimg{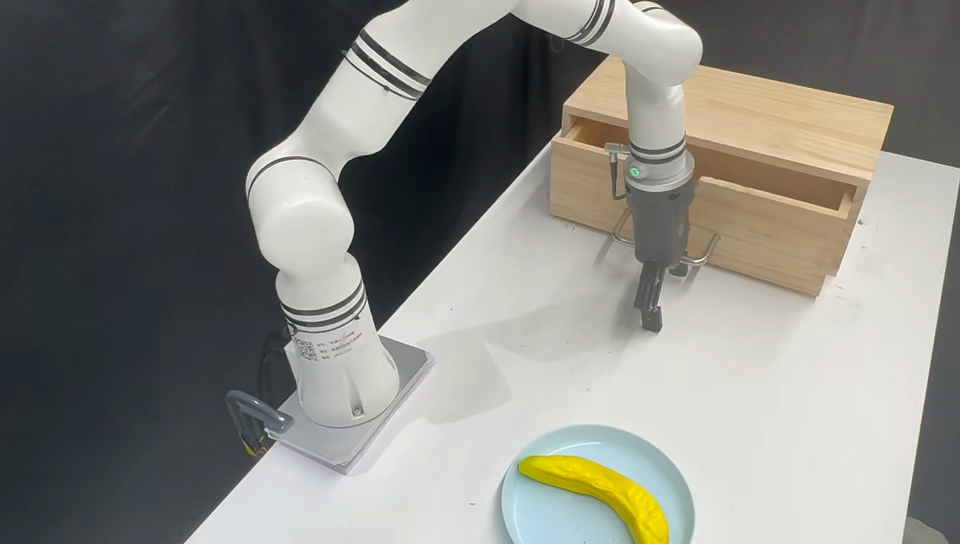}\hfill
    \frameimg{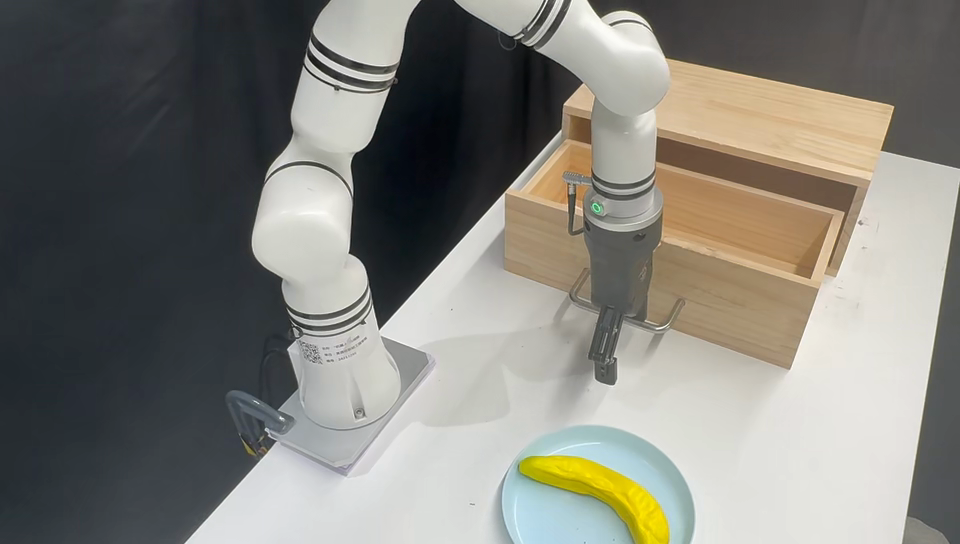}\hfill
    \frameimg{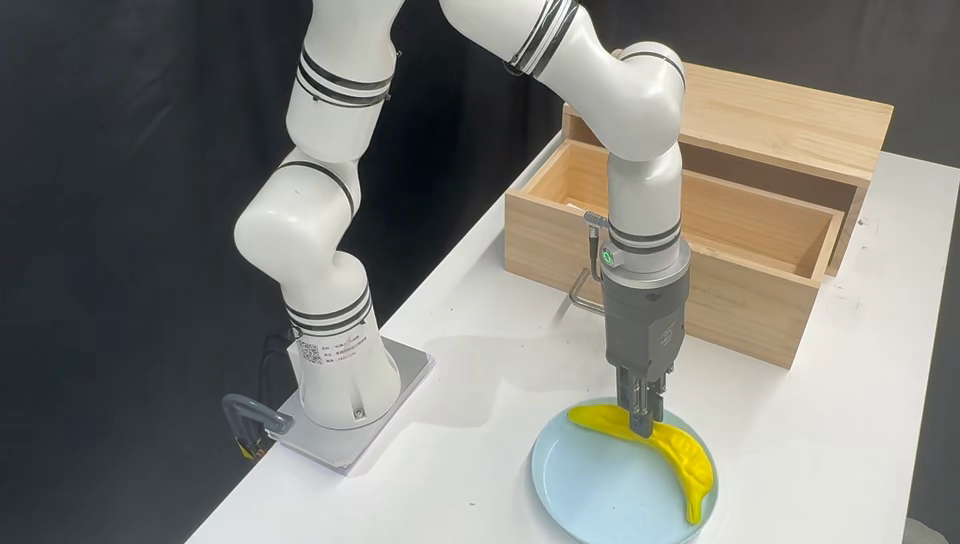}\hfill
    \frameimg{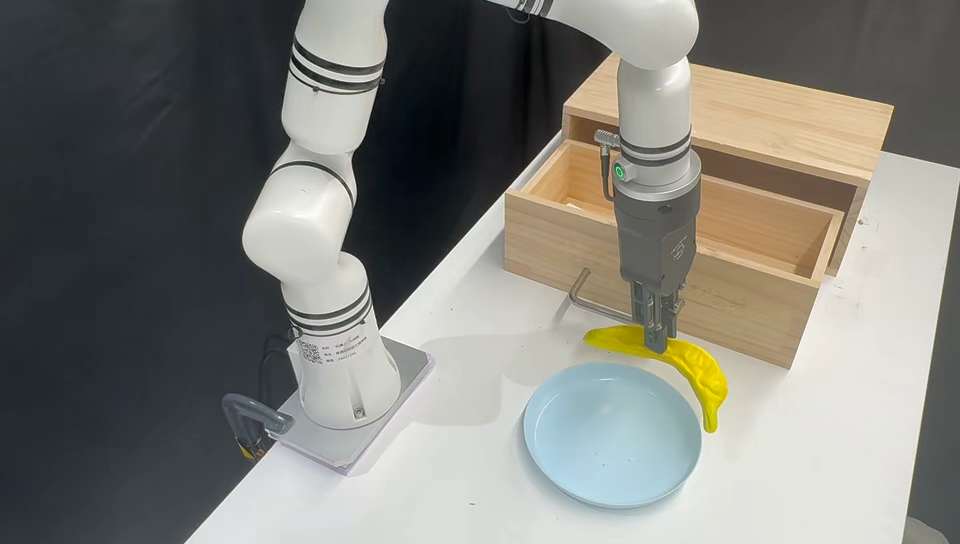}\hfill
    \frameimg{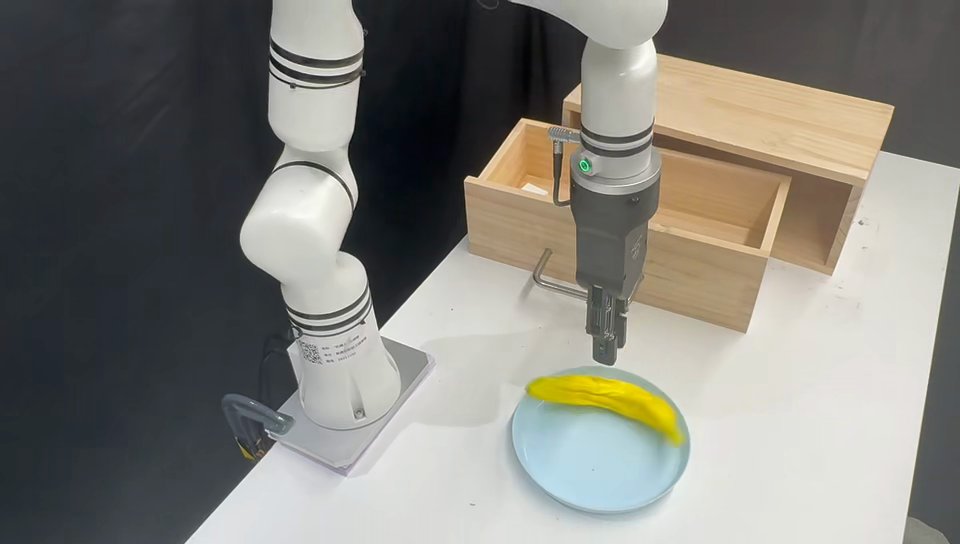}\hfill
    \frameimg{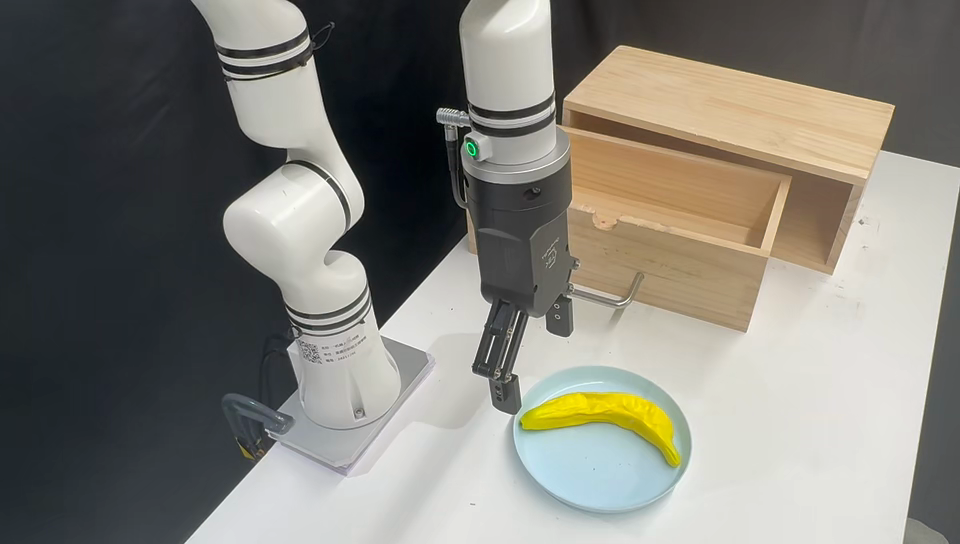}
    \caption{LISA Baseline: open drawer and pick up banana from the plate}
    \vspace{4pt}
  \end{subfigure}

  \begin{subfigure}{\textwidth}
    \centering
    \frameimg{fig/realman/open-pick-place/frames_top/episode_0/frame_000000.png}\hfill
    \frameimg{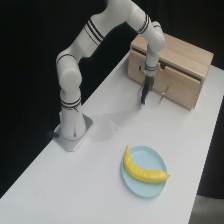}\hfill
    \frameimg{fig/realman/open-pick-place/frames_top/episode_0/frame_000024.png}\hfill
    \frameimg{fig/realman/open-pick-place/frames_top/episode_0/frame_000037.png}\hfill
    \frameimg{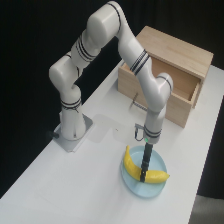}\hfill
    \frameimg{fig/realman/open-pick-place/frames_top/episode_0/frame_000089.png}\hfill
    \frameimg{fig/realman/open-pick-place/frames_top/episode_0/frame_000095.png}
    \caption{DASL (Ours): open drawer and pick up banana from the plate}
  \end{subfigure}

\caption{Qualitative comparison on compositional tasks. While LISA fails via incomplete closing (a) and unstable grasping (c), DASL (b, d) ensures precise execution and robust skill transitions.}
  \label{fig:realman_comparison}

  \vspace{1.5em}

  \begin{subfigure}{\textwidth}
    \centering
    \frameimg{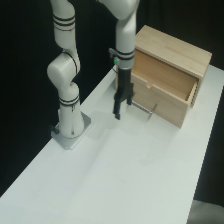}\hfill
    \frameimg{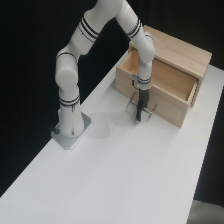}\hfill
    \frameimg{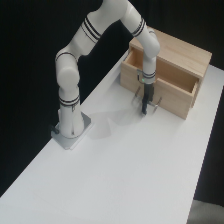}\hfill
    \frameimg{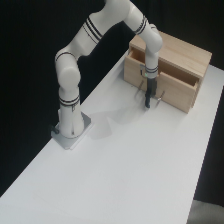}\hfill
    \frameimg{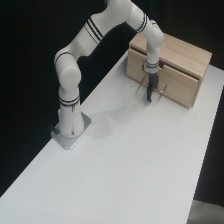}\hfill
    \frameimg{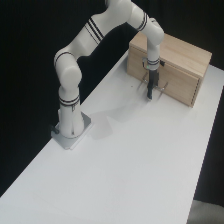}\hfill
    \frameimg{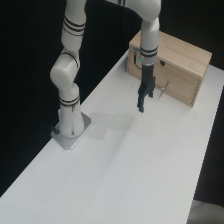}
    \caption{Atomic Task: close drawer}
    \vspace{4pt} 
  \end{subfigure}

  \begin{subfigure}{\textwidth}
    \centering
    \frameimg{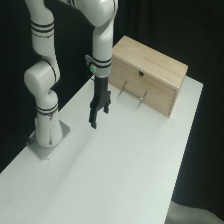}\hfill
    \frameimg{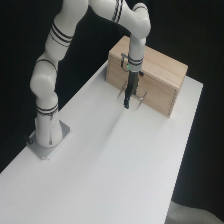}\hfill
    \frameimg{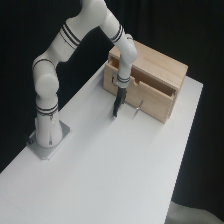}\hfill
    \frameimg{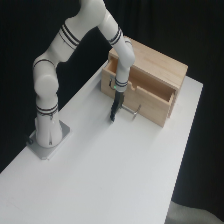}\hfill
    \frameimg{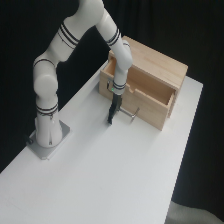}\hfill
    \frameimg{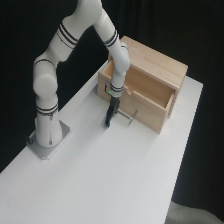}\hfill
    \frameimg{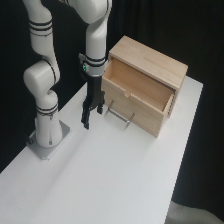}
    \caption{Atomic Task: open drawer}
    \vspace{4pt}
  \end{subfigure}

  \begin{subfigure}{\textwidth}
    \centering
    \frameimg{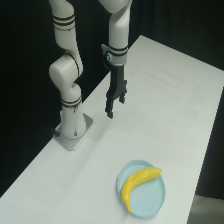}\hfill
    \frameimg{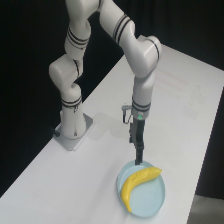}\hfill
    \frameimg{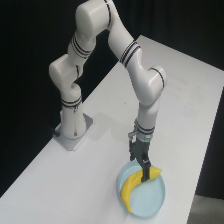}\hfill
    \frameimg{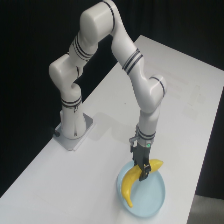}\hfill
    \frameimg{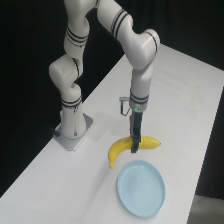}\hfill
    \frameimg{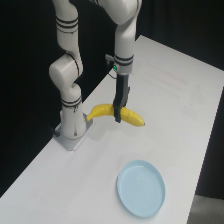}\hfill
    \frameimg{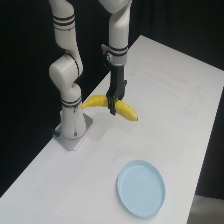}
    \caption{Atomic Task: pick up the banana from the plate}
    \vspace{4pt}
  \end{subfigure}

  \begin{subfigure}{\textwidth}
    \centering
    \frameimg{fig/realman/open-close/frames_top/episode_0/frame_000000.png}\hfill
    \frameimg{fig/realman/open-close/frames_top/episode_0/frame_000024.png}\hfill
    \frameimg{fig/realman/open-close/frames_top/episode_0/frame_000027.png}\hfill
    \frameimg{fig/realman/open-close/frames_top/episode_0/frame_000049.png}\hfill
    \frameimg{fig/realman/open-close/frames_top/episode_0/frame_000055.png}\hfill
    \frameimg{fig/realman/open-close/frames_top/episode_0/frame_000067.png}\hfill
    \frameimg{fig/realman/open-close/frames_top/episode_0/frame_000092.png}
    \caption{Compositional Task: open drawer and close drawer}
    \vspace{4pt}
  \end{subfigure}

  \begin{subfigure}{\textwidth}
    \centering
    \frameimg{fig/realman/open-pick-place/frames_top/episode_0/frame_000000.png}\hfill
    \frameimg{fig/realman/open-pick-place/frames_top/episode_0/frame_000018.png}\hfill
    \frameimg{fig/realman/open-pick-place/frames_top/episode_0/frame_000024.png}\hfill
    \frameimg{fig/realman/open-pick-place/frames_top/episode_0/frame_000037.png}\hfill
    \frameimg{fig/realman/open-pick-place/frames_top/episode_0/frame_000053.png}\hfill
    \frameimg{fig/realman/open-pick-place/frames_top/episode_0/frame_000089.png}\hfill
    \frameimg{fig/realman/open-pick-place/frames_top/episode_0/frame_000095.png}
    \caption{Compositional Task: open drawer and pick up banana from the plate}
  \end{subfigure}

  \caption{Qualitative visualizations of the proposed DASL framework executing atomic and compositional tasks on a Realman robot. Each row shows a key-frame sequence of a successful trial.}
  \label{fig:realman_results_all}

\end{figure}

This performance gap is particularly evident on the complex compositional tasks. Figure~\ref{fig:realman_results_all} presents qualitative visualizations of DASL successfully executing all defined atomic and compositional tasks. To further illustrate our advantages, Figure~\ref{fig:realman_comparison} provides a direct visual comparison against LISA on compositional tasks. As observed, the baseline struggles with seamless execution. Specifically, in the open and close drawer task (Figure~\ref{fig:realman_comparison}a), LISA fails to apply sufficient displacement, leaving the drawer partially open. Furthermore, in the open drawer and pick up banana task (Figure~\ref{fig:realman_comparison}c), LISA exhibits an unstable grasp, causing the banana to drop immediately after being lifted. In contrast, DASL demonstrates smoother and more robust skill transitions. It achieves precise execution, completely closing the drawer (Figure~\ref{fig:realman_comparison}b) and maintaining a firm grasp on the object (Figure~\ref{fig:realman_comparison}d). These visualizations confirm that the asynchronous architecture effectively chains distinct skills for complex instructions, validating its practical utility in physical environments.

\end{document}